\documentclass{article} 
\usepackage{iclr2027_conference,times}

\usepackage{amsmath,amsfonts,bm}

\def\eqref#1{equation~\ref{#1}}

\def\1{\bm{1}}

\newcommand{\test}{\mathcal{D_{\mathrm{test}}}}

\DeclareMathAlphabet{\mathsfit}{\encodingdefault}{\sfdefault}{m}{sl}
\SetMathAlphabet{\mathsfit}{bold}{\encodingdefault}{\sfdefault}{bx}{n}

\usepackage{hyperref}
\usepackage{url}
\usepackage{booktabs}
\usepackage{longtable}
\usepackage{algorithm}
\usepackage{amsmath}
\usepackage{algpseudocode}
\usepackage{fix-cm}
\usepackage{comment}

\usepackage[table]{xcolor}
\usepackage{graphicx}
\definecolor{ModelBand}{HTML}{E8F2F9}
\definecolor{ModelInk}{HTML}{365B76}
\definecolor{GainInk}{HTML}{406C89}
\definecolor{DropInk}{HTML}{95635E}
\newcommand{\modelband}[1]{\rowcolor{ModelBand}\multicolumn{21}{l}{\strut\color{ModelInk}\itshape #1}\\}
\newcommand{\gainpos}[1]{\textcolor{GainInk}{#1}}
\newcommand{\gainneg}[1]{\textcolor{DropInk}{#1}}

\definecolor{AblationBand}{HTML}{E8F2F9}
\definecolor{AblationInk}{HTML}{365B76}

\definecolor{AgiBand}{HTML}{E8F2F9}
\definecolor{AgiInk}{HTML}{365B76}
\definecolor{AgiGain}{HTML}{406C89}
\definecolor{AgiDrop}{HTML}{95635E}
\definecolor{citecolor}{HTML}{0071BC}

\definecolor{OverallBg}{RGB}{240,240,240}

\title{Towards a Systematic Study of Embodied System 2 over Full Task Execution Cycles}

\author{
\parbox{0.95\textwidth}{\centering
Yuxi Wei\textsuperscript{1,2}\quad
Ye Fang\textsuperscript{2,3}\quad
Shuai Yang\textsuperscript{2,4}\\[4pt]
Tong Wu\textsuperscript{2,3}\quad
Tai Wang\textsuperscript{2,*}\quad
Xiaojuan Qi\textsuperscript{1,*}\\[8pt]
\normalfont\small
\textsuperscript{1}The University of Hong Kong\quad
\textsuperscript{2}Morphi Robot\\[3pt]
\textsuperscript{3}Fudan University\quad
\textsuperscript{4}Shanghai Jiao Tong University\\[5pt]
\textsuperscript{*}Co-corresponding authors
}
}

\iclrfinalcopy
\hypersetup{pdfauthor={Yuxi Wei, Ye Fang, Shuai Yang, Tong Wu, Tai Wang, Xiaojuan Qi}}
\begin{document}

\maketitle
\lhead{Preprint}
\begin{center}
\small\textbf{Project website:} \url{https://vfishc.github.io/scoper/}
\end{center}

\begin{abstract}
An embodied brain, or System 2, must turn an abstract goal into actions while discovering relevant scene conditions and checking whether executed skills achieved their intended effects. Open-loop evaluations cannot test observations shaped by an agent’s decisions, while some closed-loop evaluations provide limited evidence during execution or assume idealized skill outcomes. We introduce SCOPER, a suite for studying observation-grounded planning, execution, and recovery. SCOPER-bench combines high-level goals, ongoing visual observations, a shared rule-based System 1 for low-level actions, and controlled execution failures. Its construction and validation pipeline yields 870 tasks across BEHAVIOR and ProcTHOR, complemented by about 3,500 situated questions tied to the evaluated agent’s execution. Benchmark results reveal difficulties in retaining scene evidence, monitoring progress, and recovering from unexpected outcomes. To address these difficulties, SCOPER-harness combines Evidence Memory and Progress Verification with a dynamic task agenda and bounded exploration, without additional model training. Across two evaluated open-source backbones and two simulators, the harness improves task success by 11.1–28.6\%; a harnessed 9B model outperforms a directly executed 27B model on both simulators. In an offline next-action evaluation using observations from 200 real-robot demonstrations, the harness also improves full-call accuracy by 14.7\% and 15.3\%. SCOPER provides a setting for diagnosing and improving embodied reasoning throughout a complete task execution cycle.
\end{abstract}

\section{Introduction}
\label{sec:introduction}

An embodied agent must do more than turn a goal into a sequence of actions~\citep{huang2022zeroshotplanner,singh2022progprompt,song2022llmplanner}. Its high-level controller, which we call the embodied brain or System 2, must decide what information to seek, interpret visual observations~\citep{driess2023palme,zeng2022socratic}, select skills~\citep{mu2023embodiedgpt,ahn2024autort}, and determine whether those skills achieved their intended effects~\citep{guo2023doremi,liu2023reflect}. A low-level System 1 executes the selected skills, following the separation of language-based planning and skill execution explored in prior embodied systems~\citep{ahn2022saycan,huang2022innermonologue,liang2022codeaspolicies,huang2023voxposer}. In a partially observed environment, exploration may reveal new requirements, while an unsuccessful action may invalidate the current plan~\citep{guo2023doremi,liu2023reflect}. These demands are coupled: what the agent observes shapes what it should do~\citep{wu2023tapa,wu2023tidybot}, and what happens after it acts shapes what it should do next. Studying them together is essential to understanding whether an embodied brain can guide long-horizon behavior through a complete closed-loop task~\citep{yenamandra2023homerobot}.

  \begin{figure}[t]
      \centering
      \includegraphics[width=1.0\linewidth,trim=0 3bp 0 2bp,clip]{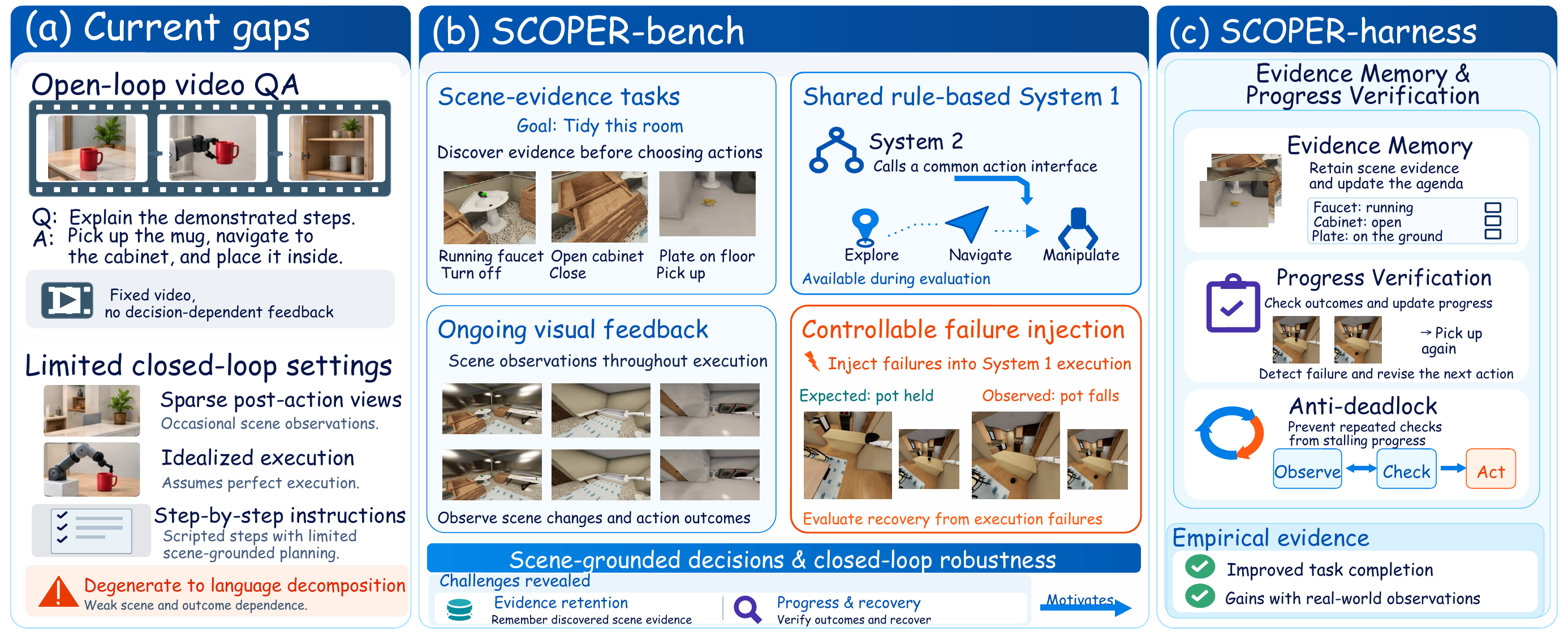}
      \vspace{-1mm}
      \caption{ \small
          Embodied System~2 throughout closed-loop task execution.
          Open-loop reference trajectories cannot test decision-dependent interaction;
          closed-loop settings omitting execution-time observations or failures leave
          key capabilities untested.
          The SCOPER suite combines high-level goals, ongoing visual observations,
          and controlled failures in SCOPER-bench, with Evidence Memory and
          Progress Verification in SCOPER-harness.
      }
      \label{fig:teaser}
  \vspace{-2mm}
  \end{figure}

  Figure~\ref{fig:teaser} summarizes these demands and their treatment by SCOPER-bench and SCOPER-harness. Existing evaluations inadequately capture these coupled demands. Open-loop evaluations based on recorded egocentric videos assess interpretation of fixed observations~\citep{qiu2024egoplanbench2,dang2025ecbench,majumdar2024openeqa,chen2023egoplan,jia2022egotaskqa,cheng2024videgothink}, but
  cannot test whether an agent acquires missing evidence, adjusts task decomposition through interaction, or recovers from its own execution failures. Existing closed-loop benchmarks~\citep{yang2025embodiedbench,cheng2025embodiedeval,srivastava2021behavior}
  also leave gaps when they omit intermediate execution observations, assume idealized execution, or provide detailed procedural instructions. These choices limit evaluation of
  evidence accumulation, outcome monitoring, and recovery, and can reduce planning to instruction-driven subtask sequencing with limited need for visual reasoning. The action--result loop alone therefore cannot establish conditions for studying embodied System~2.

  We introduce the \textbf{SCOPER suite} for studying System~2 in closed-loop, observation-grounded planning, execution, and recovery, comprising \textbf{SCOPER-bench} and
  \textbf{SCOPER-harness}. SCOPER-bench requires agents to infer subtasks from abstract goals and multi-frame visual observations throughout execution; construction-time decompositions serve validation. A unified rule-based System~1 supplies shared execution infrastructure, and a scalable construction and validation pipeline yields 870 tasks across two simulators (Section~\ref{sec:benchmark}). Controlled failure injection tests whether agents detect unmet expectations and replan to resume progress. Evaluations reveal persistent difficulties in acquiring and maintaining scene evidence for planning, monitoring outcomes, and recovering from failures.

  SCOPER-harness addresses these
  difficulties through \textbf{Evidence Memory}, which maintains explicit language descriptions of observed scene information, and \textbf{Progress Verification}, which checks expected
  outcomes against observations after each execution step. A dynamic agenda organizes pending subtasks and is updated using accumulated evidence and verification results. Because
  repeated exploration and checking can delay execution, an anti-deadlock mechanism limits unproductive repetition to balance information gathering with task advancement.

  Experiments show that complete execution remains challenging even with shared low-level skills: the strongest directly evaluated model achieves 50.5\% task success on BEHAVIOR and 35.1\% on ProcTHOR. Across evaluated open-source models, SCOPER-harness consistently improves performance, with gains of up to 28.6 percentage points, enabling a smaller model to perform comparably to the strongest evaluated model. Evaluated embodied-finetuned models also underperform general-purpose models, indicating that embodied specialization does not necessarily yield
  stronger closed-loop reasoning in this setting. Tests with real-world observations further show that benchmark-guided harness refinements improve performance, particularly in memory-
  dependent and failure-recovery cases.

  Our contributions are threefold:   
(1) \textbf{SCOPER-bench:}
      A validated cross-simulator benchmark jointly testing observation-grounded task decomposition, execution, and failure recovery.
(2) \textbf{SCOPER-harness:}
      A harness connecting scene evidence and verified outcomes to adaptive task agendas and sustained execution.
(3) \textbf{Experiments and findings:}
      We identify gaps in scene evidence discovery and failure verification and design a targeted harness that consistently improves performance across evaluated models, with further gains on real-world observation tests.

\section{Related Work}

  \paragraph{Open-loop Evaluation of Embodied Reasoning.}
  Recent benchmarks evaluate embodied understanding and planning
  from recorded egocentric observations by asking the models to answer some open-loop questions.~\citep{grauman2021ego4d,grauman2023egoexo4d,cheng2023egothink,mangalam2023egoschema,yang2025egolife}. EgoPlan-Bench2~\citep{qiu2024egoplanbench2} tests planning
  in everyday scenarios, and ECBench~\citep{dang2025ecbench} examines diverse embodied cognitive
  abilities. These protocols probe
  reasoning over visual evidence, but observations remain independent
  of the evaluated model's decisions. They therefore cannot establish
  whether an agent can acquire missing evidence through action,
  revise task decomposition as execution unfolds, or recover from
  its own execution failures. SCOPER-bench tests these demands through decision-dependent observations and outcome-guided planning and recovery.

  \paragraph{Closed-loop Embodied Benchmarks.}
  Interactive benchmarks~\citep{shridhar2019alfred,shridhar2020alfworld,padmakumar2021teach,puig2018virtualhome} increasingly evaluate embodied agents through
  simulated task execution~\citep{savva2019habitat,szot2021habitat2,puig2023habitat3,deitke2020robothor,mees2021calvin,liu2023libero}. EmbodiedBench~\citep{yang2025embodiedbench} and EmbodiedEval~\citep{cheng2025embodiedeval} extend
  evaluation to visual navigation, interaction, and planning, but their
  standard stepwise interfaces emphasize observations at action
  boundaries rather than evidence acquired within skill execution.
  Moreover, action-validity feedback does not systematically test
  whether agents detect and recover from deviations between intended
  and actual skill effects. BEHAVIOR~\citep{srivastava2021behavior,li2024behavior1k,li2021igibson2} evaluates complex household
  activities through integrated planning and control, making high-level
  reasoning difficult to disentangle from low-level execution in
  end-to-end performance. SCOPER-bench isolates these System~2 demands using shared, controllable System~1 skills, multi-frame execution observations, explicit failure injection, and high-level goals.

    \paragraph{Agentic designs for embodied intelligence.}
  Embodied frameworks combine planning, memory, and execution
  feedback to sustain long-horizon behavior~\citep{rana2023sayplan,wang2023voyager,yao2022react,shinn2023reflexion,wang2023deps}. Cortex~\citep{peng2026cortex} aligns VLM
  planning with VLA execution through standardized subtasks and
  semantic memory; Thea~\citep{wang2026thea} integrates persistent scene context with
  action-outcome evaluation; Mimir~\citep{xu2026mimir} grounds an evolving task agenda
  in world memory and explores under insufficient evidence.
  SCOPER-harness emphasizes how scene discovery determines an abstract goal's required work, coupling exploration, evidence-driven agenda revision, and verified progress (Section~\ref{sec:method}).

  \section{SCOPER-bench}
  \label{sec:benchmark}

  We test System~2 information acquisition, goal decomposition, subtask selection, and plan revision from observations and outcomes. Instructions specify requirements, not action sequences; completion and situated question answering (QA) assess execution and scene/task understanding.

  \begin{figure}[t]
      \centering
      \includegraphics[width=1.0\linewidth,trim=0 4bp 0 7bp,clip]{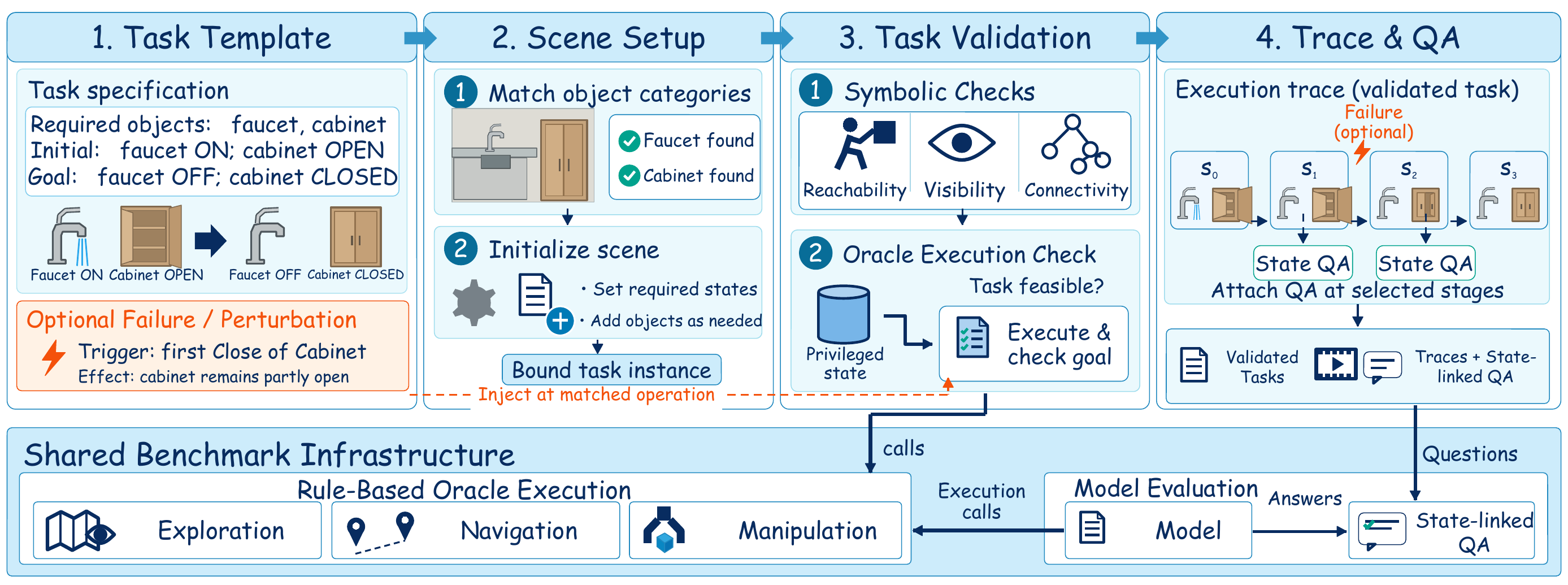}
      \vspace{-1mm}
      \caption{\small
          SCOPER-bench construction.
          Scene-grounded templates undergo static checks and privileged Oracle System~1 execution, including failures and recovery. Traces support situated QA construction; evaluation answers follow the evaluated episode's state. Average steps serve as a reference metric.
      }
      \label{fig:benchmark_construction}
      \vspace{-4mm}
  \end{figure}

  \subsection{Task Formulation and Evaluation Protocol}
  \label{sec:benchmark_design}
  \label{sec:benchmark_protocol}

  \paragraph{Task groups.}
  We organize the benchmark into four groups.
  \textbf{Routine execution} specifies concrete task requirements while leaving action selection and sequencing to the agent.
  \textbf{Observation-dependent} covers state-dependent execution, scene inspection, and responses to task updates.
  \textbf{Failure recovery} covers initially present anomalies, controlled execution failures, and mid-execution disturbances that require corrective actions or replanning.
  \textbf{Composite} combines multiple subtasks, requiring agents to maintain progress and reason over cross-step dependencies; these subtasks can draw from the other
  three groups.

  \paragraph{Observation-dependent decisions.}
  Tasks differ in their information needs before acting. For example, ``put the cup in the cabinet'' identifies the object and destination, whereas ``inspect the room,
  clean the dishware that needs cleaning, and put it away'' requires discovering which objects need attention. Observation remains necessary after acting: following a placement
  request, the agent must inspect whether the object reached its destination and assess further action needs.

    \paragraph{Task representation.}
  A task instance is represented as $\mathcal{T}=(x,\mathcal{R},\mathcal{C}_0,\mathcal{G},\mathcal{E},\mathcal{Q})$, where $x$ is the instruction, $\mathcal{R}$ specifies object-role constraints, $\mathcal{C}_0$ defines initial conditions, $\mathcal{G}$ contains goal predicates, $\mathcal{E}$
  configures optional execution events, and $\mathcal{Q}$ contains attached questions. This configuration supports construction and evaluation. Evaluated System~2 uses instructions, visual observations, and prescribed public tool information; construction-time role bindings, reference decompositions, and privileged simulator state
  provide no planning guidance.

  \paragraph{Agent inputs and evaluation protocol.}
  System~2 receives instructions, tool descriptions, permitted vocabulary, decision history, and tool-returned images. Exploration and navigation return eight uniformly sampled execution frames; state-changing actions return before-and-after frames. Direct execution retains up to 32 frames; harness context routing appears in Section~\ref{sec:method_decision} and Appendix~\ref{app:harness}. Execution ends on \texttt{stop} or step-budget exhaustion. Success requires both \texttt{stop} and the resulting environment state satisfying the template's goal predicates. Triggered questions use dedicated answer-only turns with the same context available to policy decisions. Observation settings, budgets, and scoring rules appear in Appendix~\ref{app:evaluation}; QA activation and answering details appear in Appendix~\ref{app:qa}.

  \subsection{Oracle System~1 and Controlled Execution}
  \label{sec:benchmark_execution}

  \paragraph{Unified skill infrastructure.}
  Oracle System~1 provides rule-based active observation, navigation, and manipulation skills shared by evaluated models and the construction-time teacher. Standardizing execution controls low-level variation without training a separate policy for each model. Complete skill definitions and object binding are given in Appendices~\ref{app:system1} and~\ref{app:binding}. Navigation uses A* search~\citep{hart1968astar} over traversability maps derived from scene collision geometry. Manipulation bypasses low-level physical execution, directly updating object poses, holding or containment relations, and task-relevant states with synchronized visual representations. Navigation, scanning, and target search collect multi-frame observations along paths and viewpoints; state-changing actions expose their visual outcomes for subsequent System~2 decisions.

    \paragraph{Controlled failures and perturbations.}
  Template-defined events alter System~1 behavior or outcomes at designated stages, requiring inspection of deviations, identification of unmet requirements, and recovery or replanning. This combines shared execution infrastructure and controllable imperfect outcomes; backend implementations and failure mechanisms appear in Appendices~\ref{app:system1} and~\ref{app:failures}.

  \subsection{Task Construction and Validation}
  \label{sec:benchmark_construction}

  \paragraph{Template-to-scene construction.}
  Figure~\ref{fig:benchmark_construction} summarizes the shared construction pipeline in BEHAVIOR~\citep{srivastava2021behavior} and ProcTHOR~\citep{deitke2022procthor,kolve2017ai2thor}, which diversify scene styles and layouts.
Using template constraints, the constructor matches scenes and binds roles to objects, checking required objects and fixtures against interaction and spatial constraints, using scene profiles in ProcTHOR and scene--template matching with asset materialization in BEHAVIOR. Initialization sets object placements, containment, agent pose, and task-relevant states with synchronized visual representations. For dishware inspection, a dirty item is placed where exploration is needed to discover it; the instruction specifies inspection and handling requirements without revealing the target. Rule-based subtask decomposition guides execution validation. Configuration details appear in Appendix~\ref{app:templates}.

  \paragraph{Static and execution validation.}
  Static checks test role and condition consistency, initial-state reconstruction, nontrivial goals, and reachability. A privileged teacher follows the reference decomposition through Oracle System~1, checking goal satisfaction and configured events; failure-injection instances must recover to completion with failures enabled. Traces and videos support inspection. We construct 870 executable tasks from 35 base templates and their compositions (Appendix~\ref{app:templates}).

  \paragraph{Situated QA attachment.}
  About 3,500 questions from eight QA templates probe object states, spatial relations, action outcomes, anomalies, new requirements, and task progress. Candidates derived from task semantics are screened against teacher traces for valid activation and resolvable answers. Evaluation triggers questions at configured actions, steps, or subtask milestones, with answers determined from the evaluated episode's corresponding state (Appendix~\ref{app:qa}).

\section{SCOPER-harness}
  \label{sec:method}

\begin{figure}[t]
      \centering
      \includegraphics[width=1.0\linewidth,trim=0 5bp 0 2bp,clip]{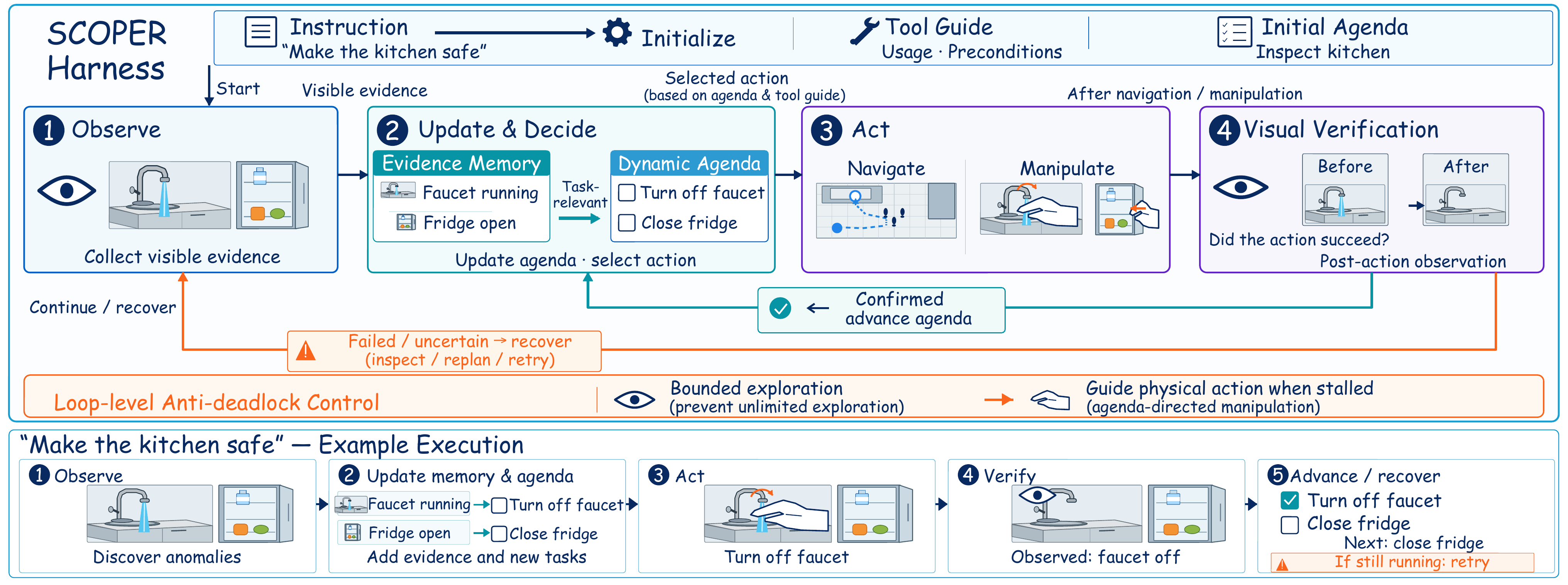}
      \vspace{-1mm}
      \caption{ \small
          SCOPER-harness. Evidence Memory: agenda revision, action selection; Progress Verification: execution outcomes for planning, recovery; anti-deadlock control: balancing exploration and task advancement.
      }
      \label{fig:method}
      \vspace{-4mm}
  \end{figure}

  Prior embodied systems use environmental feedback and explicit memory to support planning~\citep{huang2022innermonologue,wang2026thea,xu2026mimir,huang2022vlmaps,jatavallabhula2023conceptfusion,gu2023conceptgraphs}.
  Our evaluations expose two difficulties in closed-loop task execution: maintaining scene evidence to guide task decomposition and subsequent decisions, and detecting unsuccessful
  execution to support recovery. We introduce \textbf{SCOPER-harness}, a training-free harness that addresses these difficulties through \textbf{Evidence Memory} and \textbf{Progress
  Verification}. A dynamic agenda connects accumulated evidence to task requirements, while anti-deadlock control balances information gathering with task advancement. All model-based
  components share the same fixed-parameter backbone through role-specific prompts and use no additional privileged simulator information.
  Figure~\ref{fig:method} summarizes the architecture.

  \subsection{Evidence Memory and Adaptive Task Decomposition}
  \label{sec:method_state}

  \paragraph{Structured evidence memory.}
  Partial observations require retaining evidence without treating missing observations as state changes. We maintain $M_t=(F_t,I_t,O_t)$: facts store object properties, relations, and scene descriptions with evidence and validity annotations; incidents record execution discrepancies and handling status; obligations track outstanding observation or action needs linked to relevant objects and incidents.

  \paragraph{Instruction-derived agenda.}
  The agenda $A_t$ organizes required outcomes, ongoing constraints, and conditional requirements, while memory obligations track observation or handling needs. Initialization uses only the instruction, retaining unresolved conditions as inspection needs without assuming unseen states or fixing an action sequence. Items track status and supporting evidence.

  \paragraph{Observation-driven memory updates.}
  Before decisions, discovery extracts task-relevant objects and states from current and available execution-process images; reconciliation checks facts against new evidence. Confirmation refreshes evidence, contradiction invalidates facts, and newer scene descriptions can supersede older ones; invisibility alone does not negate a fact. A deterministic commit layer checks references, evidence links, and consistency before committing model updates.

  \paragraph{Evidence-grounded agenda revision.}
  Committed memory drives agenda revision: the model combines the instruction, existing items, and new evidence to activate conditional requirements, add necessary work within task
  scope, merge duplicates, and update or reopen affected items. Inspection items become supported when the required information is obtained; execution items require verified effects,
  while ongoing constraints remain monitored. Unresolved inspection needs guide observation actions toward relevant objects or conditions. For example, ``inspect the kitchen and turn
  off any running faucets'' initializes inspection and conditional shutoff requirements. Observing running water supports the corresponding state check and activates shutoff work, but
  does not establish that the faucet has been turned off.

  \paragraph{Context Routing and Action Selection.}
  \label{sec:method_decision}

  Deterministic routing selects committed memory within a bounded context, prioritizing relevant facts and unresolved needs while retaining evidence references and validity annotations. The policy receives the routed memory $\widetilde M_t$ together with the instruction $x$, recent images $V_t$, public action-request history $H_t$, agenda $A_t$, and control context
  $c_t$: $a_t=\pi_\theta(x,V_t,H_t,A_t,\widetilde M_t,c_t)$. The control context specifies available tools and arguments, the recovery phase, and action restrictions. The policy outputs one tool call with arguments. Request history records
  issued tool requests but provides no ground-truth success labels.

  \subsection{Progress Verification and Recovery}
  \label{sec:method_verification}

  \paragraph{Visual verification and recovery feedback.}
  Issuing an action does not establish progress. Navigation and manipulation therefore receive outcome verification; exploration supplies memory observations without separate verification. Given the tool request, expected state and visual effects, and immediate before-and-after images, the verifier returns success, failure, or ambiguity with visual evidence. Success commits supported effects and resolves satisfied needs; failure records discrepancies and recovery obligations; ambiguity leaves outcomes unconfirmed. Memory adopts the verifier's effect assessment without readjudicating the same images; later observations may revise it. In the faucet example, stopped flow supports completion; continued flow leaves shutoff unresolved and triggers recovery. The original request, evidence, and recovery state guide further observation, renewed approach, or retry. Verified subsequent effects update memory and the agenda, returning local recovery to task planning.

  \subsection{Anti-Deadlock Control}
  \label{sec:method_antideadlock}

  \paragraph{Balancing exploration, execution, and recovery budgets.}
  Evidence maintenance and verification can encourage excessive exploration or conservative checking. The controller tracks scans, searches, and navigation by room/target, and observations and retries per incident. At exploration limits, it restricts tools and requests reselection to promote approach and interaction. The agenda guides planning; recovery and anti-deadlock rules constrain execution. Reselection and recovery are budgeted. Exhaustion terminates execution or permits continuation with unresolved state, depending on the recovery branch. Temporary planning assumptions remain separate from verified facts, cannot support agenda completion, and remain revisable by later evidence. Appendix~\ref{app:harness} details memory and agenda records and states, routing order and capacities, recovery fields and transitions, counter updates and resets, branch rules, prompt responsibilities, and execution pseudocode.

\section{Experiments}
\begin{figure}
    \centering
    \includegraphics[width=1.0\linewidth,trim=0 3bp 0 5bp,clip]{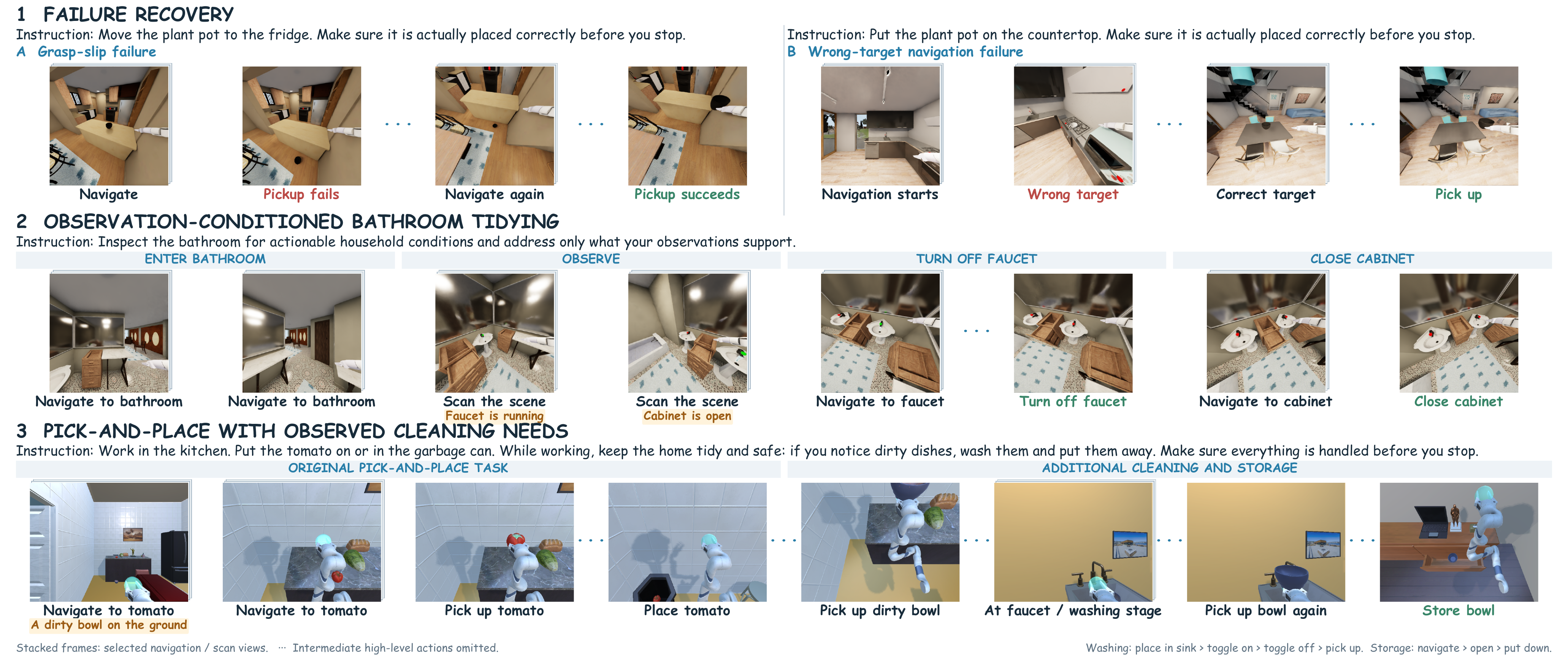}
    \vspace{-1mm}
    \caption{\small Example task traces: (1) recovery from controlled grasp and navigation failures; (2) observation-conditioned household actions identifying targets undisclosed in instructions; and (3) observation activating and fulfilling an instruction-specified cleaning requirement during placement.}
    \label{fig:trace_samples}
    \vspace{-4mm}
\end{figure}

\subsection{Experimental Setup}

  We evaluate GPT-6 Astra~\citep{openai2026astra}, Gemini-3.8-flash~\citep{google2026gemini38flash}, Embodied-R1.5~\citep{yuan2026embodiedr15}, RoboBrain2.5-8B~\citep{tan2026robobrain25},
  RynnBrain11-9B~\citep{damo2026rynnbrain11}, Molmo2-ER-4B~\citep{fang2026molmoact2}, Qwen3.5-9B~\citep{qwen2026qwen35}, and Qwen3.8-27B~\citep{qwen2026qwen38}
  in ProcTHOR~\citep{deitke2022procthor} and BEHAVIOR~\citep{srivastava2021behavior}. All eight models follow the shared Oracle System~1 interface and input protocol (Section~\ref{sec:benchmark_protocol}), using $512\times512$ RGB observations without privileged simulator state.
  For Qwen3.5-9B and Qwen3.8-27B, we compare direct execution against the same-backbone, training-free SCOPER-harness (Section~\ref{sec:method}). Given the limited gains and substantially higher cost of pairing GPT-6-Astra with the harness, we exclude this configuration from the full comparison.
  Across methods, atomic/two-subtask/three-subtask episodes allow
  24/36/48 high-level calls, including QA answers.
  Tool interfaces, observation processing, and model/harness configurations
  appear in Appendices~\ref{app:evaluation}, \ref{app:system1}, and~\ref{app:harness};
  model-call overhead is described in Appendix~\ref{app:call_overhead}.

  SR is the task success fraction; Prog./Cov.\ are task-averaged
  fractions of satisfied goal predicates/covered required actions.
  QA (T/A) denotes triggered/configured questions and
  correctly answered/triggered questions, respectively, counting
  triggered but unanswered questions as incorrect.
  Aggregates weight task metrics by task counts and QA metrics
  by their respective question denominators.
  Success conditions and detailed scoring rules appear in Appendix~\ref{app:evaluation}.
  Additional action-step and blind evaluations appear in Appendices~\ref{app:action_steps} and~\ref{app:blind_testing}. Figure~\ref{fig:trace_samples} illustrates why fixed instruction-derived action sequences can be insufficient. Further examples appear in Appendix~\ref{app:traces}.

\subsection{Main Results}

  \begin{table*}[t]
  \centering
  \caption{\small BEHAVIOR and ProcTHOR results by task group (\%). QA reports trigger rate / triggered answer accuracy; $\Delta$ denotes the change from the corresponding direct-execution baseline in percentage points. Bold indicates the best result among non-API models within each backend.}
  \label{tab:main-results}
  \label{tab:behavior-horizontal}
  \label{tab:procthor-horizontal}
  \begingroup
  \fontsize{7.2}{9.4}\selectfont
  \setlength{\tabcolsep}{2.0pt}
  \renewcommand{\arraystretch}{1.0}
  \resizebox{\textwidth}{!}{%
  \begin{tabular}{@{}l@{\hspace{7pt}}
  rrrr@{\hspace{7pt}}
  rrrr@{\hspace{7pt}}
  rrrr@{\hspace{7pt}}
  rrrr@{\hspace{7pt}}
  >{\columncolor{OverallBg}}r
  >{\columncolor{OverallBg}}r
  >{\columncolor{OverallBg}}r
  >{\columncolor{OverallBg}}r@{}}
  \toprule
Model & \multicolumn{4}{c}{Routine Execution} & \multicolumn{4}{c}{Observation-dependent} & \multicolumn{4}{c}{Failure Recovery} & \multicolumn{4}{c}{Composite} & \multicolumn{4}
  {>{\columncolor{OverallBg}}c}{Overall} \\
  \cmidrule(lr){2-5}\cmidrule(lr){6-9}\cmidrule(lr){10-13}\cmidrule(lr){14-17}\cmidrule(lr){18-21}
   & SR & Prog. & Cov. & QA (T / A) & SR & Prog. & Cov. & QA (T / A) & SR & Prog. & Cov. & QA (T / A) & SR & Prog. & Cov. & QA (T / A) & SR & Prog. & Cov. & QA (T / A) \\
  \midrule
  \multicolumn{21}{l}{\strut\bfseries BEHAVIOR} \\
\modelband{API model}
  GPT-6 Astra & 42.0 & 74.6 & 72.6 & 81.8 / 54.3 & 40.0 & 56.9 & 69.1 & 85.9 / 82.5 & 76.2 & 85.1 & 88.2 & 84.4 / 74.7 & 35.0 & 75.4 & 59.4 & 43.9 / 58.1 & 50.5 & 74.2 & 73.5 & 74.4 /
  68.9 \\
    Gemini-3.8-flash & 25.0 & 44.6 & 42.3 & 48.5 / 30.6 & 25.6 & 35.4 & 43.5 & 54.4 / 53.2 & 50.8 & 57.3 & 59.2 & 57.0 / 51.1 & 20.0 & 41.2 & 32.4 & 24.2 / 31.9 & 31.9 & 45.7 & 45.4 &
  46.5 / 44.1 \\

  \modelband{Embodied models}
  Embodied-R1.5 & 5.0 & 35.4 & 29.1 & 48.2 / \textbf{70.6} & 15.6 & 26.8 & 44.3 & 52.9 / 71.5 & 16.9 & 25.2 & 38.4 & 63.7 / 51.4 & 2.0 & 27.5 & 18.5 & 18.5 / 43.6 & 10.2 & 28.5 & 32.7
  & 46.9 / 60.1 \\
  RoboBrain2.5-8B & 1.0 & 27.0 & 16.2 & 28.0 / 51.8 & 11.1 & 13.7 & 29.3 & 29.3 / 72.4 & 10.8 & 18.9 & 27.7 & 43.3 / 44.6 & 1.0 & 24.3 & 12.4 & 11.1 / 60.6 & 6.2 & 21.0 & 21.7 & 29.0 /
  53.6 \\
  RynnBrain11-9B & 2.0 & 30.3 & 21.3 & 45.1 / 57.5 & 10.0 & 12.0 & 22.7 & 44.0 / 75.4 & 6.9 & 17.3 & 30.7 & 51.8 / 38.3 & 0.0 & 24.6 & 9.6 & 15.8 / 46.8 & 4.8 & 21.0 & 21.7 & 39.9 /
  52.8 \\
  Molmo2-ER-4B & 0.0 & 28.5 & 13.1 & 40.4 / 46.7 & 3.3 & 5.4 & 11.5 & 41.3 / 67.3 & 13.1 & 21.9 & 18.1 & 50.3 / 33.3 & 0.0 & 22.0 & 5.5 & 12.8 / 29.0 & 4.8 & 20.0 & 12.5 & 37.0 / 44.4 \\
  \modelband{Open-source models}
  Qwen3.5-9B & 5.0 & 35.7 & 25.6 & 47.1 / 60.7 & 15.6 & 19.8 & 33.0 & 46.7 / 66.1 & 15.4 & 24.9 & 35.8 & 49.5 / 43.8 & 0.0 & 25.0 & 13.2 & 15.4 / 56.5 & 9.3 & 26.4 & 27.4 & 40.2 / 55.1
  \\
  Qwen3.5-9B + harness & 28.0 & 58.9 & 54.0 & 68.5 / 56.7 & 30.0 & 41.0 & 53.6 & 68.5 / 74.3 & 50.8 & 60.9 & 67.1 & 71.0 / 65.0 & 22.0 & 56.1 & 41.7 & 33.0 / 59.6 & 34.1 & 55.0 & 55.0
  & 60.8 / 64.2 \\
  \textcolor{GainInk}{\quad $\Delta$ vs. base (pp)} & \gainpos{+23.0} & \gainpos{+23.3} & \gainpos{+28.5} & \gainpos{+21.4} / \gainneg{-4.1} & \gainpos{+14.4} & \gainpos{+21.2} &
  \gainpos{+20.6} & \gainpos{+21.8} / \gainpos{+8.2} & \gainpos{+35.4} & \gainpos{+36.0} & \gainpos{+31.3} & \gainpos{+21.5} / \gainpos{+21.2} & \gainpos{+22.0} & \gainpos{+31.1} &
  \gainpos{+28.5} & \gainpos{+17.6} / \gainpos{+3.1} & \gainpos{+24.8} & \gainpos{+28.6} & \gainpos{+27.6} & \gainpos{+20.6} / \gainpos{+9.1} \\[2pt]
  Qwen3.8-27B & 13.0 & 44.5 & 43.2 & 60.9 / 49.2 & 14.4 & 28.7 & 46.7 & 65.6 / 62.4 & 20.8 & 39.4 & 55.1 & 75.8 / 50.0 & 9.0 & 46.2 & 35.8 & 42.3 / \textbf{82.5} & 14.8 & 39.9 & 45.9 &
  62.1 / 57.9 \\
  Qwen3.8-27B + harness & \textbf{36.0} & \textbf{68.7} & \textbf{66.8} & \textbf{78.0} / 52.4 & \textbf{34.4} & \textbf{50.6} & \textbf{64.1} & \textbf{80.9} / \textbf{78.1} &
  \textbf{65.4} & \textbf{75.8} & \textbf{81.2} & \textbf{83.0} / \textbf{69.6} & \textbf{30.0} & \textbf{70.3} & \textbf{55.3} & \textbf{43.8} / 63.5 & \textbf{43.3} & \textbf{67.4} &
  \textbf{67.9} & \textbf{72.0} / \textbf{66.2} \\
  \textcolor{GainInk}{\quad $\Delta$ vs. base (pp)} & \gainpos{+23.0} & \gainpos{+24.2} & \gainpos{+23.5} & \gainpos{+17.1} / \gainpos{+3.2} & \gainpos{+20.0} & \gainpos{+21.9} &
  \gainpos{+17.4} & \gainpos{+15.2} / \gainpos{+15.7} & \gainpos{+44.6} & \gainpos{+36.5} & \gainpos{+26.0} & \gainpos{+7.2} / \gainpos{+19.6} & \gainpos{+21.0} & \gainpos{+24.1} &
  \gainpos{+19.5} & \gainpos{+1.6} / \gainneg{-19.1} & \gainpos{+28.6} & \gainpos{+27.5} & \gainpos{+22.0} & \gainpos{+9.9} / \gainpos{+8.4} \\[2pt]
  \midrule
\multicolumn{21}{l}{\strut\bfseries ProcTHOR} \\
\modelband{API model}
  GPT-6 Astra & 35.4 & 75.7 & 68.8 & 79.0 / 60.5 & 41.1 & 66.4 & 74.1 & 82.0 / 68.4 & 43.9 & 69.6 & 78.8 & 80.9 / 59.7 & 18.0 & 67.3 & 50.8 & 52.3 / 65.6 & 35.1 & 70.2 & 68.7 & 74.2 /
  62.7 \\
    Gemini-3.8-flash & 20.8 & 47.1 & 42.4 & 49.1 / 37.4 & 25.6 & 43.8 & 49.6 & 55.3 / 45.4 & 27.7 & 43.5 & 49.0 & 49.3 / 39.7 & 14.0 & 49.7 & 37.5 & 38.8 / 48.2 & 22.2 & 46.0 & 44.7 &
  48.1 / 41.8 \\

  \modelband{Embodied models}
  Embodied-R1.5 & 3.9 & 34.1 & 15.7 & 36.0 / 63.5 & 15.6 & 18.3 & 32.9 & 42.0 / 69.2 & 7.7 & 21.7 & 30.4 & 59.0 / \textbf{58.9} & 1.0 & 29.1 & 12.8 & 19.4 / 30.4 & 6.7 & 26.2 & 22.7 &
  40.2 / 59.1 \\
  RoboBrain2.5-8B & 1.5 & 26.9 & 11.8 & 21.5 / 46.3 & 10.0 & 14.4 & 21.7 & 25.9 / 60.6 & 9.2 & 22.7 & 24.8 & 40.7 / 44.2 & 3.0 & 28.7 & 16.2 & 19.4 / 55.4 & 5.8 & 23.6 & 18.5 & 27.5 /
  49.4 \\
  RynnBrain11-9B & 2.3 & 33.1 & 16.7 & 35.2 / 51.5 & 7.8 & 12.8 & 18.5 & 32.6 / 60.2 & 2.3 & 24.6 & 20.9 & 35.5 / 38.2 & 0.0 & 26.4 & 13.2 & 18.3 / 52.8 & 2.9 & 25.1 & 17.5 & 31.0 /
  49.0 \\
  Molmo2-ER-4B & 0.0 & 28.7 & 7.3 & 16.3 / 58.1 & 1.1 & 5.6 & 4.8 & 12.9 / 51.5 & 0.8 & 14.6 & 5.8 & 23.5 / 28.9 & 0.0 & 18.9 & 2.5 & 5.5 / 25.0 & 0.4 & 17.8 & 5.3 & 15.4 / 41.3 \\
  \modelband{Open-source models}
  Qwen3.5-9B & 0.8 & 34.1 & 25.8 & 44.1 / 55.4 & 18.9 & 25.6 & 40.9 & 50.6 / 62.8 & 8.5 & 29.2 & 43.6 & 64.5 / 44.9 & 2.0 & 35.8 & 23.5 & 31.1 / 75.6 & 6.9 & 31.4 & 33.4 & 48.5 / 55.7
  \\
  Qwen3.5-9B + harness & 20.8 & 59.1 & 51.9 & 65.6 / 58.2 & 32.2 & 49.6 & 60.8 & 69.5 / 66.1 & 30.0 & 53.5 & 64.7 & 74.3 / 54.5 & 12.0 & 54.6 & 39.7 & 43.7 / 69.8 & 23.8 & 54.6 & 54.7
  & 64.1 / 60.3 \\
  \textcolor{GainInk}{\quad $\Delta$ vs. base (pp)} & \gainpos{+20.0} & \gainpos{+25.0} & \gainpos{+26.1} & \gainpos{+21.5} / \gainpos{+2.8} & \gainpos{+13.3} & \gainpos{+24.0} &
  \gainpos{+19.9} & \gainpos{+18.9} / \gainpos{+3.3} & \gainpos{+21.5} & \gainpos{+24.2} & \gainpos{+21.1} & \gainpos{+9.8} / \gainpos{+9.5} & \gainpos{+10.0} & \gainpos{+18.8} &
  \gainpos{+16.2} & \gainpos{+12.5} / \gainneg{-5.7} & \gainpos{+16.9} & \gainpos{+23.2} & \gainpos{+21.2} & \gainpos{+15.6} / \gainpos{+4.7} \\[2pt]
  Qwen3.8-27B & 16.2 & 50.2 & 46.5 & 65.6 / \textbf{64.8} & 30.0 & 44.2 & 54.0 & 71.0 / \textbf{79.0} & 26.2 & 46.7 & 57.4 & 72.3 / 52.7 & 13.0 & 57.4 & 48.1 & \textbf{55.7} /
  \textbf{83.9} & 21.1 & 49.6 & 51.5 & 66.4 / \textbf{67.4} \\
  Qwen3.8-27B + harness & \textbf{33.1} & \textbf{71.1} & \textbf{64.6} & \textbf{76.5} / 61.7 & \textbf{37.8} & \textbf{61.5} & \textbf{70.0} & \textbf{79.8} / 70.4 & \textbf{39.2} &
  \textbf{64.7} & \textbf{74.4} & \textbf{79.1} / 58.5 & \textbf{17.0} & \textbf{65.3} & \textbf{50.3} & 53.0 / 69.6 & \textbf{32.2} & \textbf{66.1} & \textbf{65.4} & \textbf{72.7} /
  63.8 \\
  \textcolor{GainInk}{\quad $\Delta$ vs. base (pp)} & \gainpos{+16.9} & \gainpos{+21.0} & \gainpos{+18.1} & \gainpos{+10.9} / \gainneg{-3.1} & \gainpos{+7.8} & \gainpos{+17.3} &
  \gainpos{+16.0} & \gainpos{+8.8} / \gainneg{-8.6} & \gainpos{+13.1} & \gainpos{+18.0} & \gainpos{+17.0} & \gainpos{+6.8} / \gainpos{+5.8} & \gainpos{+4.0} & \gainpos{+8.0} &
  \gainpos{+2.2} & \gainneg{-2.7} / \gainneg{-14.2} & \gainpos{+11.1} & \gainpos{+16.5} & \gainpos{+13.8} & \gainpos{+6.3} / \gainneg{-3.6} \\[2pt]
  \bottomrule
\end{tabular}%
  }
  \endgroup
   \vspace{-3mm}
  \end{table*}

  \paragraph{General-purpose capability matters.}
  Table~\ref{tab:main-results} shows that GPT-6 Astra leads direct execution with overall SRs of 50.5\% on
  BEHAVIOR and 35.1\% on ProcTHOR, versus 14.8\% and 21.1\%
  for Qwen3.8-27B. Its necessary-action coverage is higher on both backends; only one of its 500 failed episodes ended in a parsing error.
  These results suggest strengths in reliable tool use, multi-step progression, observation-grounded action selection, outcome monitoring, and recovery. Embodied-specialized models show no
  consistent advantage, suggesting that gains on open-loop embodied
  evaluations need not transfer to closed-loop execution.
  Nevertheless, 58.6\% of Astra's failures exhausted the action budget,
  indicating limits in exploration efficiency and recovery planning.

  \paragraph{Harness gains across models and backends.}
  SCOPER-harness raises overall SR by {24.8/28.6 percentage
  points on BEHAVIOR and 16.9/11.1 on ProcTHOR} for
  Qwen3.5-9B/Qwen3.8-27B, respectively. SR, progress, and necessary-action
  coverage improve in every task group for both models on both
  backends.
  {Qwen3.5-9B with SCOPER-harness surpasses direct
  Qwen3.8-27B execution on both backends.}
  Additional inference calls organizing Evidence Memory, adaptive planning, and recovery thus improve fixed-model performance.

  \paragraph{Situated QA complements execution metrics.}
  GPT-6 Astra leads overall QA trigger rates on both backends while
  maintaining high accuracy on triggered questions, consistent with
  its task performance. SCOPER-harness increases trigger rates in
  all four backbone--backend combinations and triggered accuracy in
  three. {Accuracy must be read alongside trigger rate}:
  execution paths activate different question sets, precluding an
  identical-question interpretation.

  \paragraph{Failure patterns motivate evidence and progress control.}
  Across the six open-source models' 4,813 failed normal-image
  episodes, {47.7\% stopped prematurely and 30.7\% exhausted
  the action budget}, reflecting difficulties in completion
  assessment, state tracking, and sustained progress.
  At least three consecutive identical tool calls occurred in
  40.2\%, suggesting difficulty adapting unsuccessful strategies;
  parsing failures accounted for 12.8\%, and reference-binding
  issues appeared in 17.3\%.
  This motivates evidence-based completion checks, state tracking, repetition-aware recovery, and clearer tool/reference conventions. These indicators overlap rather than partition failure causes.

\subsection{Harness Component Ablations}

  \begin{table*}[t]
  \centering
  \vspace{-2mm}
  \caption{\small Qwen3.8-27B harness component ablation results on ProcTHOR and BEHAVIOR}
  \label{tab:ablation-horizontal-no-bold}
  \begingroup
  \fontsize{7.2}{8}\selectfont       
  \renewcommand{\arraystretch}{0.95} 
  \setlength{\tabcolsep}{3pt}
  \setlength{\aboverulesep}{1pt}
  \setlength{\belowrulesep}{1pt}

  \resizebox{\textwidth}{!}{%
  \begin{tabular}{@{}l@{\hspace{8pt}}
  rrr@{\hspace{9pt}}
  rrr@{\hspace{9pt}}
  rrr@{\hspace{9pt}}
  rrr@{\hspace{9pt}}
  >{\columncolor{OverallBg}}r
  >{\columncolor{OverallBg}}r
  >{\columncolor{OverallBg}}r@{}}
  \toprule
  Variant & \multicolumn{3}{c}{Routine Execution} & \multicolumn{3}{c}{Observation-dependent} & \multicolumn{3}{c}{Failure Recovery} & \multicolumn{3}{c}{Composite} & \multicolumn{3}
  {>{\columncolor{OverallBg}}c}{Overall} \\
  \cmidrule(lr){2-4}\cmidrule(lr){5-7}\cmidrule(lr){8-10}\cmidrule(lr){11-13}\cmidrule(lr){14-16}
   & SR & Prog. & Cov. & SR & Prog. & Cov. & SR & Prog. & Cov. & SR & Prog. & Cov. & SR & Prog. & Cov. \\
  \midrule
  \rowcolor{AblationBand}\multicolumn{16}{l}{\strut\color{AblationInk}\itshape ProcTHOR (450 episodes per variant)} \\
  Full harness & 33.1 & 71.1 & 64.6 & 37.8 & 61.5 & 70.0 & 39.2 & 64.7 & 74.4 & 17.0 & 65.3 & 50.3 & 32.2 & 66.1 & 65.4 \\
  w/o Anti-deadlock & 21.5 & 45.5 & 40.5 & 24.4 & 41.7 & 47.5 & 23.9 & 42.5 & 48.0 & 11.0 & 40.7 & 30.4 & 20.4 & 42.8 & 41.8 \\
  \textcolor{GainInk}{\quad $\Delta$ vs. full (pp)} & \textcolor{DropInk}{-11.5} & \textcolor{DropInk}{-25.7} & \textcolor{DropInk}{-24.2} & \textcolor{DropInk}{-13.3} &
  \textcolor{DropInk}{-19.8} & \textcolor{DropInk}{-22.5} & \textcolor{DropInk}{-15.4} & \textcolor{DropInk}{-22.2} & \textcolor{DropInk}{-26.4} & \textcolor{DropInk}{-6.0} &
  \textcolor{DropInk}{-24.7} & \textcolor{DropInk}{-19.9} & \textcolor{DropInk}{-11.8} & \textcolor{DropInk}{-23.3} & \textcolor{DropInk}{-23.5} \\
  w/o Recovery & 34.6 & 74.0 & 67.1 & 42.2 & 61.1 & 70.8 & 27.7 & 50.7 & 62.4 & 18.0 & 60.3 & 49.3 & 30.4 & 61.7 & 62.5 \\
  \textcolor{GainInk}{\quad $\Delta$ vs. full (pp)} & \textcolor{GainInk}{+1.5} & \textcolor{GainInk}{+2.9} & \textcolor{GainInk}{+2.5} & \textcolor{GainInk}{+4.4} &
  \textcolor{DropInk}{-0.3} & \textcolor{GainInk}{+0.8} & \textcolor{DropInk}{-11.5} & \textcolor{DropInk}{-14.0} & \textcolor{DropInk}{-12.0} & \textcolor{GainInk}{+1.0} &
  \textcolor{DropInk}{-5.1} & \textcolor{DropInk}{-1.1} & \textcolor{DropInk}{-1.8} & \textcolor{DropInk}{-4.4} & \textcolor{DropInk}{-2.8} \\
  w/o Memory & 31.5 & 70.0 & 63.7 & 33.3 & 47.3 & 56.1 & 40.0 & 65.2 & 75.7 & 16.0 & 62.7 & 50.1 & 30.9 & 62.5 & 62.6 \\
  \textcolor{GainInk}{\quad $\Delta$ vs. full (pp)} & \textcolor{DropInk}{-1.5} & \textcolor{DropInk}{-1.1} & \textcolor{DropInk}{-0.9} & \textcolor{DropInk}{-4.5} &
  \textcolor{DropInk}{-14.2} & \textcolor{DropInk}{-14.0} & \textcolor{GainInk}{+0.8} & \textcolor{GainInk}{+0.5} & \textcolor{GainInk}{+1.3} & \textcolor{DropInk}{-1.0} &
  \textcolor{DropInk}{-2.7} & \textcolor{DropInk}{-0.3} & \textcolor{DropInk}{-1.3} & \textcolor{DropInk}{-3.6} & \textcolor{DropInk}{-2.7} \\
  \addlinespace[1pt]
  \rowcolor{AblationBand}\multicolumn{16}{l}{\strut\color{AblationInk}\itshape BEHAVIOR (420 episodes per variant)} \\
  Full harness & 36.0 & 68.7 & 66.8 & 34.4 & 50.6 & 64.1 & 65.4 & 75.8 & 81.2 & 30.0 & 70.3 & 55.3 & 43.3 & 67.4 & 67.9 \\
  w/o Anti-deadlock & 22.0 & 41.1 & 39.4 & 20.0 & 31.2 & 38.1 & 38.5 & 47.5 & 50.1 & 18.0 & 43.0 & 32.4 & 25.7 & 41.4 & 40.8 \\
  \textcolor{GainInk}{\quad $\Delta$ vs. full (pp)} & \textcolor{DropInk}{-14.0} & \textcolor{DropInk}{-27.5} & \textcolor{DropInk}{-27.4} & \textcolor{DropInk}{-14.4} &
  \textcolor{DropInk}{-19.4} & \textcolor{DropInk}{-26.0} & \textcolor{DropInk}{-26.9} & \textcolor{DropInk}{-28.4} & \textcolor{DropInk}{-31.1} & \textcolor{DropInk}{-12.0} &
  \textcolor{DropInk}{-27.4} & \textcolor{DropInk}{-22.9} & \textcolor{DropInk}{-17.6} & \textcolor{DropInk}{-26.0} & \textcolor{DropInk}{-27.2} \\
  w/o Recovery & 38.0 & 71.3 & 67.5 & 34.4 & 51.0 & 63.7 & 21.5 & 40.9 & 58.1 & 29.0 & 68.3 & 53.6 & 30.0 & 56.8 & 60.5 \\
  \textcolor{GainInk}{\quad $\Delta$ vs. full (pp)} & \textcolor{GainInk}{+2.0} & \textcolor{GainInk}{+2.6} & \textcolor{GainInk}{+0.7} & \textcolor{GainInk}{+0.0} &
  \textcolor{GainInk}{+0.4} & \textcolor{DropInk}{-0.4} & \textcolor{DropInk}{-43.8} & \textcolor{DropInk}{-35.0} & \textcolor{DropInk}{-23.1} & \textcolor{DropInk}{-1.0} &
  \textcolor{DropInk}{-2.1} & \textcolor{DropInk}{-1.6} & \textcolor{DropInk}{-13.3} & \textcolor{DropInk}{-10.6} & \textcolor{DropInk}{-7.5} \\
  w/o Memory & 36.0 & 67.7 & 65.4 & 17.8 & 33.0 & 49.8 & 64.6 & 74.8 & 82.0 & 23.0 & 63.0 & 49.4 & 37.9 & 61.3 & 63.4 \\
  \textcolor{GainInk}{\quad $\Delta$ vs. full (pp)} & \textcolor{GainInk}{+0.0} & \textcolor{DropInk}{-1.0} & \textcolor{DropInk}{-1.4} & \textcolor{DropInk}{-16.7} &
  \textcolor{DropInk}{-17.6} & \textcolor{DropInk}{-14.3} & \textcolor{DropInk}{-0.8} & \textcolor{DropInk}{-1.1} & \textcolor{GainInk}{+0.8} & \textcolor{DropInk}{-7.0} &
  \textcolor{DropInk}{-7.3} & \textcolor{DropInk}{-5.9} & \textcolor{DropInk}{-5.5} & \textcolor{DropInk}{-6.1} & \textcolor{DropInk}{-4.5} \\
  \bottomrule
  \end{tabular}%
  }
  \endgroup
  \vspace{-5mm}
  \end{table*}

 Table~\ref{tab:ablation-horizontal-no-bold} separately ablates anti-deadlock control, explicit recovery, and Evidence Memory using Qwen3.8-27B on both backends. Each \textit{w/o} disables only the named component; Recovery denotes Progress Verification and its associated recovery mechanism.

  \paragraph{Anti-deadlock control sustains progress.}
  Removing this component lowers overall SR by 11.8/17.6 percentage
  points on ProcTHOR/BEHAVIOR, reducing progress and necessary-action coverage in every task group. Trajectories show prolonged exploration and repeated confirmation without physical operations, also seen in direct execution and potentially reinforced by harness evidence and verification. Exploration limits and action commitment prevent budget exhaustion.

  \paragraph{Explicit recovery supports anomaly handling.}
  Without recovery, failure-recovery SR drops from 39.2\% to 27.7\%
  on ProcTHOR and 65.4\% to 21.5\% on BEHAVIOR, while routine
  execution and observation-dependent performance remain largely
  stable or improve slightly. This selective decline supports verification and recovery for identifying unmet action effects and organizing corrective actions, with benefits concentrated in anomaly handling.

\paragraph{Memory supports observation-dependent decisions.}
  Removing Evidence Memory lowers observation-dependent SR from
  37.8\% to 33.3\% on ProcTHOR and from 34.4\% to 17.8\% on
  BEHAVIOR. Composite SR also falls from 30.0\% to 23.0\% on
  BEHAVIOR, with a smaller change on ProcTHOR. This supports retaining evidence for subsequent observation-dependent decisions.

\subsection{Real-Observation Evaluation}

  \begin{table*}[t]
  \centering
  \caption{\small Evaluation of normal execution, observation, and failure recovery on real execution data.}
  \label{tab:real-observation-results}
  \begingroup
  \fontsize{7.2}{8}\selectfont
  \renewcommand{\arraystretch}{0.95}
  \setlength{\tabcolsep}{4pt}
  \setlength{\aboverulesep}{1pt}
  \setlength{\belowrulesep}{1pt}
  \resizebox{\textwidth}{!}{%
  \begin{tabular}{@{}l@{\hspace{8pt}}
  rr@{\hspace{9pt}}
  rr@{\hspace{9pt}}
  rr@{\hspace{10pt}}
  rr@{\hspace{12pt}}
  >{\columncolor{OverallBg}}r
  >{\columncolor{OverallBg}}r@{}}
  \toprule
  Model & \multicolumn{2}{c}{Routine Execution} & \multicolumn{2}{c}{Observation-dependent} & \multicolumn{2}{c}{Failure Recovery} & \multicolumn{2}{c}{Critical Decisions} &
  \multicolumn{2}{>{\columncolor{OverallBg}}c}{Overall} \\
  \cmidrule(lr){2-3}\cmidrule(lr){4-5}\cmidrule(lr){6-7}\cmidrule(lr){8-9}\cmidrule(lr){10-11}
   & Tool & Call & Tool & Call & Tool & Call & Obs. Call & Repair Call & Tool & Call \\
  \midrule
    \rowcolor{AgiBand}\multicolumn{11}{l}{\strut\color{AgiInk}\itshape API model} \\
  GPT-6 Astra & 86.8 & 79.6 & 78.2 & 69.4 & 73.8 & 63.5 & 64.4 & 56.7 & 79.5 & 70.8 \\
  \rowcolor{AgiBand}\multicolumn{11}{l}{\strut\color{AgiInk}\itshape Embodied models} \\
  Embodied-R1.5 & 55.4 & 45.4 & 44.3 & 31.7 & 40.4 & 28.3 & 22.2 & 20.0 & 46.7 & 35.1 \\
  RoboBrain2.5-8B & 51.9 & 40.8 & 38.1 & 26.0 & 35.4 & 22.7 & 16.7 & 13.3 & 41.8 & 29.8 \\
  RynnBrain11-9B & 52.2 & 42.0 & 41.0 & 28.1 & 35.7 & 23.5 & 20.0 & 15.0 & 42.9 & 31.1 \\
  Molmo2-ER-4B & 49.6 & 36.4 & 37.9 & 25.4 & 34.5 & 21.3 & 15.6 & 11.7 & 40.7 & 27.7 \\
  \rowcolor{AgiBand}\multicolumn{11}{l}{\strut\color{AgiInk}\itshape Open-source models} \\
  Qwen3.5-9B & 55.8 & 45.0 & 42.8 & 30.3 & 38.2 & 26.4 & 20.0 & 13.3 & 45.6 & 33.9 \\
  Qwen3.5-9B + harness & 62.2 & 52.9 & 59.3 & 48.9 & 55.4 & 44.3 & 46.7 & 41.7 & 58.9 & 48.6 \\
  \textcolor{AgiGain}{\quad $\Delta$ vs. base (pp)} & \textcolor{AgiGain}{+6.4} & \textcolor{AgiGain}{+7.9} & \textcolor{AgiGain}{+16.5} & \textcolor{AgiGain}{+18.6} &
  \textcolor{AgiGain}{+17.2} & \textcolor{AgiGain}{+17.9} & \textcolor{AgiGain}{+26.7} & \textcolor{AgiGain}{+28.4} & \textcolor{AgiGain}{+13.3} & \textcolor{AgiGain}{+14.7} \\
  Qwen3.8-27B & 63.8 & 54.1 & 52.6 & 41.8 & 49.2 & 37.0 & 28.9 & 23.3 & 55.2 & 44.3 \\
  Qwen3.8-27B + harness & 73.8 & 64.8 & 68.8 & 60.0 & 65.0 & 54.4 & 56.7 & 51.7 & 69.1 & 59.6 \\
  \textcolor{AgiGain}{\quad $\Delta$ vs. base (pp)} & \textcolor{AgiGain}{+10.0} & \textcolor{AgiGain}{+10.7} & \textcolor{AgiGain}{+16.2} & \textcolor{AgiGain}{+18.2} &
  \textcolor{AgiGain}{+15.8} & \textcolor{AgiGain}{+17.4} & \textcolor{AgiGain}{+27.8} & \textcolor{AgiGain}{+28.4} & \textcolor{AgiGain}{+13.9} & \textcolor{AgiGain}{+15.3} \\
  \bottomrule
  \end{tabular}%
  }
  \endgroup
  \vspace{-5mm}
  \end{table*}

  We test transfer of simulation findings through offline next-tool-call evaluation on 200 distinct AgiBotWorld2026~\citep{agibot2026world}
  demonstration episodes: 80 normal-execution, 60 observation-dependent,
  and 60 involving anomalous or incomplete actions and repair.
  At each node, direct execution and SCOPER-harness predict the next tool call from the instruction, current semantic action segment's head-camera images, and preceding reference actions; predictions are not propagated as executed actions. We report tool accuracy and full-call accuracy
  (correct tool and all required arguments), separately evaluating
  observation-critical and repair-decision nodes.
  This fixed-history evaluation tests real-visual decisions, but not active observation acquisition, consequences of agents' own errors, or sustained multi-step recovery. It complements closed-loop simulation testing task completion through continued interaction.
  Construction and scoring appear in Appendix~\ref{app:agibot}.
  Consistent with simulation (Table~\ref{tab:real-observation-results}), GPT-6 Astra leads overall full-call accuracy at 70.8\%,
  while embodied-specialized models show no consistent advantage.
  SCOPER-harness raises full-call accuracy from 33.9\% to 48.6\%
  for Qwen3.5-9B and 44.3\% to 59.6\% for Qwen3.8-27B.
  Observation-critical call accuracy improves by 26.7/27.8 percentage
  points, respectively, and repair-call accuracy by 28.4 points
  for both, benefiting real visual decisions, especially critical nodes.

\section{Conclusion}
We introduced the SCOPER suite for studying embodied System~2 through closed-loop,
observation-grounded planning and recovery. SCOPER-bench combines high-level goals,
execution-time observations, and controlled failures in 870 tasks across two
simulators. SCOPER-harness connects accumulated scene evidence and verified action
outcomes to an evolving task agenda, improving performance across the evaluated
models and on real-observation decision tests. These findings highlight the
importance of using observations to determine required work and revise decisions
throughout execution. Future work will extend the current skill interface toward
richer, more continuous action spaces and broader action coverage, and adapt the
benchmark and harness to more complex scenes with greater diversity, dynamics,
and interaction demands.

\subsection*{AI use statement}
We used Codex and the models GPT-5.6-Sol, GPT-6-Astra, and Gemini-3.8-flash
for AI-assisted research discussions and code implementation. AI assistance also
supported manuscript drafting and revision, literature search and reference
checking, and figure-script editing. The benchmark additionally uses AI models
as evaluated agents, as described in the experimental setup. The authors directed these uses, reviewed and revised the assisted content, and take responsibility for the final paper and its scientific claims.

\subsection*{Reproducibility statement}
Section~\ref{sec:benchmark} describes benchmark construction and evaluation, and
Section~\ref{sec:method} presents the harness. The appendices document task
templates and dataset statistics (Appendix~\ref{app:templates}), execution
infrastructure and controlled failures (Appendices~\ref{app:system1}
and~\ref{app:failures}), evaluation protocols and situated QA
(Appendices~\ref{app:evaluation} and~\ref{app:qa}), and harness state definitions
and pseudocode (Appendix~\ref{app:harness}). Real-observation evaluation details
are provided in Appendix~\ref{app:agibot}. Additional supporting files will be
provided as supplementary material.

\bibliographystyle{iclr2027_conference}
\bibliography{iclr2027_conference}

\clearpage
\appendix

\section*{Appendix Guide}
\phantomsection
\pdfbookmark[0]{Appendix Guide}{appendix-guide}
The links below lead to each appendix section.
\begin{itemize}
  \item \hyperref[app:evaluation]{Appendix~\ref*{app:evaluation}: Evaluation Protocol}
  \item \hyperref[app:action_steps]{Appendix~\ref*{app:action_steps}: High-Level Action Counts}
  \item \hyperref[app:blind_testing]{Appendix~\ref*{app:blind_testing}: Blind Evaluation}
  \item \hyperref[app:agibot]{Appendix~\ref*{app:agibot}: Real-Robot Next-Tool-Call Evaluation}
  \item \hyperref[app:system1]{Appendix~\ref*{app:system1}: Oracle System 1 and Teacher Policy}
  \item \hyperref[app:failures]{Appendix~\ref*{app:failures}: Controlled Failure Injection and Scene Disturbances}
  \item \hyperref[app:templates]{Appendix~\ref*{app:templates}: Task Templates and Construction}
  \item \hyperref[app:binding]{Appendix~\ref*{app:binding}: Binding Procedure}
  \item \hyperref[app:qa]{Appendix~\ref*{app:qa}: Situated QA}
  \item \hyperref[app:harness]{Appendix~\ref*{app:harness}: Harness Details}
  \item \hyperref[app:traces]{Appendix~\ref*{app:traces}: Execution Trace Visualization}
  \item \hyperref[app:call_overhead]{Appendix~\ref*{app:call_overhead}: Harness Model-Call Overhead}
\end{itemize}

\section{Evaluation Protocol}
\label{app:evaluation}

Evaluation compares closed-loop execution through a shared skill interface and task-level scorer. This section specifies episode interaction, available information, and aggregation; detailed QA definitions appear in Appendix~\ref{app:qa}.

\subsection{Interaction, Observations, and Information Access}

Each episode loads a task, initializes the scene and event schedulers, and produces initial RGB images. The model emits one skill request from its current decision context; Binding resolves public references, System~1 executes admitted requests, and the interface supplies images and request records for the next decision. A separate evaluator checks goals and events.

All evaluated methods receive public instructions, tool definitions,
RGB images, and permitted request records. Simulator inventories and
identifiers, role mappings, ground-truth states and poses, distances,
goal progress, injection labels, teacher trajectories, and reference
decompositions remain internal. Public vocabulary defines valid references
rather than scene contents. Harness memory, agenda, and visual assessments
derive from public inputs. Execution-admission signals are used by the
deterministic controller only for gating and budget accounting; executor
success flags and diagnostic records are excluded from model inputs.

Navigation, scanning, and target search each produce up to eight
process-observation frames, retaining the current view; ordinary
manipulation supplies before-and-after RGB images. These observations
enter a visual history maintained across actions. Direct evaluation
models receive the latest available frames, up to 32, together with
public action-request history at each decision. The harness uniformly
samples up to eight frames from navigation, scanning, and search
for observation maintenance; its policy uses up to the latest four
frames, public request history, and routed memory. Since memory retains
evidence from earlier observations, the policy uses fewer recent frames
than direct evaluation. When fewer frames
are available, all available frames within the respective limit are
used. Instructions, public vocabulary, and tool definitions remain
accessible throughout execution.

The harness uses Qwen3.5-9B and Qwen3.8-27B; policy, verification, memory, and agenda calls within an experiment use the same fixed-parameter base model. Additional model calls contribute computational cost rather than additional environmental observations; acquiring a different viewpoint or position requires a public tool request and consumes the corresponding interaction budget.

Direct-model evaluation uses an input image resolution of
$512\times512$; GPT-6 uses the \texttt{max} reasoning-effort setting.

\subsection{Evaluation Prompts and Per-Decision Context}
\label{app:prompts}

The fixed system prompt assigns the model the role of an embodied agent completing the task with the listed skills. It supplies the instruction, tool names and argument definitions, public reference rules, and general action prerequisites. Instructions specify objectives, scope, and conditional requirements. Shared guidance covers approaching objects before interaction, assessing effects from images, observing again when needed, and stopping after all task clauses are satisfied; the model chooses the task-specific action sequence.

Each user turn contains the current step, remaining tool budget, the previous requested skill and public arguments, chronologically ordered action-request history, and the current image input. The request-history limit follows the episode's high-level budget, capped at 48 entries, allowing preceding requests to be retained within the prescribed 24-, 36-, or 48-step episodes. Each entry states what was requested without an action-success label. Full user--assistant conversations are not replayed by default, and old images are not repeatedly embedded in the history text; previously generated reasoning is not retained as permanent environmental evidence.

The model outputs one tool name and its arguments, optionally accompanied by a short rationale. The interface parses the request before Binding and execution. The next-turn summary repeats the requested call and states that it has finished; the model assesses its outcome from images.

The following layout distinguishes fixed interaction rules from
turn-specific content. Images are attached chronologically as image
inputs; dynamic content comes from the public task and interaction
records. New task messages enter subsequent task context, whereas
QA messages use a dedicated answer format.
\begin{quote}\small
\textbf{System:} Complete the task using the available tools.
Task: [instruction and public task updates].
Tools: [names and argument definitions].
Follow the public reference rules and interaction prerequisites.
Use images to assess outcomes; a requested action does not establish
success. Return one tool call with arguments.\par
\textbf{Decision turn:} Step: [current step].
Remaining budget: [remaining calls].
Previous requests: [public action history].
Last request: [tool and arguments; call finished].
Visual input: [ordered frames]. Choose the next call.\par
\textbf{QA turn:} Question: [identifier, text, and choices].
Use the permitted current context and return the question identifier
and selected answer value.
\end{quote}

When a situated question is posted, a separate answer call supplies its public identifier, text, choices, and the visual and execution context permitted at that time; the model responds with the question identifier and selected answer value. Newly introduced task requirements are supplied as public natural-language messages and distinguished from questions requesting only an answer. After stopping, the prompt specifies an answer-only phase without further physical actions; goal checking and answer correctness remain evaluator-side operations.

\subsection{Budgets and Termination}

Atomic, two-subtask, and three-subtask episodes use high-level call budgets of 24, 36, and 48, respectively, shared across compared methods. Observation, manipulation, answering, and stopping are counted under a common protocol; appended QA answers also consume this budget. Internal verification, memory maintenance, and action reselection calls are recorded separately, enabling separate comparisons of environment-interaction efficiency and model-call cost.

Under the standard stop-and-QA finishing protocol, \texttt{stop} ends physical execution and freezes its task score. Pending questions are handled in an answer-only finishing phase using the remaining budget, without resuming physical actions. Physical completion and subsequent answers can therefore be scored separately.

Task completion is determined from environment state at scoring time and applicable process conditions, rather than a model's completion claim or earlier goal attainment. Let $\mathcal G_i$ denote an episode's goal predicates. We set $s_i=1$ exactly when all predicates hold and the task's configured stopping requirement is satisfied. A stop call freezes the goal score; if the budget expires without stopping, the last available state is evaluated and any unmet stopping requirement remains unsatisfied. Budget exhaustion is additionally recorded as a runner failure, so runner status and physical task success are reported separately.

For composite tasks, handoff records local child completion, whereas overall scoring uses the parent task's required goal set; handoff records do not replace the overall goal check. Event records separately establish whether an intervention was triggered and applied. An untriggered intervention provides no evidence of successful recovery. Goal scores, intervention application, and QA scores remain distinct.

\subsection{Metrics and Aggregation}
\label{app:metrics}

We compute task metrics over a common set $\mathcal D$ of scored episodes, including both successful tasks and agent failures during valid evaluation. Infrastructure errors and unfinished runs are recorded separately rather than counted as task successes or failures. Task-group and template results use the corresponding subsets; proportions below are reported as percentages.

\paragraph{Success Rate (SR).}
SR measures the fraction of episodes satisfying their final success conditions:
\begin{equation}
\mathrm{SR}=\frac{1}{|\mathcal D|}\sum_{i\in\mathcal D}s_i,
\qquad s_i\in\{0,1\}.
\end{equation}
Here, $s_i$ is the evaluator's \texttt{task\_success} indicator, requiring the scored state to satisfy the task goals, including applicable process conditions, and stopping conditions. It is neither the model's completion claim nor merely a successful runner termination.

\paragraph{Task Progress (Prog.).}
Progress measures partial completion. Let $\mathcal G_i$ be the goal predicates of episode $i$ and $g_{ij}\in\{0,1\}$ indicate whether predicate $j$ holds at scoring time. We first compute the satisfied fraction within each episode and then average over episodes:
\begin{equation}
\mathrm{Prog.}=\frac{1}{|\mathcal D|}\sum_{i\in\mathcal D}
\frac{\sum_{j\in\mathcal G_i}g_{ij}}{|\mathcal G_i|}.
\end{equation}
Thus, tasks with more predicates do not automatically receive greater weight.

Goal items can be environmental predicates or evaluator-recorded
process conditions. Container inspection uses two goal items: a valid
interior inspection and the condition-dependent final openness.
Logical branches of that condition are not separate scoring items.

\paragraph{Necessary Action Coverage (Cov.).}
Coverage measures how many required actions from the teacher policy's reference execution are covered. For each task, a validated teacher trajectory is compiled into a deduplicated reference action set with dependency constraints $\mathcal R_i$, which serves as the ground-truth reference for this metric. Matching uses the normalized tool, the resolved execution target, and the manipulated object for placement or release. Same-object causal dependencies and composite-stage constraints are retained, while unrelated actions may be reordered. Only execution-confirmed matches count; matching request names or repeated attempts alone do not increase coverage. References remain private to the evaluator.

Let $\mathcal C_i\subseteq\mathcal R_i$ contain the matched reference actions. Coverage is normalized within each task before averaging:
\begin{equation}
\mathrm{Cov.}=\frac{1}{|\mathcal D|}\sum_{i\in\mathcal D}
\frac{|\mathcal C_i|}{|\mathcal R_i|}.
\end{equation}
This measures execution coverage relative to the teacher reference, rather than shortest-path efficiency or final goal correctness, and is reported separately from SR and progress.

QA aggregation includes only questions with valid evaluator answer
keys. Questions whose keys cannot be resolved are recorded separately
and excluded from the corresponding configured and posted counts,
as well as correct-answer counts. Missing, incorrect, or unparseable
model responses are not answer-key resolution failures and remain
incorrect responses to valid posted questions.

\paragraph{QA Trigger Rate and Triggered QA Accuracy (T / A).}
These metrics distinguish reaching question checkpoints from answering them correctly. Let $q_i$, $p_i$, and $c_i$ denote the numbers of valid configured, valid posted, and correctly answered situated questions in episode $i$. The main tables report
\begin{equation}
T=\frac{\sum_{i\in\mathcal D}p_i}{\sum_{i\in\mathcal D}q_i},
\qquad
A=\frac{\sum_{i\in\mathcal D}c_i}{\sum_{i\in\mathcal D}p_i}.
\end{equation}
Questions eligible for activation but still awaiting delivery in the queue do not count as posted.

Triggered but unanswered questions count as incorrect in $A$; untriggered questions are excluded from its denominator but reduce $T$. A metric with a zero denominator is reported as unavailable. Different execution paths can expose different question sets, so accuracy is interpreted alongside trigger rate. These metrics concern situated QA.

\paragraph{Average Steps.}
Let $L_i$ be the number of high-level calls counted by the interaction protocol in episode $i$. We report
\begin{equation}
\mathrm{AvgSteps}=\frac{1}{|\mathcal D|}\sum_{i\in\mathcal D}L_i.
\end{equation}
Both successful and failed episodes are included. Observation, navigation, manipulation, answering, and stopping follow the shared counting protocol; low-level simulator actions and internal harness model calls are excluded. Atomic, two-subtask, and three-subtask budgets are 24, 36, and 48 calls, respectively. Fewer steps can result from early failure, so this metric is not designated lower-is-better.

\paragraph{Aggregation.}
SR, progress, coverage, and average steps are weighted by episode counts; QA metrics are weighted by their respective question denominators. We do not directly average template or category percentages or assign equal weight to per-episode QA accuracies. Overall results are computed from underlying counts or equivalent denominator-weighted aggregates.

\section{High-Level Action Counts}
\label{app:action_steps}
\begin{table}[t]
\centering
\small
\caption{Average high-level calls over all scored episodes, including agent failures. Lower values do not necessarily indicate better efficiency.}
\label{tab:average_steps}
\begin{tabular}{lrr}
\toprule
Model & ProcTHOR & BEHAVIOR \\
\midrule
GPT-6 Astra & 23.07 & 22.45 \\
Embodied-R1.5 & 12.88 & 14.46 \\
RoboBrain2.5-8B & 10.08 & 12.54 \\
RynnBrain11-9B & 21.88 & 21.21 \\
Molmo2-ER-4B & 18.08 & 18.31 \\
Qwen3.5-9B & 20.06 & 19.45 \\
Qwen3.5-9B + Harness & 22.91 & 22.29 \\
Qwen3.8-27B & 20.68 & 21.68 \\
Qwen3.8-27B + Harness & 22.94 & 22.41 \\
\bottomrule
\end{tabular}
\par\smallskip
\begin{minipage}{\linewidth}
\footnotesize

\end{minipage}
\end{table}

Table~\ref{tab:average_steps} reports average high-level calls; these should be interpreted jointly with task completion.
On ProcTHOR, RoboBrain2.5-8B uses 10.08 steps with 5.8\% success,
whereas Qwen3.8-27B uses 20.68 steps and achieves 21.1\% success.
Shorter trajectories therefore do not necessarily imply more effective execution.

\section{Blind Evaluation}
\label{app:blind_testing}
To verify that SCOPER-bench requires visual observations rather than textual shortcuts, we replace all visual inputs with black images. Tables~\ref{tab:behavior-blind} and~\ref{tab:procthor-blind} report the results for BEHAVIOR and ProcTHOR, respectively. All models exhibit near-zero task completion, relying largely on prior-based guesses. These results demonstrate that the benchmark requires visual perception and cannot be solved through simple instruction decomposition.

\begin{table*}[t]
  \centering
  \caption{\small BEHAVIOR blind testing by task group (\%).}
  \label{tab:behavior-blind}
  \begingroup
  \renewcommand{\modelband}[1]{\rowcolor{ModelBand}\multicolumn{16}{l}{\strut\color{ModelInk}\itshape #1}\\}
  \fontsize{7.2}{9.4}\selectfont
  \setlength{\tabcolsep}{2.0pt}
  \renewcommand{\arraystretch}{1.12}
  \resizebox{\textwidth}{!}{%
  \begin{tabular}{@{}l@{\hspace{7pt}}
  rrr@{\hspace{7pt}}
  rrr@{\hspace{7pt}}
  rrr@{\hspace{7pt}}
  rrr@{\hspace{7pt}}
  >{\columncolor{OverallBg}}r
  >{\columncolor{OverallBg}}r
  >{\columncolor{OverallBg}}r@{}}
  \toprule
  Model & \multicolumn{3}{c}{Routine Execution}
        & \multicolumn{3}{c}{Observation-dependent}
        & \multicolumn{3}{c}{Failure Recovery}
        & \multicolumn{3}{c}{Composite}
        & \multicolumn{3}{>{\columncolor{OverallBg}}c}{Overall} \\
  \cmidrule(lr){2-4}\cmidrule(lr){5-7}\cmidrule(lr){8-10}\cmidrule(lr){11-13}\cmidrule(lr){14-16}
   & SR & Prog. & Cov.
   & SR & Prog. & Cov.
   & SR & Prog. & Cov.
   & SR & Prog. & Cov.
   & SR & Prog. & Cov. \\
  \midrule

  \modelband{API model}
  GPT-6 Astra
  & 0.0 & 22.2 & 37.5
  & 1.1 & 18.5 & 34.8
  & 0.8 & 28.5 & 48.1
  & 2.0 & 29.5 & 30.1
  & 1.0 & 25.1 & 38.5 \\

  Gemini-3.8-flash
  & 0.0 & 14.5 & 24.7
  & 1.1 & 10.9 & 20.4
  & 0.8 & 17.8 & 30.0
  & 1.0 & 15.5 & 15.7
  & 0.7 & 15.0 & 23.3 \\

  \modelband{Embodied models}
  Embodied-R1.5
  & 2.0 & 24.8 & 16.0
  & 7.8 & 20.6 & 25.6
  & 8.5 & 18.9 & 23.5
  & 3.0 & 21.8 & 13.0
  & 5.5 & 21.3 & 19.6 \\

  RoboBrain2.5-8B
  & 1.0 & 20.6 & 11.0
  & 3.3 & 9.7 & 18.3
  & 4.6 & 14.2 & 16.8
  & 2.0 & 20.6 & 6.2
  & 2.9 & 16.3 & 13.2 \\

  RynnBrain11-9B
  & 1.0 & 22.5 & 11.4
  & 4.4 & 8.6 & 10.4
  & 3.1 & 11.4 & 13.8
  & 2.0 & 18.0 & 3.6
  & 2.6 & 15.0 & 10.1 \\

  Molmo2-ER-4B
  & 0.0 & 21.4 & 6.4
  & 1.1 & 3.7 & 5.9
  & 6.2 & 17.3 & 9.8
  & 2.0 & 14.7 & 2.9
  & 2.6 & 14.7 & 6.5 \\

  \modelband{Open-source models}
  Qwen3.5-9B
  & 2.0 & 24.9 & 14.8
  & 10.0 & 14.9 & 16.8
  & 5.4 & 16.0 & 19.6
  & 2.0 & 20.3 & 7.6
  & 4.8 & 18.9 & 15.0 \\

  Qwen3.5-9B + harness
  & 2.0 & 23.4 & 28.5
  & 4.4 & 16.5 & 27.2
  & 1.5 & 23.3 & 36.5
  & 2.0 & 25.8 & 21.1
  & 2.4 & 22.5 & 28.9 \\

  Qwen3.8-27B
  & 5.0 & 24.9 & 21.5
  & 5.6 & 17.4 & 25.9
  & 7.7 & 23.2 & 27.4
  & 3.0 & 21.7 & 14.2
  & 5.5 & 22.0 & 22.5 \\

  Qwen3.8-27B + harness
  & 1.0 & 22.7 & 34.2
  & 2.2 & 18.4 & 33.2
  & 2.3 & 27.5 & 43.5
  & 2.0 & 28.1 & 27.1
  & 1.9 & 24.5 & 35.2 \\

  \bottomrule
  \end{tabular}%
  }
  \endgroup
  \vspace{-3mm}
\end{table*}

\begin{table*}[t]
  \centering
  \caption{\small ProcTHOR blind testing by task group (\%). }
  \label{tab:procthor-blind}
  \begingroup
  \renewcommand{\modelband}[1]{\rowcolor{ModelBand}\multicolumn{16}{l}{\strut\color{ModelInk}\itshape #1}\\}
  \fontsize{7.2}{9.4}\selectfont
  \setlength{\tabcolsep}{2.0pt}
  \renewcommand{\arraystretch}{1.12}
  \resizebox{\textwidth}{!}{%
  \begin{tabular}{@{}l@{\hspace{7pt}}
  rrr@{\hspace{7pt}}
  rrr@{\hspace{7pt}}
  rrr@{\hspace{7pt}}
  rrr@{\hspace{7pt}}
  >{\columncolor{OverallBg}}r
  >{\columncolor{OverallBg}}r
  >{\columncolor{OverallBg}}r@{}}
  \toprule
  Model & \multicolumn{3}{c}{Routine Execution}
        & \multicolumn{3}{c}{Observation-dependent}
        & \multicolumn{3}{c}{Failure Recovery}
        & \multicolumn{3}{c}{Composite}
        & \multicolumn{3}{>{\columncolor{OverallBg}}c}{Overall} \\
  \cmidrule(lr){2-4}\cmidrule(lr){5-7}\cmidrule(lr){8-10}\cmidrule(lr){11-13}\cmidrule(lr){14-16}
   & SR & Prog. & Cov.
   & SR & Prog. & Cov.
   & SR & Prog. & Cov.
   & SR & Prog. & Cov.
   & SR & Prog. & Cov. \\
  \midrule

  \modelband{API model}
  GPT-6 Astra
  & 4.6 & 38.6 & 43.4
  & 3.3 & 35.4 & 46.6
  & 3.8 & 35.1 & 51.4
  & 2.0 & 32.2 & 36.2
  & 3.6 & 35.5 & 44.7 \\

  Gemini-3.8-flash
  & 2.3 & 24.4 & 27.4
  & 3.3 & 24.0 & 31.1
  & 2.3 & 21.6 & 32.0
  & 1.0 & 20.4 & 23.6
  & 2.2 & 22.6 & 28.6 \\

  \modelband{Embodied models}
  Embodied-R1.5
  & 1.5 & 23.0 & 6.2
  & 5.6 & 12.6 & 15.0
  & 3.1 & 14.5 & 14.8
  & 1.0 & 22.2 & 4.9
  & 2.7 & 18.3 & 10.1 \\

  RoboBrain2.5-8B
  & 0.0 & 16.4 & 4.1
  & 3.3 & 9.9 & 8.9
  & 3.1 & 12.8 & 8.5
  & 2.0 & 15.6 & 6.4
  & 2.0 & 13.9 & 6.8 \\

  RynnBrain11-9B
  & 1.5 & 22.6 & 5.8
  & 2.2 & 8.8 & 6.7
  & 0.8 & 18.8 & 8.6
  & 0.0 & 14.9 & 5.4
  & 1.1 & 17.0 & 6.7 \\

  Molmo2-ER-4B
  & 0.0 & 18.4 & 3.2
  & 1.1 & 3.7 & 2.4
  & 0.0 & 10.6 & 2.9
  & 1.0 & 10.7 & 1.1
  & 0.4 & 11.5 & 2.5 \\

  \modelband{Open-source models}
  Qwen3.5-9B
  & 0.8 & 22.7 & 11.6
  & 6.7 & 16.1 & 15.4
  & 2.3 & 19.3 & 19.5
  & 2.0 & 24.1 & 11.9
  & 2.7 & 20.7 & 14.7 \\

  Qwen3.5-9B + harness
  & 3.8 & 31.8 & 31.2
  & 4.4 & 28.5 & 35.2
  & 2.3 & 28.0 & 37.3
  & 2.0 & 28.3 & 26.6
  & 3.1 & 29.3 & 32.7 \\

  Qwen3.8-27B
  & 2.3 & 29.9 & 29.5
  & 4.4 & 25.0 & 33.2
  & 3.8 & 30.6 & 33.3
  & 3.0 & 33.3 & 25.7
  & 3.3 & 29.9 & 30.5 \\

  Qwen3.8-27B + harness
  & 4.6 & 37.0 & 40.8
  & 3.3 & 32.7 & 43.3
  & 3.8 & 34.6 & 48.7
  & 2.0 & 32.6 & 33.6
  & 3.6 & 34.5 & 42.0 \\

  \bottomrule
  \end{tabular}%
  }
  \endgroup
  \vspace{-3mm}
\end{table*}

\section{Real-Robot Next-Tool-Call Evaluation}
\label{app:agibot}

\subsection{Data and Decision-Node Construction}

The evaluation uses real-robot demonstrations from AgiBotWorld2026,
excluding simulation data. The 200 distinct source episodes form
three mutually exclusive groups: 80 normal-execution, 60
observation-dependent, and 60 involving anomalous or incomplete
actions and their repair. Each sample covers a contiguous task scope
with a defined objective and termination condition within one source
video; videos are not spliced together. An episode can yield multiple
decision nodes, so episode counts differ from scored-decision counts.

Videos are segmented into semantic actions such as navigation,
picking, placing, and opening or closing containers. Local camera,
torso, and gripper adjustments are generally absorbed into their
semantic segment, whereas necessary base approaches are represented
as navigation. For reference calls $(a_1,\ldots,a_K)$, the node after
segment $i$ receives the reference prefix through $a_i$ and targets
$a_{i+1}$; the final segment is followed by \texttt{finish()}.
Scoring begins after the first segment and does not predict the
initial action. Every node uses reference history rather than
predictions from preceding nodes.

\subsection{Observations and Challenge Categories}

Visual inputs use the head-camera stream
\texttt{observation.images.top\_head}, sampled at 0.5 FPS on a fixed
temporal grid anchored at the start of the selected task scope.
Each node receives only grid frames within the current segment and
no later than its endpoint, without accumulating raw images from
earlier segments. Textual action history retains the complete
reference prefix. Future frames, manually selected keyframes,
bounding-box annotations, success or failure labels, category
labels, and reference answers are withheld.

English instructions are reconstructed from source annotations and
videos, specifying execution order where necessary. Observation-dependent
instructions describe inspection requirements without revealing the
object or state to discover; repair instructions do not identify a
failed action. Each observation-dependent episode includes at least
one node whose next call requires current visual evidence. Nodes
requiring future images to resolve ambiguity are excluded from scoring.

Recovery samples include naturally occurring anomalies with verifiable
subsequent recovery and incomplete-action nodes constructed by splitting
an original action before completion. For example, a pickup segment can
be divided into an unfinished pickup and its continuation, followed by
placement. The latter preserves the original images but does not imply
an actual grasp failure or controller retry. The two sources are labeled
separately, and decisions that resume the task after repair are evaluated.

\subsection{Harness State Along Reference Execution}

The harness maintains memory and the agenda along the reference
execution prefix. The initial agenda is derived from the instruction,
and state is updated using images and reference calls provided up to
the current node. Raw image input still covers only the current segment;
earlier observations can persist through memory. Predicted calls are
not executed, do not enter reference history, and do not authorize
committing expected effects. Subsequent updates follow the reference
video and actions without future frames or answer labels.

\subsection{Output Interface and Metrics}

Each node freely generates one structured call without candidate
options or a complete future plan. Tools include \texttt{navigate\_to},
\texttt{pick}, \texttt{place}, \texttt{open}/\texttt{close},
\texttt{turn\_on}/\texttt{turn\_off}, \texttt{soak\_under}, and
\texttt{finish}, with data-dependent extensions fixed before evaluation.
Object arguments are matched at category level using a frozen alias
mapping; instance identifiers, colors, coordinates, and hand selection
are not evaluated. Placement specifies the destination without repeating
the held object's category. The interface contains no \texttt{inspect} tool.

Tool Accuracy measures the fraction of nodes with the correct tool
name. The primary metric, Call Accuracy, requires the tool and all
necessary arguments to match one complete reference. Argument errors
invalidate the call but not a correct tool name; unparseable outputs
or unknown tools fail both metrics. The argument-free \texttt{finish()}
receives identical judgments. Equivalent references must be matched
as complete calls rather than by combining their components.

Category and overall results are aggregated over decision nodes;
overall accuracy divides total correct predictions by all scorable
nodes. Observation-critical Call Accuracy and Repair-decision Call
Accuracy use their respective subsets without adding them again to
the overall denominator. Post-repair continuation nodes contribute
to the repair category. Because predictions do not control the robot
or subsequent observations, this evaluation reports no task success,
goal progress, action coverage, or average execution steps.

\section{Oracle System 1 and Teacher Policy}
\label{app:system1}

Oracle System~1 is skill-execution infrastructure shared by evaluated models and the teacher. It implements individual skill requests through rules; the teacher policy reads the task specification and selects skill sequences for reference execution and validation. We describe skill execution and teacher organization separately below.

\subsection{Unified Skill Abstraction}

System~1 converts high-level requests into controlled scene changes and provides visual observations of their execution and outcomes. Table~\ref{tab:skills} summarizes skills by shared semantics: navigation and observation establish information and interaction poses, while manipulation updates object poses, relations, or properties. Object-state processing can be realized by a dedicated skill or a composition of existing skills.

\begin{table}[htbp]
\centering\small
\caption{Unified System~1 skills. Rows describe semantic capabilities rather than identical API names across backends.}
\label{tab:skills}
\begin{tabular}{p{.19\linewidth}p{.33\linewidth}p{.39\linewidth}}
\toprule
Skill & Preconditions and input & Controlled execution and observations \\
\midrule
Navigation & Resolved object or room target; reachable interaction pose. & Follow an A* path, update the agent pose, and provide observations along the route. \\
Scan & Available viewpoints around the current pose. & Rotate the view and acquire multiple observations of the surroundings. \\
Target search & Public target reference. & Direct observation toward relevant views to obtain visual evidence about the target. \\
Pick & Pickupable, accessible, interaction-ready object. & Establish the held relation and make the object pose follow the executor. \\
Place / release & Held object; an admissible destination for placement. &
Place at a support or containment pose, or release at the current
position; remove the held relation and update spatial relations. \\
Open / close & Openable, accessible container. & Update openness and the corresponding joint or visual configuration. \\
Toggle on / off & Toggleable, interaction-ready device. & Change activation state and synchronize its appearance and applicable semantic effects. \\
Cleaning & Compatible object and required tool or water source. & Update cleanliness and its visual representation. \\
Slice / cut & Sliceable object with required tool and access conditions. & Update the sliced state and associated geometry or object identities. \\
Heat / cool & Compatible object and appropriately configured heat or cold source. & Update heated, cooked, or cooled states. \\
\bottomrule
\end{tabular}
\end{table}

\subsection{Tool Arguments and Model-Facing Outputs}
\label{app:tool_arguments}

Table~\ref{tab:tool_arguments} summarizes model-facing tool arguments. Unless noted otherwise, listed arguments are required strings; \texttt{object\_ref} uses a public concrete category or supported alias rather than a private role or object identifier. Skills share semantics, with interface differences stated only where necessary: ProcTHOR includes room navigation in \texttt{navigate\_to}, whereas BEHAVIOR additionally exposes room navigation and several state-processing tools. Models receive only the tool definitions available in the current backend.

\begin{longtable}{p{.25\linewidth}p{.28\linewidth}p{.38\linewidth}}
\caption{Public tool parameters and observation semantics.}\label{tab:tool_arguments}\\
\toprule Tool & Arguments & Meaning and returned observation \\ \midrule
\endfirsthead
\toprule Tool & Arguments & Meaning and returned observation \\ \midrule
\endhead
\texttt{navigate\_to} & \texttt{object\_ref} & Object/receptacle, or a public room label in ProcTHOR; route and endpoint images. \\
\texttt{navigate\_to\_room} & \texttt{room} & Public room label in BEHAVIOR; process images. \\
\texttt{scan} & None. & Multiple in-place views obtained by rotating the camera. \\
\texttt{look\_for} & \texttt{object\_ref} & Target-directed camera views; does not replace approaching the object. \\
\texttt{pick\_up} & \texttt{object\_ref} & Object to pick up; before-and-after images. \\
\texttt{put\_down} & \texttt{object\_ref} & Destination receptacle or support, not the held object; before-and-after images. \\
\texttt{drop} & None. & Release at the current position in BEHAVIOR; before-and-after images. \\
\texttt{open}, \texttt{close}, \texttt{toggle\_on}, \texttt{toggle\_off} & \texttt{object\_ref} & Container or device to operate; before-and-after images. \\
\texttt{slice} / \texttt{cut} & \texttt{object\_ref} & Food to cut, not the cutting tool; before-and-after images. \\
\texttt{wipe} & \texttt{object\_ref} & Object to clean with a held tool; before-and-after images. \\
\texttt{soak\_under} & \texttt{object\_ref} & Running water source for cleaning the held item; before-and-after images. \\
\texttt{fill} & \texttt{object\_ref}, \texttt{source} & Container and fluid source; before-and-after images. \\
\texttt{pour} & \texttt{object\_ref} & Receiver for contents of the held container; before-and-after images. \\
\texttt{heat}, \texttt{cool} & \texttt{object\_ref} & Heat/cold device receiving the held item; before-and-after images. \\
\texttt{answer} & \texttt{question\_id}, \texttt{value} & Posted question identifier and selected answer text; no new scene evidence. \\
\texttt{stop} & None. & End physical execution; no goal-success oracle is returned. \\
\bottomrule
\end{longtable}

Wiping, washing under a source, filling, pouring, and dedicated heating and cooling tools are exposed by BEHAVIOR; corresponding ProcTHOR state effects arise through the container, device, and placement operations described above. State transitions occur inside System~1, while the model receives prescribed execution images and a request-completion summary, not state values, held-object inventories, ground-truth spatial relations, or effect labels. Answering and stopping are protocol operations and do not acquire additional scene information.

\subsection{Navigation and Observation Implementation}

Navigation constructs its search space from scene collision and reachability information and uses A* to reach an interaction pose near the target. ProcTHOR searches a grid of reachable positions and executes movement and rotation along the path. BEHAVIOR searches an occupancy grid accounting for the agent footprint and validates connectivity and path feasibility; its eight-connected search disallows diagonal corner cutting through blocked cells.

Navigation, scanning, and target search provide images during execution through backend observation callbacks. Scanning covers surrounding views, while target search organizes target-directed observation. The evaluation interface selects and delivers observations to the model or harness, connecting scene discovery to action selection throughout execution.

\subsection{Manipulation and State Synchronization}

Manipulation controls poses, attachment relations, and object properties through rules, bypassing low-level grasp and contact dynamics. ProcTHOR's controlled pickup and placement maintain visual attachment and containment relations alongside simulator state operations. BEHAVIOR's symbolic executor updates held states, spatial relations, and properties and synchronizes these changes with rendered objects and joints. Both retain object-affordance, access, and interaction preconditions.

State processing follows the semantics of the corresponding operation. For example, ProcTHOR implements cleaning through sink and faucet interactions and cooling or heating through container and device operations. BEHAVIOR uses corresponding state operations and execution wrappers for cleaning, cutting, and temperature changes. A shared capability therefore need not correspond to an identically named standalone tool on both backends. The resulting scene and images provide evidence for subsequent outcome checking.

Manipulation skills use controlled state transitions that bypass
low-level motion and contact simulation while retaining immediate
pre- and post-action RGB observations for outcome checking.
Navigation and exploration additionally provide multi-frame process
observations, allowing the model to discover objects and conditions
during execution.

The teacher policy uses task specifications and environment truth to organize skill calls into reference trajectories. Routine strategies satisfy goal predicates, observation-dependent strategies handle scene issues to be discovered, recovery strategies address execution deviations, and composite strategies coordinate subtasks. Reference trajectories support feasibility, goal, and event validation and provide execution context for QA attachment. Evaluated models execute through the same System~1 skill interface.

\section{Controlled Failure Injection and Scene Disturbances}
\label{app:failures}

Controlled interventions turn routine skill execution into closed-loop tasks requiring monitoring and recovery. We distinguish execution failures that alter the current skill's outcome from scene disturbances that change the environment after an execution milestone. Both are specified by activation conditions and state transformations in the task configuration, allowing the same task goal to be paired with different recovery challenges.

\subsection{Activation and Scheduling}

We represent an event as $e=(\tau,r,h,\phi,n)$, where $\tau$ specifies an eligible skill or milestone, $r$ identifies the target role, $h$ defines activation conditions, $\phi$ specifies the state intervention, and $n$ bounds the number of firings. Conditions can depend on a particular action result, elapsed action steps, containment or non-held states, and the owning subtask being active. Configurations also support probabilities and cooldown intervals to control intervention frequency.

Execution failures intercept a matching skill call or modify its outcome, whereas scene disturbances check conditions and apply changes after the relevant execution milestone. The scheduler maintains event histories and firing counts and scopes events to their owning subtask. Eligibility and whether the intended state change was actually applied are recorded separately for task validation.

\subsection{Failure Mechanisms}

Table~\ref{tab:failure_modes} summarizes six concrete failure mechanisms. They alter holding, placement, openness, or navigation outcomes, requiring subsequent actions to be revised from actual observations. For example, a dropped placement releases the object outside its destination; recovery requires localization and pickup before another placement attempt.

\begin{table}[htbp]
\centering\small
\caption{Execution failures: activation, state consequences, and recovery requirements.}
\label{tab:failure_modes}
\begin{tabular}{p{.20\linewidth}p{.43\linewidth}p{.28\linewidth}}
\toprule
Mechanism & Trigger and realized consequence & Recovery requirement \\
\midrule
Grasp slip & A pickup releases the object at a recoverable nearby location. & Relocalize, approach, and pick again. \\
Dropped placement & Placement releases the object beside the destination, violating the required relation. & Reacquire the object and place it again. \\
Partial opening & An open request leaves insufficient openness. & Inspect and complete opening. \\
Partial closing & A close request leaves the container partly open. & Inspect and complete closing. \\
Wrong-target navigation & Navigation executes toward an alternative object. & Detect the mismatch and approach the intended target. \\
Incomplete navigation & The final agent pose falls short of the required interaction position. & Reassess the current pose and navigate onward. \\
\bottomrule
\end{tabular}
\end{table}

Object release and relocation use controlled pose and held-relation updates and refresh subsequent observations. Dropped placement selects a location outside the destination; BEHAVIOR additionally checks collision, reachability, and rendered visibility and restores the prior state if the relocation cannot be committed. Partial opening and closing modify openness and its visual configuration. Wrong-target navigation substitutes the executed target while retaining the original request for outcome checking.

Incomplete navigation simulates failure to arrive by modifying the final pose. ProcTHOR repositions the agent to an intermediate location between the start and endpoint after nominal navigation; BEHAVIOR selects a pre-terminal pose from the planned route. This mechanism creates an outcome requiring further navigation without adding a physical obstacle to the scene.

\subsection{State-Changing Disturbances}

Scene disturbances can reclose or reopen containers, toggle devices, re-dirty objects, and relocate or restore objects. Activation connects these changes to meaningful task stages: for example, moving a staged object once it is no longer held, or re-dirtying an object after cleaning and intervening actions. Such events require the model to revisit previously valid facts and restore disrupted prerequisites or goals.

Both backends implement disturbances through scene-state changes and synchronize the visual representation of object poses, joints, or properties. Task validation checks final goals together with required interventions, ensuring that reference trajectories encounter the intended challenge. The evaluator retains private event records, while models assess consequences from protocol-provided observations and public interaction information.

\section{Task Templates and Construction}
\label{app:templates}

Templates translate capability requirements into instantiable object, state, and event constraints. We organize tasks into Routine Execution, Observation-dependent, Failure Recovery, and Composite. Below we describe shared template fields, instantiation, and representative instances. Tables below summarize each benchmark template through its roles, initialization requirements, and goal predicates.

\subsection{Dataset Statistics and Scene Sampling}
\label{app:dataset_statistics}

The construction target is ten task instances per base template in
 each simulator, together with 50 two-child and 50 three-child
 composite tasks per simulator. The 35 base templates thus yield a
 target of 450 tasks per backend. Limited matches across diverse
 BEHAVIOR scenes lead to the exclusion of 30 instances, leaving
 420 BEHAVIOR and 450 ProcTHOR tasks, or 870 tasks in total.
 ProcTHOR contains 130 routine-execution, 90 observation-dependent,
 130 failure-recovery, and 100 composite episodes. BEHAVIOR contains
 100, 90, 130, and 100 episodes in the corresponding groups, with
 the 30 excluded instances belonging to routine execution.
 ProcTHOR scenes are randomly sampled from its official pool of
 12,000 scenes. For BEHAVIOR, filtering out excessively large scenes
 leaves a pool of 35 distinct scenes for task matching; task objects
 are randomly added under the template's compatibility constraints.

\subsection{Template Specification and Instantiation}

Each template specifies object roles, categories or category groups, affordance requirements, and relations between roles. Initial conditions describe open/closed, cleanliness, containment, or location states, and goal predicates define the desired outcome. Instruction patterns replace roles with public object descriptions; event configurations define challenge triggers and consequences; composite tasks additionally retain subtask identities and scopes.

Scene matching first filters candidates by object type and affordance, then checks same-room, cross-room, distinctness, and spatial constraints. ProcTHOR profiles provide object and scene information. The BEHAVIOR matcher can reuse the shared JSON templates and distinguishes direct matches, matches requiring spawned assets, and matches requiring runtime validation. Bound tasks satisfy initial conditions through permitted relocation, state configuration, or asset materialization.

Initialization applies these configurations to concrete objects and synchronizes rendered state. Validation checks initial predicates, geometry and support relations, navigation and sequential interaction reachability, and whether all goals already hold. Teacher rollouts then check final goals and required events. Scenes unable to support a template retain explicit rejection reasons.

Instructions specify task objectives, scope, and handling rules without exposing construction-time instance bindings or step-by-step procedures. In observation-conditioned tasks, the model identifies the concrete objects requiring attention and their current states from scene observations. Public category terms express selection criteria rather than enumerate qualifying scene instances. The model consequently determines action targets, necessary prerequisites, and subsequent operations.

For tasks with universal inspection requirements, construction
ensures that all actionable issues within the instruction scope
are represented in template bindings and scoring goals; other
relevant objects are initialized in compliant states. Instructions
such as switching everything off or closing all containers therefore
introduce no additional initial issues outside the scoring scope.

For example, clean-dishware storage requires dishware, storage, and distinguishable dirty-dishware distractors. The matcher selects instances satisfying affordance and spatial constraints; initialization makes target dishware clean, makes distractor dishware dirty, and configures placement under the template constraints. The public instruction requests storing clean dishware and excluding dirty dishware from storage, without identifying which concrete instances belong to either group or prescribing a pickup-and-placement sequence. Final goals jointly require target storage, exclusion of distractors from storage, and container closure where applicable.

\subsection{Representative Templates}

Routine execution includes simple placement and preparation before storage. Simple placement transfers a pickupable object to a destination; preparation requires cleaning, cutting, heating, or cooling before the placement goal. Multi-stage preparation combines these requirements into an object workflow.

The dirty-dish workflow template includes dishware, sink, faucet, and storage roles and can include a staging surface. The dish is initialized dirty and outside storage, and the goal requires it to be clean and stored. The task connects identifying issues in the scene to subsequent cleaning and storage, making observations determine which objects and operations require attention.

Safety inspection initializes a running faucet and an open container in one room and requires both to be closed or switched off. Other observation templates include cross-room inspection, mixed issues, and category-dependent storage. Latent issue records specify the actual initial conditions and required handling, while the model identifies actionable conditions from visual observations.

The dropped-placement template binds a pickupable object and destination, with a goal of establishing the correct placement relation. A one-shot placement event releases the object beside the target, requiring recovery through approach, pickup, and return to the destination. Partial opening or closing, grasp slips, and incorrect navigation alter other execution effects, requiring different recovery procedures toward the same final goal.

Composite tasks contain two or three subtasks and retain each child's roles, goals, events, and identity. Runtime scheduling scopes events to their owning subtask and advances the active child at handoff. Handoffs are managed by the evaluation process rather than requested by the model. QA can cover different children and handoff progress; overall and per-subtask success are recorded separately.

\subsection{Template Structure and Predicate Catalog}

Table~\ref{tab:template_structure} summarizes template structure. Table~\ref{tab:shared_templates} describes the 35 benchmark task templates individually through object roles, initialization or event requirements, and goal predicates. Roles satisfy the affordance and access constraints of their operations; the tables highlight spatial conditions that distinguish task semantics.

\begin{table}[htbp]\centering\small
\caption{Semantic structure of a task template.}
\label{tab:template_structure}
\begin{tabular}{p{.23\linewidth}p{.68\linewidth}}\toprule
Component & Specification \\\midrule
Roles & Object categories, affordances, optionality, and inter-role relations. \\
Initialization & Object properties, initial containment and placement, and task-relevant initial visibility. \\
Goals & Object-state and spatial-relation predicates, including conditional or set-valued goals. \\
Instruction & Public descriptions grounded in bound roles and observation-conditioned handling rules. \\
Events and QA & Activation conditions, state interventions, questions, and scoring rules. \\
Composition & Child identities, scopes, and handoff relations. \\
\bottomrule\end{tabular}\end{table}

For a unified notation, $P(x,r)$ denotes the placement relation required by a template, instantiated as containment or surface support; $P(X,r)$ requires the relation for every target in the set. $\bigwedge_{d\in D}\neg P(d,r)$ requires each distractor to remain outside the destination. $C^*(r)$ requires closure only when the destination is openable. Optional roles contribute goals when instantiated. Event requirements define the challenge alongside initial states but activate during execution.

The benchmark uses 35 task templates. Routine Execution, Observation-dependent, Failure Recovery, and Composite are capability-level groups, and an individual template can involve several capabilities. Initially running faucets or dirty dishes are issues to discover through observation, whereas execution-failure templates alter outcomes during skill calls. Composite tasks are formed by composing the selected templates.

\begingroup\footnotesize
\setlength{\tabcolsep}{3pt}
\begin{longtable}{p{.22\linewidth}p{.35\linewidth}p{.35\linewidth}}
\caption{Benchmark task templates: roles, initialization, and goal predicates.}\label{tab:shared_templates}\\
\toprule Template & Roles and initialization / events & Goal predicates \\ \midrule
\endfirsthead
\toprule Template (continued) & Roles and initialization / events & Goal predicates \\ \midrule
\endhead
Simple placement & Object $x$, receptacle $r$ in the same room. \par $x$ visible; $\neg P(x,r)$. & $P(x,r)$. \\ \addlinespace
Device activation & Toggleable device $d$. \par $\neg\mathrm{On}(d)$. & $\mathrm{On}(d)$. \\ \addlinespace
Clean and place & Dirtyable $x$, sink $w$, faucet $f$, destination $r$. \par $\mathrm{Dirty}(x)$; compatible sink--faucet pair. & $\mathrm{Clean}(x)\land\allowbreak P(x,r)\land\allowbreak C^*(r)$. \\ \addlinespace
Heat and place & Object $x$, microwave $h$, destination $r$. \par $\neg\mathrm{On}(h)$. & $\mathrm{Hot}(x)\land\allowbreak P(x,r)\land\allowbreak \mathrm{Closed}(h)\land\allowbreak \neg\mathrm{On}(h)$. \\ \addlinespace
Cool and place & Object $x$, refrigerator $h$, destination $r\ne h$. \par Compatible cooling and placement roles. & $\mathrm{Cool}(x)\land\allowbreak P(x,r)\land\allowbreak \mathrm{Closed}(h)$. \\ \addlinespace
Slice and place & Sliceable $x$, knife $k$, destination $r$. \par Compatible slicing and placement roles. & $\mathrm{Sliced}(x)\land\allowbreak P(x,r)$. \\ \addlinespace
Slice, cool, and place & $x$, knife $k$, work surface $w$, refrigerator $h$, destination $r$. \par Roles co-located; distinct processing and destination roles. & $\mathrm{Sliced}(x)\land\allowbreak \mathrm{Cool}(x)\land\allowbreak P(x,r)$. \\ \addlinespace
Wash, heat, and store & $x$, sink $w$, faucet $f$, microwave $h$, storage $r$. \par $\mathrm{Dirty}(x)$. & $\mathrm{Clean}(x)\land\allowbreak \mathrm{Hot}(x)\land\allowbreak P(x,r)\land\allowbreak C^*(r)$. \\ \addlinespace
Store dishware & Dishware set $X$, storage $r$, optional distractors $D$. \par Mandatory target outside storage; optional relocation support. & $P(X,r)\land\allowbreak \bigwedge_{d\in D}\neg P(d,r)\land\allowbreak C^*(r)$. \\ \addlinespace
Store cleaning supplies & Cleaning supplies $X$, storage $r$, optional distractors $D$. \par Mandatory target outside storage; optional relocation support. & $P(X,r)\land\allowbreak \bigwedge_{d\in D}\neg P(d,r)\land\allowbreak C^*(r)$. \\ \addlinespace
Select clean dishware & Clean dishware $X$, dirty distractor $d$, storage $r$. \par Mandatory targets clean; $\mathrm{Dirty}(d)$; optional extra target. & $P(X,r)\land\allowbreak \neg P(d,r)\land\allowbreak C^*(r)$. \\ \addlinespace
Store food & Food set $X$, cold storage $r$, optional distractors $D$. \par Mandatory target outside storage. & $P(X,r)\land\allowbreak \bigwedge_{d\in D}\neg P(d,r)\land\allowbreak \mathrm{Closed}(r)$. \\ \addlinespace
Cross-room food storage & Food $x$, refrigerator $r$, source-room anchor. \par $\neg P(x,r)$; source and destination in different rooms. & $P(x,r)\land\allowbreak C^*(r)$. \\ \addlinespace
Explore and place & Object $x$, destination $r$. \par $x$ outside the initial view. & $P(x,r)\land\allowbreak C^*(r)$. \\ \addlinespace
Cross-room inspection & Faucet $f$, container $c$ in separate rooms. \par $\mathrm{On}(f)\land\allowbreak \mathrm{Open}(c)$; initially unseen issues. & $\neg\mathrm{On}(f)\land\allowbreak \mathrm{Closed}(c)$. \\ \addlinespace
Dirty-dish workflow & Dish $x$, sink $w$, faucet $f$, storage $r$, optional staging surface. \par $\mathrm{Dirty}(x)\land\allowbreak \neg P(x,r)$. & $\mathrm{Clean}(x)\land\allowbreak P(x,r)$. \\ \addlinespace
Mixed-issue inspection & Dish $x$, sink, faucet $f$, storage $r$, container $c$. \par $\mathrm{Dirty}(x)\land\allowbreak \mathrm{On}(f)\land\allowbreak \mathrm{Open}(c)$. & $\mathrm{Clean}(x)\land\allowbreak P(x,r)\land\allowbreak \neg\mathrm{On}(f)\land\allowbreak \mathrm{Closed}(c)$. \\ \addlinespace
Safety inspection & Faucet $f$, container $c$ in one room. \par $\mathrm{On}(f)\land\allowbreak \mathrm{Open}(c)$. & $\neg\mathrm{On}(f)\land\allowbreak \mathrm{Closed}(c)$. \\ \addlinespace
Storage-policy inspection & Misplaced object $x$, category-compatible storage $r$. \par $\neg P(x,r)$. & $P(x,r)$. \\ \addlinespace
Policy-based tidying & Sampled issue objects; optional faucet $f$. \par Scene-specific violations materialized by the generator. & Conjunction of instantiated issue goals; $\neg\mathrm{On}(f)$ when present. \\ \addlinespace
Place and handle dirty dish & Primary $x,r$; dirty dish $y$, sink, faucet, storage $s$. \par $\mathrm{Dirty}(y)$. & $P(x,r)\land\allowbreak \mathrm{Clean}(y)\land\allowbreak P(y,s)\land\allowbreak C^*(r)\land\allowbreak C^*(s)$. \\ \addlinespace
Handle faucet and dirty dish & Faucet $f$, sink, dish $x$. \par $\mathrm{On}(f)\land\allowbreak \mathrm{Dirty}(x)$. & $\neg\mathrm{On}(f)\land\allowbreak \mathrm{Clean}(x)$. \\ \addlinespace
Food storage with faucet issue & Food $x$, refrigerator $r$, faucet $f$. \par $\mathrm{On}(f)$. & $P(x,r)\land\allowbreak \mathrm{Closed}(r)\land\allowbreak \neg\mathrm{On}(f)$. \\ \addlinespace
Placement with open-container issue & Primary $x,r$; separate container $c$. \par $\mathrm{Open}(c)$; issue container initially unseen. & $P(x,r)\land\allowbreak \mathrm{Closed}(c)\land\allowbreak C^*(r)$. \\ \addlinespace
Dropped-placement recovery & Object $x$, destination $r$. \par Inject drop beside $r$ on placement. & $P(x,r)$. \\ \addlinespace
Grasp-slip recovery & Accessible pickupable $x$, destination $r$. \par Inject slip on pickup. & $P(x,r)$. \\ \addlinespace
Incomplete-navigation recovery & Object $x$, destination $r$. \par Inject incomplete navigation toward $x$. & $P(x,r)$. \\ \addlinespace
Wrong-navigation recovery & Object $x$, distractor $y$, destination $r$. \par Redirect navigation from $x$ to $y$. & $P(x,r)$. \\ \addlinespace
Partial-close recovery & Openable container $c$. \par $\mathrm{Open}(c)$; inject partial closing. & $\mathrm{Closed}(c)$. \\ \addlinespace
Partial-open recovery & Openable container $c$. \par $\mathrm{Closed}(c)$; inject partial opening. & $\mathrm{Open}(c)$. \\ \addlinespace
Destination reclosure & Object $x$, destination container $r$. \par $\neg P(x,r)\land\allowbreak \mathrm{Closed}(r)$; reclose during execution. & $P(x,r)\land\allowbreak \mathrm{Closed}(r)$. \\ \addlinespace
Object relocation & Object $x$, destination $r$, faucet interruption role. \par Relocate the object during execution. & $P(x,r)$. \\ \addlinespace
Cleanliness reversal & Dirtyable $x$, sink, faucet $f$, destination $r$. \par $\mathrm{Dirty}(x)\land\allowbreak \neg\mathrm{On}(f)$; re-dirty during execution. & $\mathrm{Clean}(x)\land\allowbreak P(x,r)$. \\ \addlinespace
Inspect and restore a storage container &
Container $c$ and optional dishware $D$.
\par Vary initial contents and openness independently;
require a valid observation of the interior. &
$\mathrm{Inspected}(c)\land\allowbreak
[(H_0(c)\land\mathrm{Closed}(c))\lor\allowbreak
(\neg H_0(c)\land\mathrm{Open}(c))]$.
\\ \addlinespace
Placement with added instruction & Object $x$, destination $r$, light $l$. \par $\neg P(x,r)\land\allowbreak \mathrm{On}(l)$; a runtime instruction adds a requirement. & $P(x,r)\land\allowbreak C^*(r)\land\allowbreak \neg\mathrm{On}(l)$. \\ \addlinespace
\bottomrule
\end{longtable}
\endgroup

Here, $H_0(c)$ records whether dishware is present in the container
at initialization; moving contents during execution does not change
the branch. $\mathrm{Inspected}(c)$ records acquisition of an interior
observation sufficient to assess the contents, rather than merely
requesting an observation tool. Construction validation requires
visually discriminable evidence for both presence and absence;
an occluded view does not establish inspection. The evaluator records whether an interior view delivered through the public observation interface satisfies the task's visual inspection condition; invoking a tool without obtaining such a view does not satisfy it. This template permits
initial openness to satisfy the state requirement, but the inspection
condition is initially unmet, so the complete task still requires
execution. This prevents initial openness from determining the
correct branch.

\subsection{Programmatic Instantiation and Composition}

The five observation-conditioned inspection and workflow templates instantiate scene issues through programmatic recipes. Cross-room inspection configures one faucet or container issue in each of two rooms; safety inspection configures faucet and container issues in one room; mixed inspection configures faucet, container, and dirty-dish issues in one room. The dirty-dish workflow recipe configures one dish issue, and the storage-policy recipe configures one misplaced object. Each issue expands into concrete roles, initial states, and handling goals.

The general long-horizon constructor combines two or three subtasks while retaining their roles, events, and goal ownership. Compatible object and state requirements allow the children to share an episode while preserving their individual task specifications. The overall goal collects the required child outcomes, while events and QA are scheduled within child scopes.

\subsection{Task Instructions and Execution Mechanisms}
\label{app:instruction_examples}

Table~\ref{tab:instruction_examples} pairs representative public
instructions with categories and execution mechanisms for 35
templates: 13 routine, nine observation-dependent, and 13 recovery
templates. Conditional container inspection belongs to the
observation-dependent group. Composite tasks combine
atomic templates and are not counted again as atomic entries.

Categories follow the same template-family grouping as the result
 tables, rather than mutually exclusive capability requirements.
 For example, routine clean-dishware selection still requires visual
 judgment; device activation and placement with an added instruction
 belong to the observation-dependent group. Failure Recovery includes
 four \texttt{recover\_*} templates that address initially present
 anomalies, six skill-outcome failure mechanisms, and three
 mid-execution disturbances. The mechanism column describes
 construction settings and is not additional model input.

Object names illustrate possible scene instantiations, and instructions
are semantic examples rather than verbatim evaluation prompts.
Event triggers and parameters remain task-specific; an intervention
is not implied for an entry without a listed disturbance. Situated QA
can be attached separately without changing the underlying category.

\begingroup
\footnotesize
\setlength{\tabcolsep}{3pt}
\begin{longtable}{p{.09\linewidth}p{.21\linewidth}p{.34\linewidth}p{.27\linewidth}}
\caption{Representative task instructions, template categories, and key execution mechanisms. R: routine; O: observation-dependent; F: failure recovery. Category and mechanism annotations are not shown to the model.}
\label{tab:instruction_examples}\\
\toprule
Category & Template & Representative instruction & Key execution mechanism \\
\midrule
\endfirsthead
\toprule
Category & Template & Representative instruction & Key execution mechanism \\
\midrule
\endhead

R & Simple placement &
Put the mug on the table. &
Transfer the object to its destination. \\
\addlinespace

O & Device activation &
Turn on the desk lamp. &
Device initially off; activate it. \\
\addlinespace

R & Clean and place &
Wash the bowl and put it in the cabinet.
Close the cabinet when finished. &
Object initially dirty; establish cleanliness and storage. \\
\addlinespace

R & Heat and place &
Heat the potato in the microwave, then put it on the plate.
Leave the microwave closed and switched off. &
Heating must precede final placement; leave the microwave off and closed. \\
\addlinespace

R & Cool and place &
Cool the apple in the refrigerator, then move it to the table.
Leave the refrigerator closed. &
Cooling must precede transfer to a different destination. \\
\addlinespace

R & Slice and place &
Slice the apple and put it on the plate. &
Slicing requires a compatible cutting tool. \\
\addlinespace

R & Slice, cool, and place &
Slice the apple, cool it in the refrigerator,
and then put it on the plate. &
Combine cutting, cooling, and final placement. \\
\addlinespace

R & Wash, heat, and store &
Wash the potato, heat it, and put it in the cabinet.
Close the cabinet when finished. &
Maintain cleaning and heating requirements through storage. \\
\addlinespace

R & Store dishware &
Put away the dishware in this room in the cabinet.
Do not put other items in the cabinet. Close the cabinet when finished. &
Select dishware from the scene; exclude distractors from storage. \\
\addlinespace

R & Store cleaning supplies &
Put away the cleaning supplies in this room in the cabinet.
Do not put other items in the cabinet. Close the cabinet when finished. &
Select cleaning supplies; exclude unrelated objects. \\
\addlinespace

R & Select clean dishware &
Inspect the dishware in this room and put only the clean
items in the cabinet. Do not put dirty dishware in the cabinet. Close the cabinet when finished. &
Distinguish clean targets from dirty distractors visually. \\
\addlinespace

R & Store food &
Put the food items in this room in the refrigerator.
Do not put non-food items in the refrigerator. Close the refrigerator when finished. &
Select food for cold storage; exclude non-food distractors. \\
\addlinespace

R & Cross-room food storage &
Move the apple from the dining room into the kitchen
refrigerator. Close the refrigerator when finished. &
Source object and destination are in different rooms. \\
\addlinespace

O & Explore and place &
Find the mug and put it in the cabinet. Close the cabinet when finished. &
Target object starts outside the initial view. \\
\addlinespace

O & Cross-room inspection &
Inspect the kitchen and bathroom. Ensure all switches are off and all containers are closed. &
Initially running faucet and open container occur in different rooms. \\
\addlinespace

O & Dirty-dish workflow &
Inspect the kitchen. Wash any dirty items you find and put them in their designated storage. &
Discover initially dirty dishware and complete washing and storage. \\
\addlinespace

O & Mixed-issue inspection &
Inspect the kitchen. Ensure all switches are off and all containers are closed. Wash any dirty items you find and put them in their designated storage. &
Discover coexisting faucet, container, and dirty-dish issues. \\
\addlinespace

O & Safety inspection &
Inspect this room and ensure all switches are off and all containers are closed. &
Discover an initially running faucet and open container in one room. \\
\addlinespace

O & Storage-policy inspection &
Inspect this room for misplaced items and put them away
according to these rules: food belongs in the refrigerator,
and dishware belongs in the cabinet. &
Identify misplaced objects and choose destinations by category. \\
\addlinespace

R & Policy-based tidying &
Inspect and tidy this room. Return misplaced items to
their designated locations, address dirty dishware,
and ensure that any faucets you encounter are off. &
Scene-specific issues are instantiated from the tidying rules. \\
\addlinespace

F & Place and handle dirty dish &
Put the mug on the table. Also inspect the kitchen
dishware and wash and store any items that need attention.
Close any storage containers you use. &
An initially dirty dish adds observation-dependent work to placement. \\
\addlinespace

F & Handle faucet and dirty dish &
Inspect the sink area, wash any dirty dishware, and ensure that the faucet is off. &
Initially running faucet and dirty dish require inspection and handling. \\
\addlinespace

F & Food storage with faucet issue &
Put the apple in the refrigerator and close it. Also inspect the kitchen and ensure all switches are off and all containers are closed. &
An initially running faucet must be addressed alongside food storage. \\
\addlinespace

F & Placement with open-container issue &
Put the mug on the table. Also inspect this room
and close any open cabinets or drawers you find. &
A separate container starts open and outside the initial view. \\
\addlinespace

F & Dropped-placement recovery &
Put the mug on the table. &
Placement injection releases the object beside the destination; reacquire it. \\
\addlinespace

F & Grasp-slip recovery &
Put the mug on the table. &
Pickup injection causes a slip and release; relocalize and pick again. \\
\addlinespace

F & Incomplete-navigation recovery &
Put the mug on the table. &
Navigation ends short of the interaction pose; continue approaching. \\
\addlinespace

F & Wrong-navigation recovery &
Put the mug on the table. &
Navigation is redirected to a distractor; approach the intended target. \\
\addlinespace

F & Partial-close recovery &
Close the cabinet. &
Closing leaves the container partly open; inspect and finish closing. \\
\addlinespace

F & Partial-open recovery &
Open the cabinet. &
Opening leaves insufficient openness; inspect and finish opening. \\
\addlinespace

F & Destination reclosure &
Put the mug in the cabinet and close the cabinet
when finished. &
A milestone-triggered disturbance recloses the destination container. \\
\addlinespace

F & Object relocation &
Put the mug on the table. &
A milestone-triggered disturbance relocates the object; find it again. \\
\addlinespace

F & Cleanliness reversal &
Wash the bowl and put it on the table. &
A disturbance makes the cleaned object dirty again; restore cleanliness. \\
\addlinespace

O & Inspect and restore a storage container &
Inspect the cabinet's contents.
If it contains dishware, close it; otherwise, leave it open. &
Vary contents and initial openness independently; require interior
inspection and the corresponding final state. \\
\addlinespace

O & Placement with added instruction &
\textit{Initial instruction:} Put the mug on the table.
\textit{Later message:} Also turn off the desk lamp
before finishing. &
A runtime message adds a new requirement to the initial placement task. \\
\addlinespace

\bottomrule
\end{longtable}
\endgroup

Composite tasks preserve child goals and event scopes and
can combine routine, observation-dependent, and recovery requirements.

\section{Binding Procedure}
\label{app:binding}

Binding connects action requests expressed in public vocabulary to instance-level System~1 execution. It comprises construction-time role binding and runtime reference resolution: the former determines the task's concrete scene instances, while the latter converts the model's selected skill and object reference into an executable request. Both backends use shared resolution semantics and maintain simulator-instance mappings inside their adapters.

\subsection{Construction-Time Binding and Public References}

During construction, template roles are bound to scene objects according to category, affordance, and spatial constraints. The internal mapping supports initialization, event scheduling, goal scoring, and runtime instance resolution. The model uses instructions and observations to select public object categories and operations, which the infrastructure resolves to concrete instances.

Tool instructions require object arguments to use a complete category name or an explicitly listed alias from the public vocabulary; room navigation uses public room labels from the instruction. Private roles, raw object identifiers, and invented color, position, or state modifiers are not valid references. Broad task concepts such as dishware or storage must be translated through observation into concrete categories supported by the interface.

\subsection{Runtime Resolution and Interaction Readiness}

At runtime, the resolver normalizes the public reference and filters candidates according to reference-admission rules. Resolution first considers compatible recent interaction focus and task-designated instances, followed by public aliases, activated observation or memory candidates, and permitted scene or room references. Internal task mappings can therefore participate in instance selection among objects of the same category: a task- or subtask-designated object can take precedence over background objects. The model selects the public object category and skill, and the infrastructure resolves the instance. Ambiguity that the applicable rules cannot resolve remains unresolved rather than selecting an arbitrary object.

Reference admission and interaction readiness are checked separately. By default, a model may attempt a public concrete category before seeing the object; candidates requiring observation evidence remain subject to their activation conditions. A valid name consequently guarantees neither existence, localization, nor readiness for manipulation. Interaction additionally requires approaching the target and satisfying affordance and access preconditions. After an object moves or drops, preserving its internal identity does not establish that the previous interaction pose remains valid.

For example, an instruction publicly specifies a drawer while other drawers are present in the scene. When the model requests navigation to \texttt{drawer}, the resolver can identify the corresponding drawer through the task mapping; a subsequent opening request using the same public name can retain the recent interaction focus. The model neither needs nor receives the drawer's simulator identifier, but must still select the opening operation and determine from returned images whether the drawer opened sufficiently. Likewise, when an instruction requires handling only dirty dishware, category resolution does not tell the model whether the dishware is dirty or replace its observation-based decision to wash it.

\subsection{Execution Results and the Model Information Boundary}

After reference resolution, the executor checks interaction preconditions and executes admitted skills. Unresolved or rejected requests retain their status for internal accounting. The model-facing summary contains the requested skill and public arguments and states that the call has finished; images provide evidence of the outcome. Binding diagnostics and instance mappings follow the information-access rules in Appendix~\ref{app:evaluation}.

System~1 provides instance-level execution support; System~2 selects actions, interprets scene observations, and assesses outcomes under the shared evaluation protocol.

\section{Situated QA}
\label{app:qa}

Situated QA turns understanding of execution state into explicit questions and scores. Questions are generated from template semantics and reference traces, while their actual answers are resolved in the evaluated episode. We describe question types, activation, and scoring below.

\subsection{Question Types and Construction}

The eight QA templates in Table~\ref{tab:qa_types} cover object states and spatial relations, action outcomes, active issues, new obligations, remaining requirements, and composite progress. For example, ``Is the bowl now in the cabinet?'' probes a relation, ``What was the outcome of the placement?'' probes an effect, and ``Which requirements remain incomplete?'' probes progress. Answers use the discrete choices associated with each question type.

All eight question types are generated in a unified pipeline from task
semantics, goal predicates, and reference execution traces. Candidate
questions are associated with public actions, task milestones, or other
configured activation conditions and checked for uniquely resolvable
answers. Selection considers diversity of question and action types and
subtask coverage in composite tasks, with up to six questions per atomic
episode. Scheduler replay validates activation before the selected
questions are stored in a separate QA specification.

Questions use public task vocabulary without exposing private instance
bindings or answer keys. Action-anchored activation matches the tool and
object role, avoiding step-offset mismatches introduced by intervening
answers. Answer keys are resolved from the actual evaluated episode
under the snapshot rules below, rather than copied from reference traces.
All eight QA types use a single selected answer from two to four choices.

\begin{table}[htbp]
\centering\small
\caption{The eight situated QA templates and the understanding they probe.}
\label{tab:qa_types}
\begin{tabular}{p{.30\linewidth}p{.61\linewidth}}
\toprule
Question type & Target of assessment \\
\midrule
Target state & Whether an object is currently clean, open, or activated. \\
Target relation & Whether an object is inside a container or on a designated surface. \\
Post-action outcome & Whether the preceding action achieved its intended effect. \\
Active issue & Which issue or execution discrepancy remains unresolved. \\
Inserted obligation & Which additional requirement was introduced by a new instruction. \\
Remaining requirement & Which task requirement has not yet been satisfied. \\
Mid-execution progress & Which goals are complete or incomplete at the current execution point. \\
Composite progress & Child completion and handoff status within a composite episode. \\
\bottomrule
\end{tabular}
\end{table}

The three progress-related templates use different scopes. Remaining-requirement questions ask which listed requirements remain unmet at a stopping milestone, for example, whether cleaning or storage remains incomplete. Mid-execution progress questions ask whether one or two specified requirements hold at an intermediate point, such as whether a bowl is clean and stored, with choices such as both, only the first, only the second, or neither. Composite-progress questions ask which subtask is active after a handoff, with choices identifying a child or indicating that all children are complete.

Each posted question is answered by the same task-executing model in a separate answer-only call. This call uses the currently available public execution context and performs no additional exploration or manipulation. QA therefore measures understanding of acquired information and is scored separately from physical task completion.

\subsection{Activation, Queuing, and Answer Time}

Activation types include a specified tool result, a task-action step threshold, subtask handoff, a new human instruction, and a stop milestone. Each question fires at most once; child-specific questions activate only within their owning subtask. Multiple eligible questions enter a queue, and the next question is posted only when no previous question remains pending.

Action-outcome questions snapshot the evaluator state of the triggering action; delayed questions identify the original action step. State, relation, and intermediate-progress questions resolve and freeze their answer when posted, so later actions do not change their scoring reference. Resolution failures are recorded separately. Public questions contain the question, choices, and identifier; answer keys remain with the evaluator.

\subsection{Scoring}

Situated QA uses the trigger rate and triggered accuracy in
Appendix~\ref{app:metrics}: valid posted questions divided by valid
configured questions, and correct answers divided by valid posted
questions. Posted but unanswered questions count as incorrect;
unposted questions are excluded from accuracy, and queued questions
are not counted as posted.

Correctness uses exact matching between normalized responses and
evaluator-held answer keys. Questions with unresolved answer keys
are excluded as invalid and reported separately; missing or
unparseable responses to valid posted questions remain incorrect.
The main tables report situated QA. Because execution paths may
trigger different question sets, accuracy is interpreted together
with trigger rate.

\section{Harness Details}
\label{app:harness}

SCOPER uses role-specific calls to the same fixed-parameter model for observation interpretation, agenda revision, action selection, and effect verification. Deterministic code handles structural and reference checks, memory routing, and bounded execution control. The following specification uses the same state definitions and information flow as the main text.

\subsection{Component Inputs and Outputs}

Initialization produces an agenda from the task instruction without images. Observation calls receive the instruction, memory to maintain, and available images, and propose evidence-linked updates. Agenda revision receives the instruction, existing items, and committed memory, and proposes additions and modifications. The policy receives the instruction, the latest four frames, up to the latest 48 public tool-request records, routed memory, the agenda, allowed tools and argument definitions, and recovery and anti-deadlock context, and outputs one structured tool call. Procedural guidance is supplied through item-level next intents and public tool descriptions.

Process observations from scan, navigation, and look-for calls are uniformly sampled into at most eight frames for observation maintenance; the policy uses the latest four available frames. Verification uses the existing immediate pre- and post-action frames, the request, and expected effects, without implicitly moving the camera. When QA is triggered, the same model answers in a dedicated turn using the question and the policy context, including the current instruction, images, public request history, agenda, routed memory, and control context; it outputs only the answer.

\subsection{Agenda Representation and Updates}

Each agenda item stores a stable identifier, requirement, mode, status, evidence references, and optional next intent. Modes are \texttt{required}, \texttt{ongoing}, or \texttt{conditional}; statuses are \texttt{pending}, \texttt{active}, \texttt{watching}, \texttt{supported}, or \texttt{blocked}, indicating outstanding work, current focus, awaiting relevant evidence, evidence satisfying the current criterion, or a missing prerequisite. The agenda contains at most eight items with at most one active item; item order does not prescribe execution order.

Supported status is interpreted according to the requirement. A state inspection is supported when the requested information is observed; an execution requirement is supported when visual verification establishes its intended effect; an ongoing constraint is supported only for the current state and remains monitored. Discovering that a condition holds activates the associated work without establishing its fulfillment. Request history and temporary planning assumptions cannot independently support completion. Later observations contradicting supporting evidence reopen affected items.

Changes in public task requirements or relevant committed memory trigger revision rather than rewriting the agenda at every call. Prompts ask the model to propose task-scoped additions, necessary prerequisites, recovery needs, and duplicate merging; the commit layer checks fields, identifiers, references, and evidence eligibility for supported status. The model judges task relevance and semantic duplication; deterministic validation enforces the record structure and evidence requirements. The agenda advises the policy without directly defining action admissibility.

Initialization establishes explicit requirements and conditional
inspection needs from the instruction. Unresolved inspection items
and next intents guide the policy's choice of observation tools and
targets. New public instructions are incorporated into task context;
either instruction changes or relevant memory changes trigger agenda
revision. QA questions are not treated as additional task requirements.

For a task to inspect a kitchen and turn off running faucets, initialization establishes an inspection need and a conditional shutoff requirement. Observing running water creates a scene fact and activates shutoff work without creating a failure incident. The shutoff requirement becomes supported only when post-action visual evidence establishes that flow has stopped. Continued flow instead records an execution incident and preserves outstanding work. Discovering one faucet does not automatically complete an inspection covering the entire kitchen.

\subsection{Memory Maintenance and Routing}

Facts store an identifier, public subject, property or relation, value, confidence, frame and action references, observation step, and validity status: \texttt{current}, \texttt{stale}, or \texttt{contradicted}. These fields inform the model's semantic judgments, while deterministic routines enforce record consistency and manage memory storage and routing. Incidents store an identifier, source action, discrepancy type, evidence, creation and update steps, handling status, and recovery attempts. Obligations store an identifier, \texttt{observe} or \texttt{act} type, relevant public objects, optional source incident, creation step, and \texttt{open} or \texttt{resolved} status.

Pre-decision observation maintenance distinguishes discovery from reconciliation. Discovery adds visible objects and conditions as scene-observation facts without independently creating incidents or obligations. Reconciliation assigns candidate facts a confirmed, contradicted, unobservable, or unknown assessment. Confirmation refreshes supporting evidence; contradiction marks a fact as \texttt{contradicted}; unobservable and unknown assessments alone do not authorize negation. A newer scene description of the same subject can supersede an earlier one. Proposals reference the relevant facts and images; the commit layer checks memory revision, reference existence, evidence eligibility, and consistency between assessments and mutations before commitment.

Default capacity is 64 facts, 16 incidents, and 16 obligations; routed context contains at most six facts, two incidents, and three obligations. Fact ordering first prioritizes currently relevant objects, then penalizes scene observations unrelated to the instruction, and then favors recent observations with stable identifiers breaking ties. Currently relevant objects come from unresolved needs and recent public interaction targets. Incidents prioritize unresolved entries and newer creation times; obligations prioritize unresolved entries and older creation times. Routing preserves validity and evidence references to distinguish historical records from current evidence.

Capacity maintenance preferentially removes stale facts and resolved records; unresolved incidents and obligations cannot be treated as handled through silent deletion. Post-action effect commitment complements pre-decision observation maintenance: the former records verification outcomes and corresponding recovery needs, while the latter handles additions and contradictions from later observations without readjudicating the same before/after evidence.

\subsection{Verification, Recovery, and Planning Assumptions}

Requests record selected actions, admission determines whether they may
execute, and visual verification assesses their effects. The evaluator
separately determines task success under Appendix~\ref{app:metrics};
control decisions and budget exhaustion do not establish success.

Verification success indicates visual support for the intended effect, failure indicates an observable discrepancy, and ambiguity indicates insufficient evidence. Outputs include evidence and observability assessments; occlusion or a lack of obvious image change does not automatically establish failure. Success commits only supported effects; failure creates or updates incidents and associated obligations; ambiguity preserves unconfirmed effects and requests observation when needed.

Recovery state contains the original request, target, expected effect, evidence, handling phase, and remaining attempts. The controller can require observation, approach, prerequisite satisfaction, or retry. For placement recovery, \texttt{held}, \texttt{released}, and \texttt{unknown} respectively call for returning toward the destination, localization and pickup, or further observation first. Effects of recovery calls remain subject to verification.

If uncertainty persists after bounded observation or retry, recovery branches that permit continuation may retain temporary planning assumptions. These reside in recovery or control context, cannot be committed as verified facts, and cannot support agenda completion. An explicitly observed failure cannot be assumed successful; subsequent evidence may revise or revoke an assumption. Terminal branches end execution with unresolved state retained.

\subsection{Anti-Deadlock Control and Execution Order}

The exploration ledger groups scans, searches, and navigation by room or semantic target and tracks observation and preparation calls without advancing interaction. At the corresponding limits, the controller rejects repetitive candidates before execution, supplies a reselection reason, and restricts tools, promoting approach and then interaction where appropriate. Relevant new evidence or admitted physical calls reset the corresponding counters under the control rules; precondition-rejected calls and changes in timestamps or step indices do not.

Recovery observations, retries of the original action, and model reselection each have finite budgets. Controller-required recovery calls take precedence over ordinary action selection, with the agenda serving as policy context. QA uses dedicated answer turns outside exploration candidate selection. Execution, answer, and stop requests each consume a high-level step; internal model calls and reselection do not add environment steps.

\paragraph{Default control limits.}
Within a no-progress phase, the default limits are two scans, two
searches and two navigation calls per semantic target, four observation
calls in total, and six observation/navigation preparation calls in total.
A rejected candidate permits at most two additional model reselection
requests. Recovery contexts default to two retry rounds, while dedicated
placement recovery additionally permits at most eight recovery actions.
These limits apply within their respective ledgers and do not extend
the episode's environment-step budget.

Exhausted dedicated placement recovery terminates with a control failure.
Generic recovery branches that permit continuation retain unconfirmed
state and separately marked planning assumptions. Later evidence revises
these assumptions.

Algorithm~\ref{alg:harness} orders observation maintenance, agenda revision, routing, action selection, and effect commitment; Algorithm~\ref{alg:guard} details bounded action selection. $M$ denotes memory, $A$ the agenda, $R$ recovery state, $L$ the control ledger, $H$ public request history, and $W$ the available image buffer. Calls parameterized by $\theta$ share the same model; other operations perform deterministic control or environment interaction.

\begin{algorithm}[htbp]
\caption{SCOPER execution with evidence memory and an agenda}
\label{alg:harness}
\begin{algorithmic}[1]
\Require Instruction $x$, initial image $o$, tools $\mathcal S$, step budget $B$
\State $M,R,L,H\gets\emptyset$; $W\gets[o]$; $b\gets B$; $z\gets\textsc{Active}$
\State $X\gets x$; $E\gets\emptyset$
\State $A\gets\Call{AcceptInitial}{\mathrm{Initialize}_{\theta}(X)}$
\While{the episode is not terminated and $b>0$}
    \State $X\gets\Call{IncorporatePublicTaskUpdates}{X}$
    \State $J\gets\Call{UnprocessedEvidence}{W,E}$
    \If{$J\ne\emptyset$}
        \State $\Delta_M\gets\mathrm{Observe}_{\theta}(X,M,J)$
        \State $M\gets\Call{Commit}{M,\Delta_M}$
        \State $E\gets E\cup\Call{EvidenceIdentifiers}{J}$
    \EndIf
    \If{task context or relevant committed memory has changed}
        \State $\Delta_A\gets\mathrm{ReviseAgenda}_{\theta}(X,A,M)$
        \State $A\gets\Call{AcceptAgenda}{A,\Delta_A,M}$
    \EndIf
    \State $\widetilde M\gets\Call{Route}{M,X,H,R}$; $V\gets\Call{LatestFour}{W}$
    \State $c\gets\Call{ControlContext}{\mathcal S,R,L}$
    \State $\kappa\gets(X,V,\Call{RecentRequests}{H},A,\widetilde M,c)$
    \If{a question $q$ is pending}
        \State $a\gets\mathrm{Answer}_{\theta}(q,\kappa)$
        \State \Call{SubmitAnswer}{$a$}; $b\gets b-1$
        \State $H\gets H\mathbin{\Vert}[\Call{PublicRequest}{a}]$; \textbf{continue}
    \EndIf
    \If{$z=\textsc{AnswerOnly}$} \State \textbf{break} \EndIf
    \State $(a,R,L)\gets\Call{GuardedSelect}{\theta,\kappa,R,L}$
    \If{$a=\textsc{Exhausted}$}
        \State \textbf{terminate} with control-exhausted status
    \EndIf
    \State $H\gets H\mathbin{\Vert}[\Call{PublicRequest}{a}]$
    \If{$a=\texttt{stop}$}
        \State \Call{StopAndFreezeScore}{}; $b\gets b-1$; $z\gets\textsc{AnswerOnly}$
        \State \textbf{continue}
    \EndIf
    \State $o^-\gets\Call{CurrentImage}{W}$
    \State $(o^+,\mathcal V,e)\gets\Call{BindAndExecute}{a}$; $b\gets b-1$
    \If{the admitted action requires effect verification}
        \State $g\gets\Call{ExpectedEffect}{a}$
        \State $v\gets\mathrm{Verify}_{\theta}(a,g,o^-,o^+)$
        \State $M\gets\Call{CommitEffect}{M,a,v}$
        \State $R\gets\Call{UpdateRecovery}{R,a,v,M}$
    \EndIf
    \State $L\gets\Call{UpdateLedger}{L,a,e,M}$
    \State $W\gets\Call{AppendImages}{W,\mathcal V,o^+}$
\EndWhile
\end{algorithmic}
\end{algorithm}

\begin{algorithm}[htbp]
\caption{Bounded action selection}
\label{alg:guard}
\begin{algorithmic}[1]
\Require Model $\theta$, policy context $\kappa$, recovery state $R$, ledger $L$
\If{a required recovery call is due and its budget permits}
    \State \Return $(\text{required call},R,L)$
\EndIf
\State Apply recovery-exhaustion rules to $R$ when applicable
\If{these rules terminate execution} \State \Return $(\textsc{Exhausted},R,L)$ \EndIf
\For{a bounded number of policy requests}
    \State $\kappa\gets\Call{RefreshControlContext}{\kappa,R,L}$
    \State $a\gets\pi_\theta(\kappa)$
    \If{$\Call{Admissible}{a,R,L}$} \State \Return $(a,R,L)$ \EndIf
    \State $L\gets\Call{AdvanceCommitment}{L,a}$
    \State $\kappa\gets\Call{AddReselectionReason}{\kappa,a,L}$
\EndFor
\State \Return $(\textsc{Exhausted},R,L)$
\end{algorithmic}
\end{algorithm}

Observe comprises discovery and reconciliation with their respective update permissions. \textsc{RecentRequests} retains up to 48 public requests; policy and QA calls share context $\kappa$. UpdateRecovery maintains phases, budgets, and permitted planning assumptions. Agenda acceptance checks structure, references, and evidence eligibility, while the model handles task scope and semantic merging.

$X$ contains the initial instruction and subsequent public task
messages, while $E$ tracks processed image identifiers. Task updates
exclude QA questions and can trigger agenda revision without waiting
for a new visual fact.

Updated recovery state and control ledgers are returned to the main loop.

Memory maintenance tracks frame identifiers and update responsibilities.
A post-action image can support discovery of other scene information,
while its action-effect assessment is committed through verification.
Repeated frames retain their existing evidence identity.

\section{Execution Trace Visualization}
\label{app:traces}
The simulated execution traces in Figure~\ref{fig:trace_samples} and
this section are produced by Qwen3.8-27B with SCOPER-harness.
Some redundant intermediate exploration steps and frames are omitted
for readability. Real-data examples illustrate
next-tool-call evaluation conditioned on reference execution prefixes.

\begin{figure}[htbp]
\centering
\includegraphics[width=1.0\linewidth]{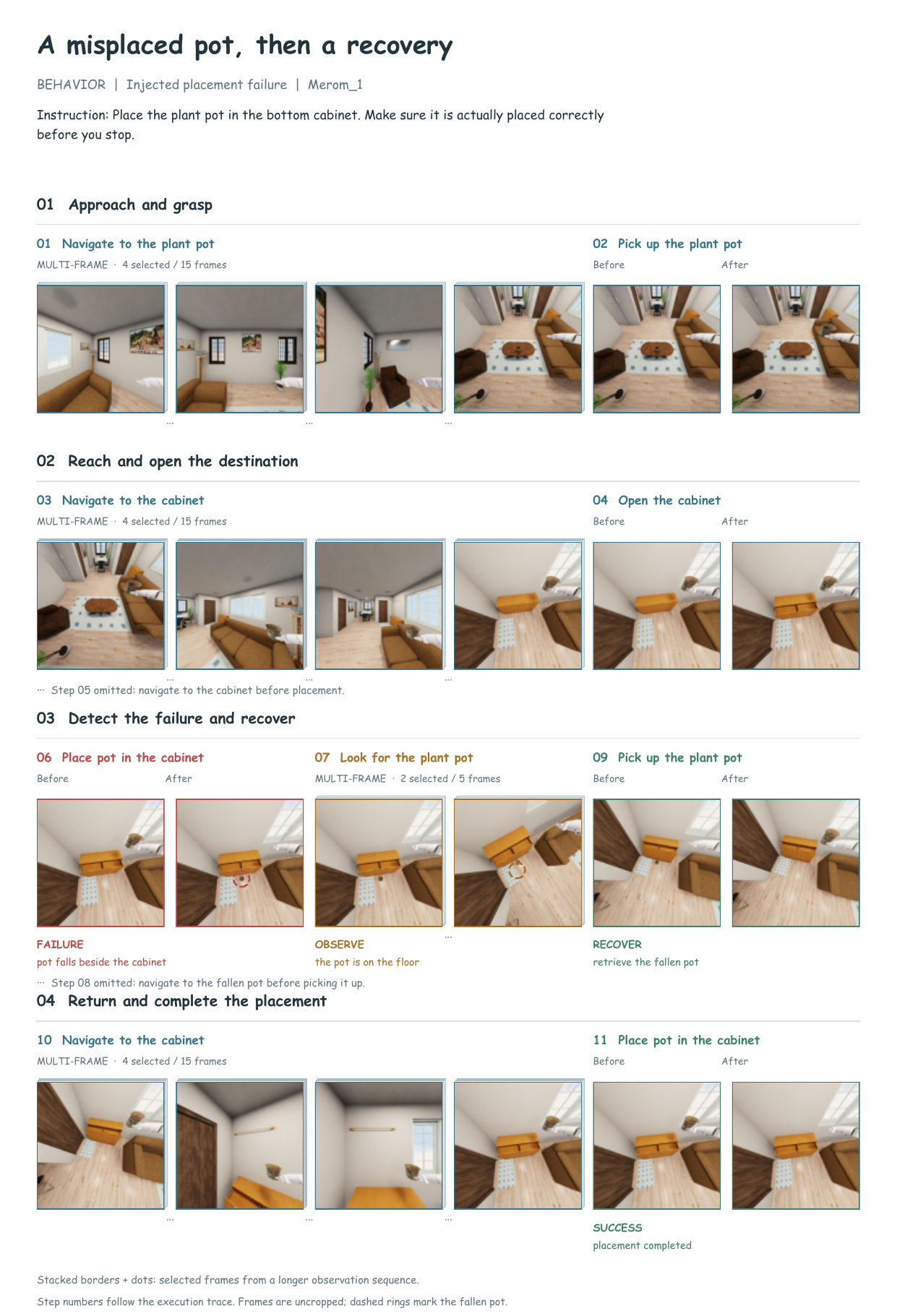}
\caption{Sample 1: Retrieving a fallen plant pot and completing placement.}
\label{fig:trace_sample_1}
\end{figure}

\begin{figure}[htbp]
\centering
\includegraphics[width=1.0\linewidth]{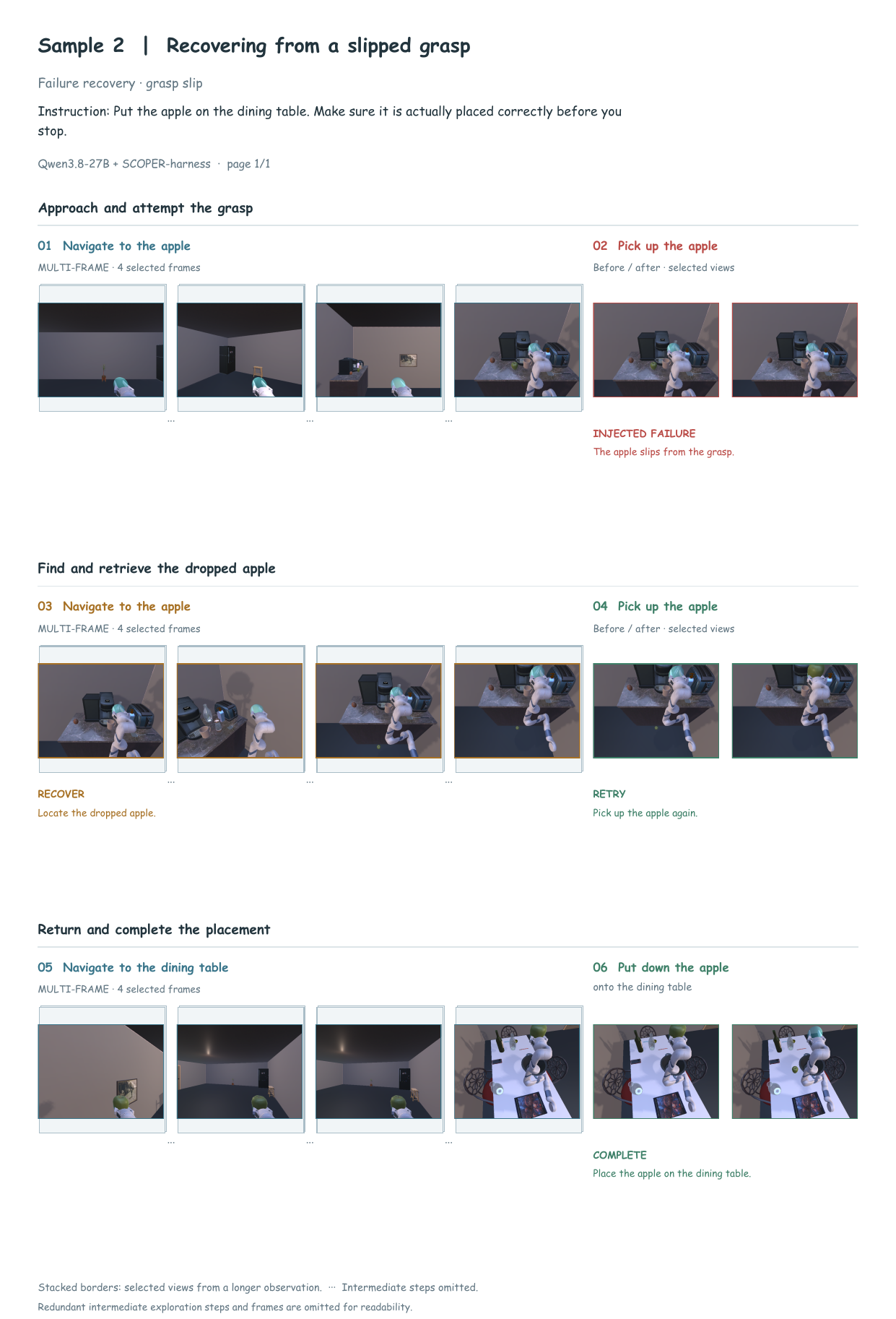}
\caption{Sample 2: Retrieving an apple after a slipped grasp.}
\label{fig:trace_sample_2}
\end{figure}

\begin{figure}[htbp]
\centering
\includegraphics[width=1.0\linewidth]{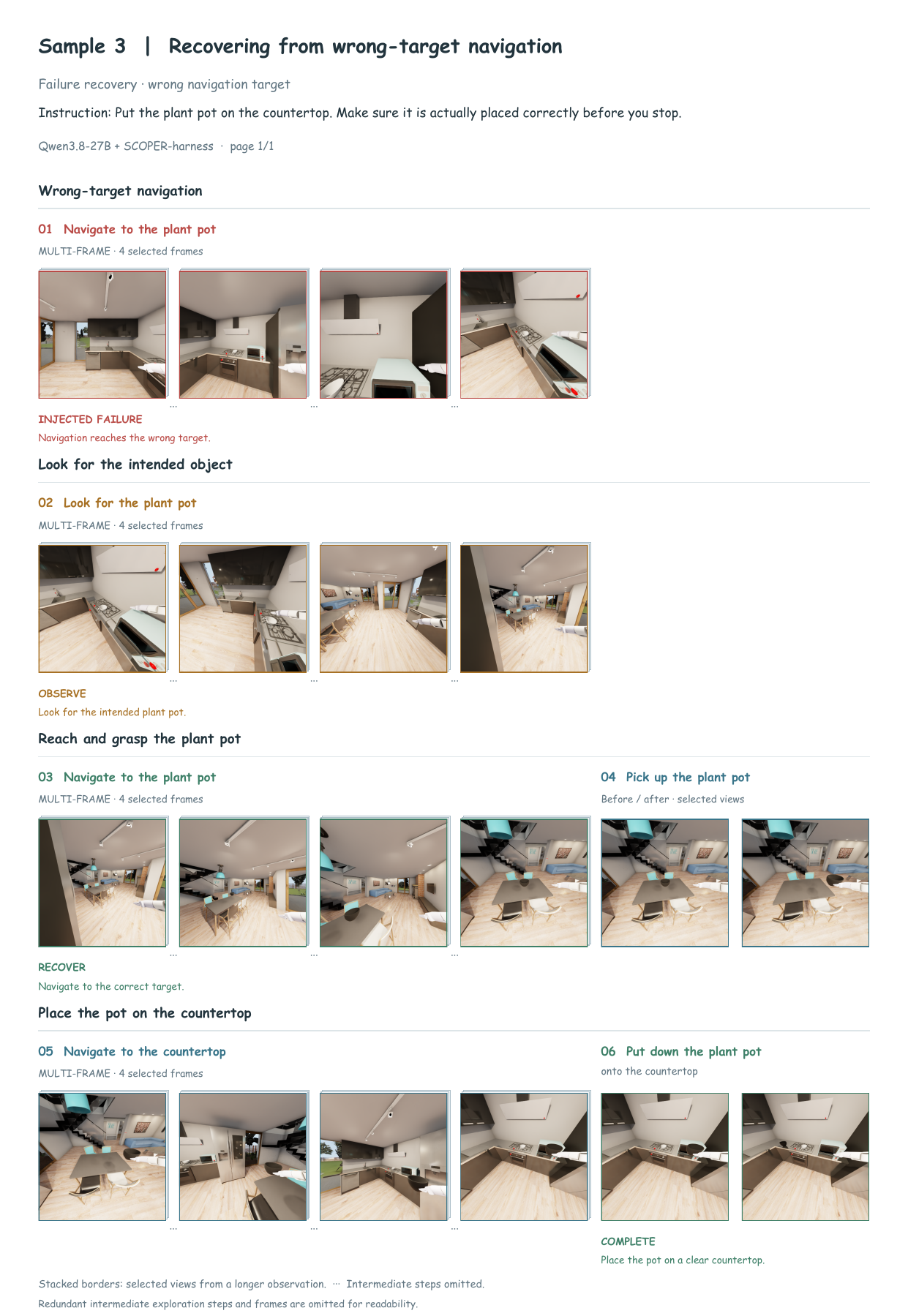}
\caption{Sample 3: Correcting navigation to the wrong target.}
\label{fig:trace_sample_3}
\end{figure}

\begin{figure}[htbp]
\centering
\includegraphics[width=1.0\linewidth]{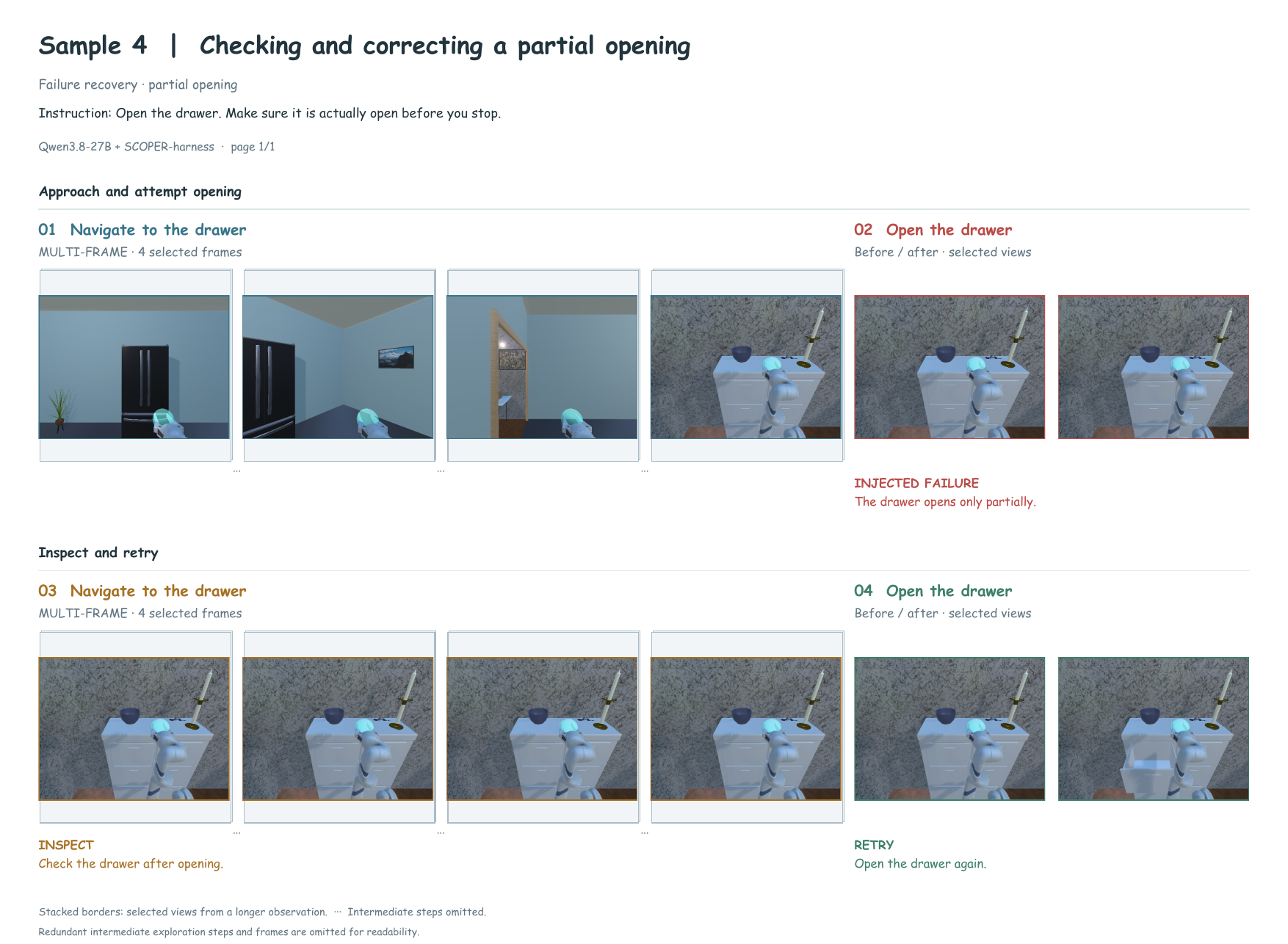}
\caption{Sample 4: Inspecting and retrying a partial drawer opening.}
\label{fig:trace_sample_4}
\end{figure}

\begin{figure}[htbp]
\centering
\includegraphics[width=1.0\linewidth]{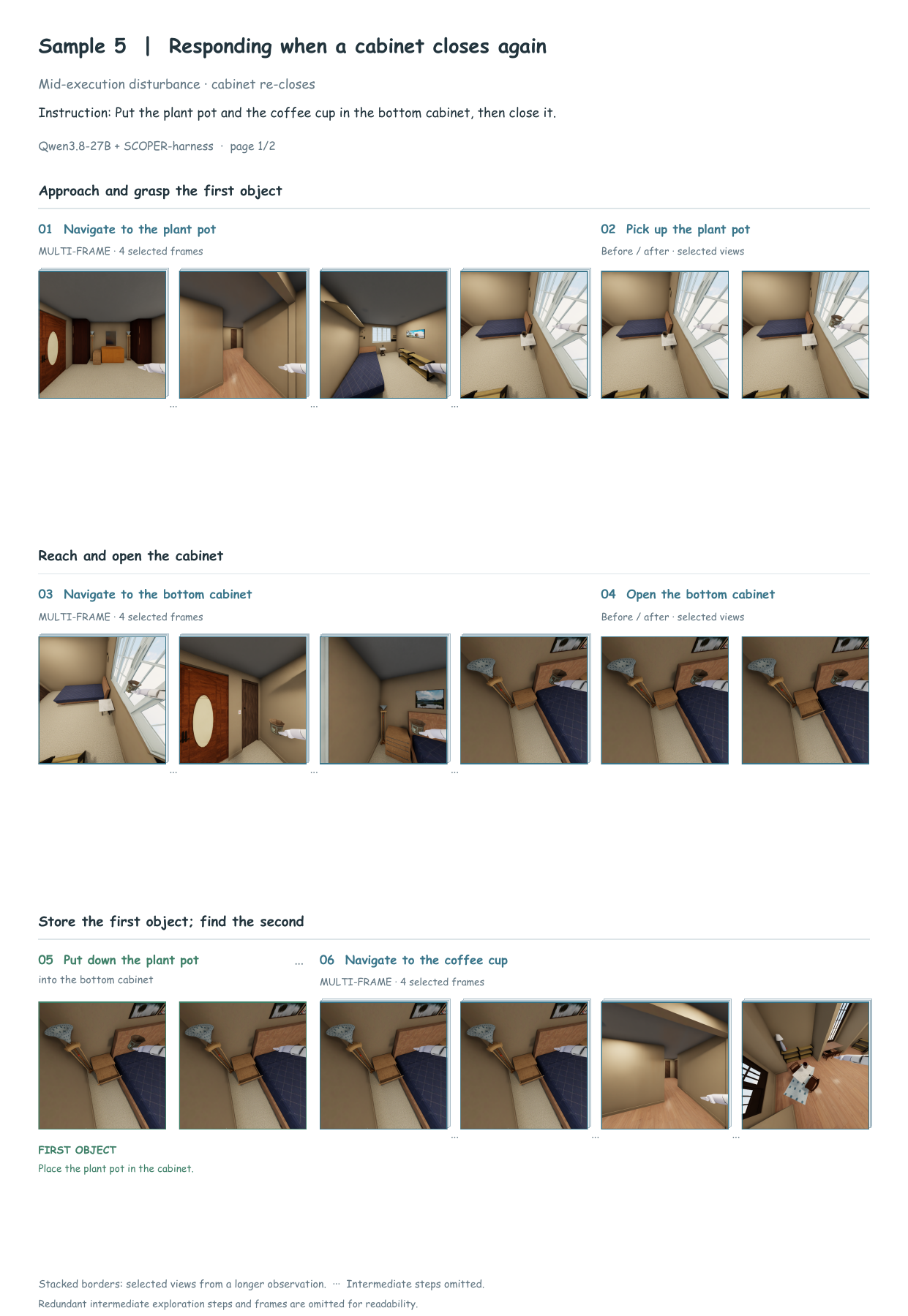}
\caption{Sample 5, part 1: Storing the plant pot and approaching the coffee cup.}
\label{fig:trace_sample_5}
\end{figure}

\begin{figure}[htbp]
\centering
\includegraphics[width=1.0\linewidth]{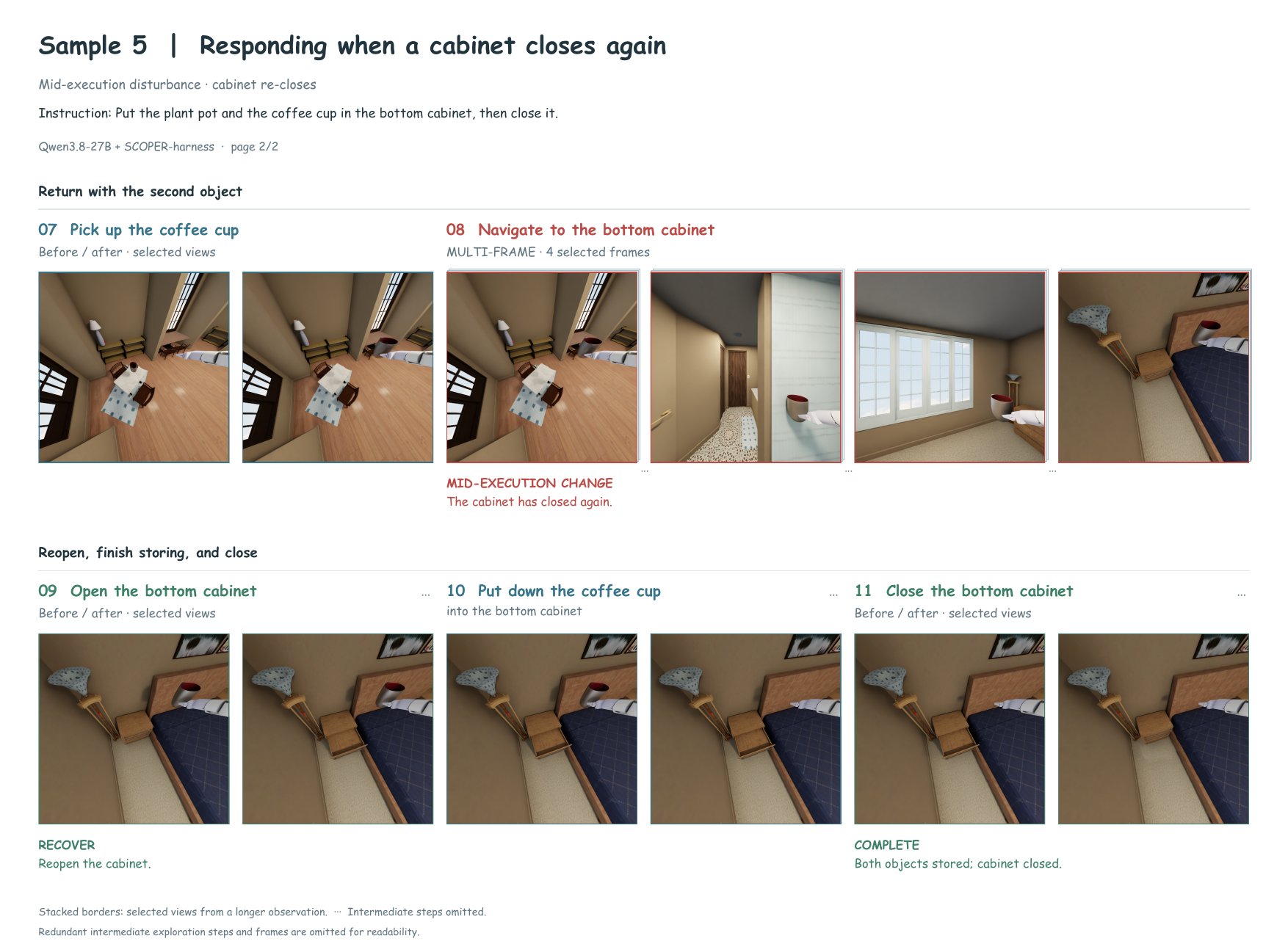}
\caption{Sample 5, part 2: Reopening the cabinet to store the coffee cup.}
\label{fig:trace_sample_5_p2}
\end{figure}

\paragraph{Execution failures and disturbances.}
Figure~\ref{fig:trace_sample_1} shows a misplaced plant pot being located, retrieved, and placed correctly. Figure~\ref{fig:trace_sample_2} illustrates retrieval after a slipped grasp, while Figure~\ref{fig:trace_sample_3} shows visual search and corrected navigation after reaching the wrong target. Figure~\ref{fig:trace_sample_4} illustrates inspection and retry after a partial drawer opening. Figures~\ref{fig:trace_sample_5} and~\ref{fig:trace_sample_5_p2} follow a two-object storage task: after storing the first object, the cabinet closes again and must be reopened before storing the second.

\begin{figure}[htbp]
\centering
\includegraphics[width=1.0\linewidth]{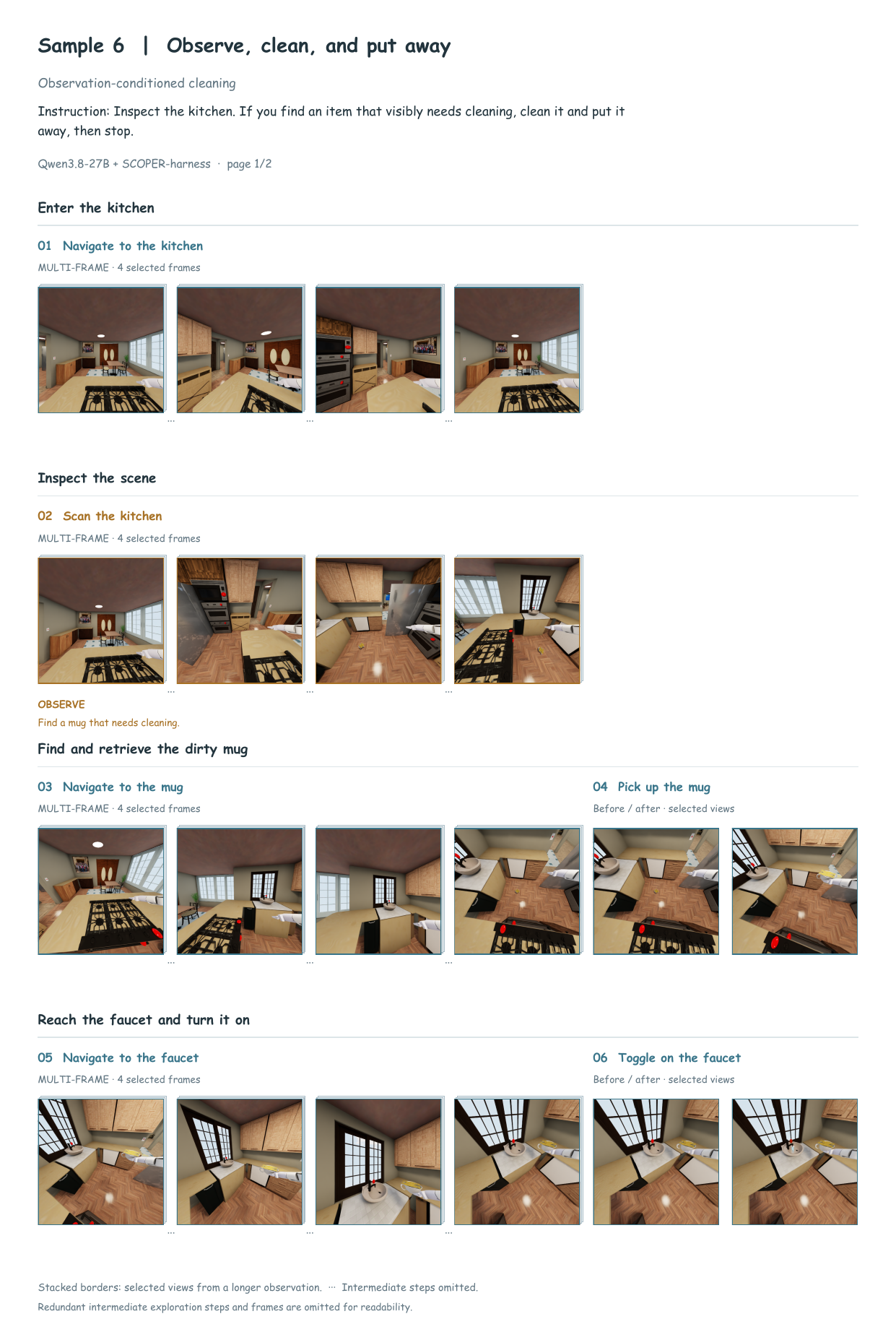}
\caption{Sample 6, part 1: Finding a dirty mug and bringing it to the faucet.}
\label{fig:trace_sample_6}
\end{figure}

\begin{figure}[htbp]
\centering
\includegraphics[width=1.0\linewidth]{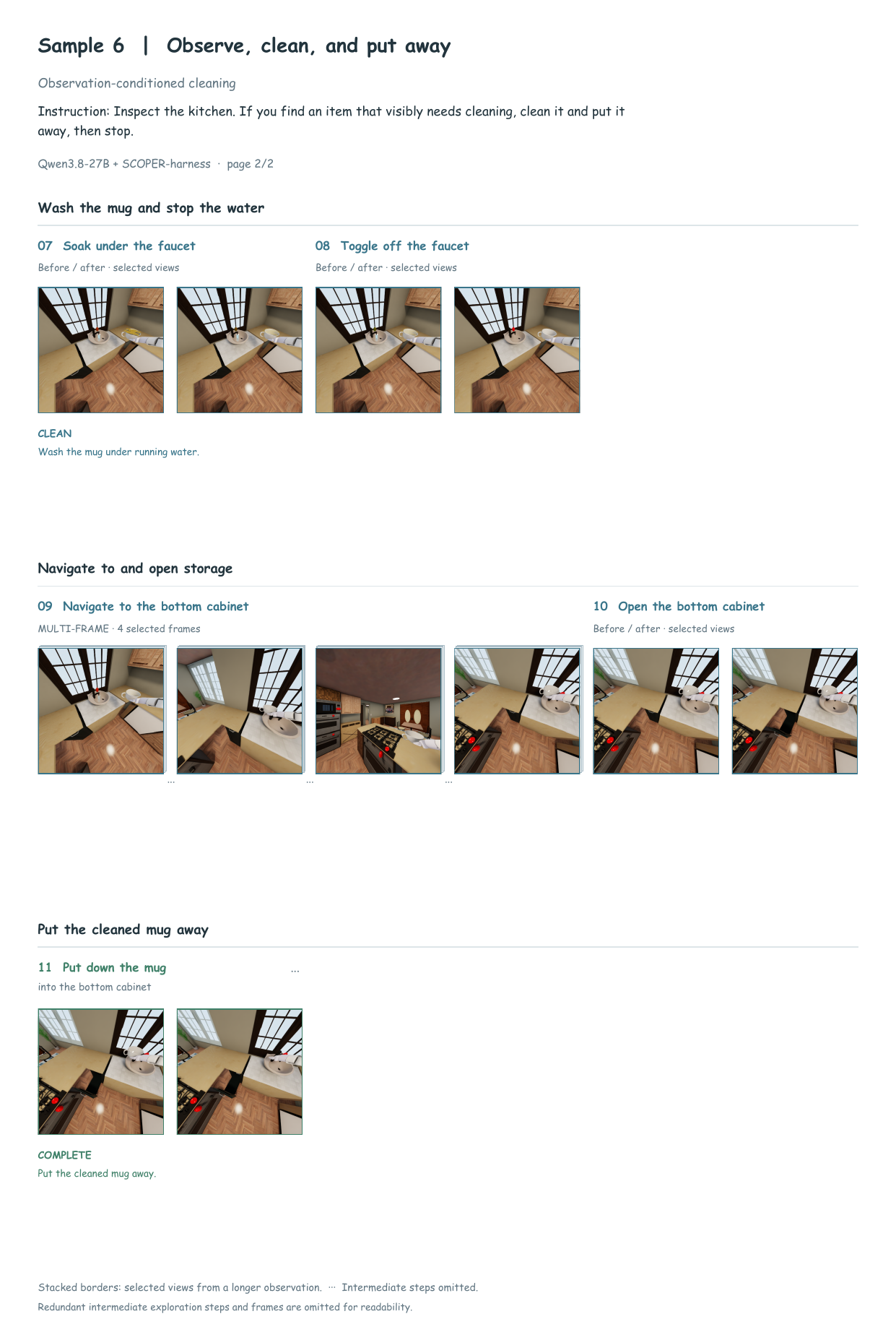}
\caption{Sample 6, part 2: Washing the mug and putting it away.}
\label{fig:trace_sample_6_p2}
\end{figure}

\begin{figure}[htbp]
\centering
\includegraphics[width=1.0\linewidth]{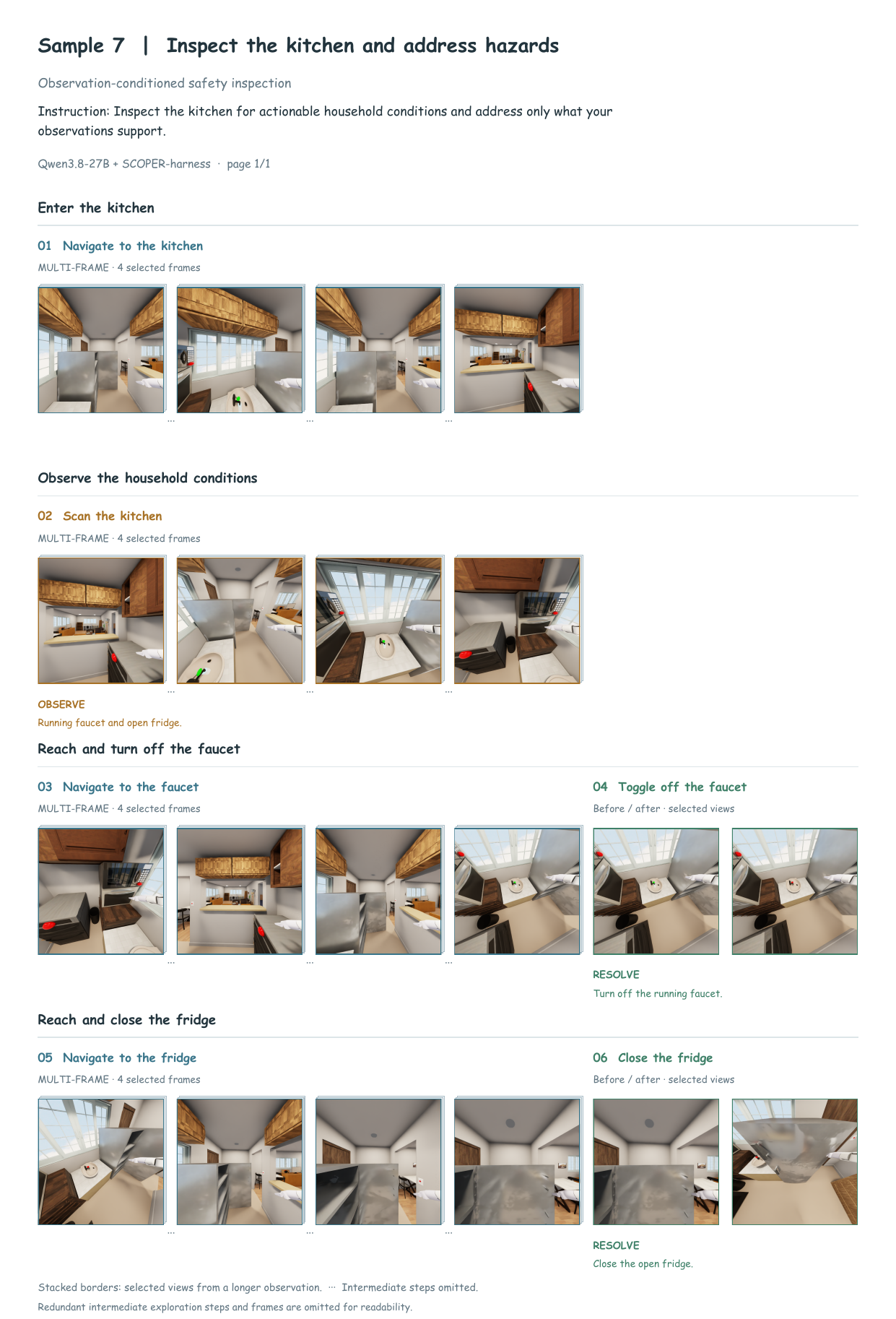}
\caption{Sample 7: Turning off an observed running faucet and closing an open fridge.}
\label{fig:trace_sample_7}
\end{figure}

\begin{figure}[htbp]
\centering
\includegraphics[width=1.0\linewidth]{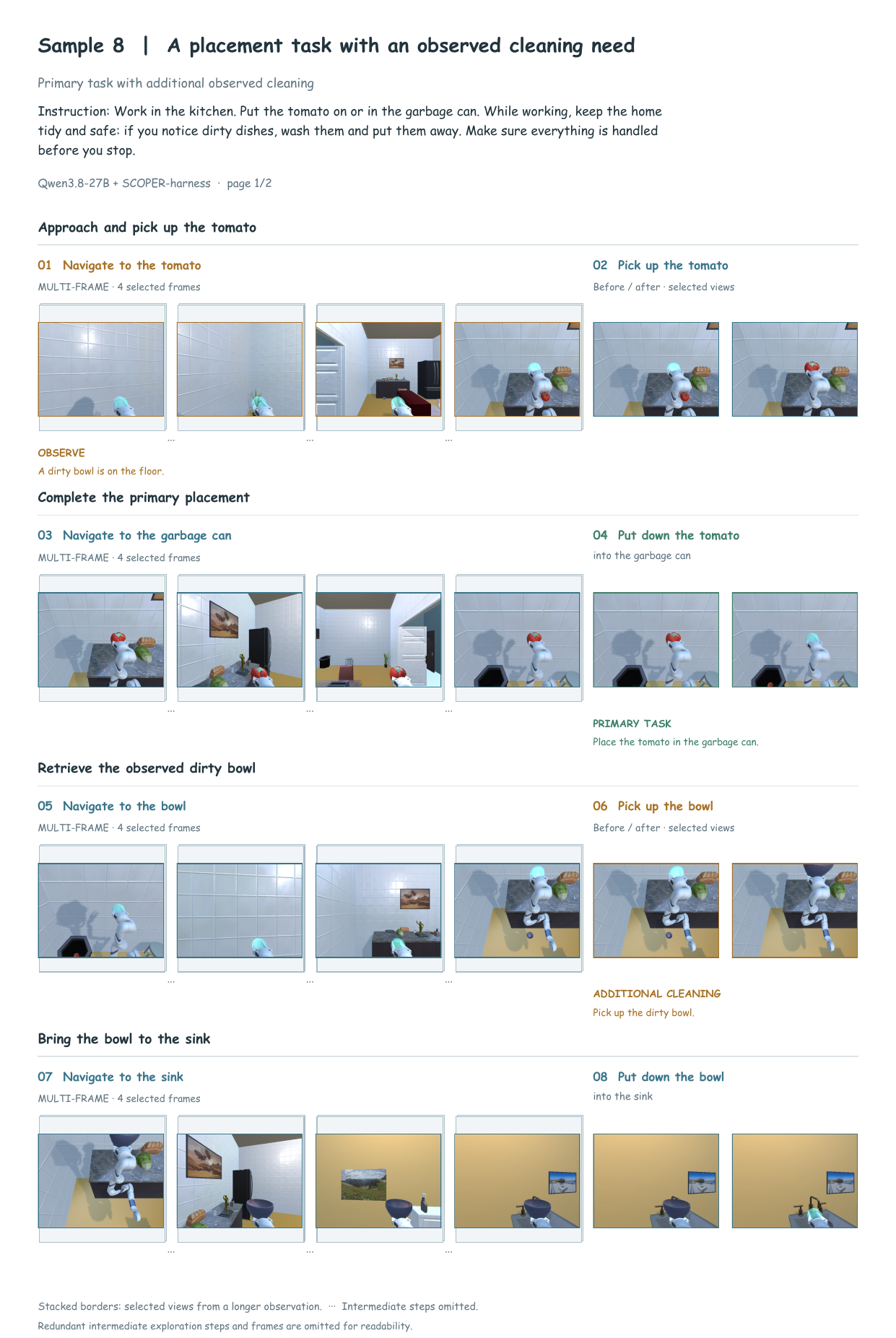}
\caption{Sample 8, part 1: Placing the tomato and retrieving an observed dirty bowl.}
\label{fig:trace_sample_8}
\end{figure}

\begin{figure}[htbp]
\centering
\includegraphics[width=1.0\linewidth]{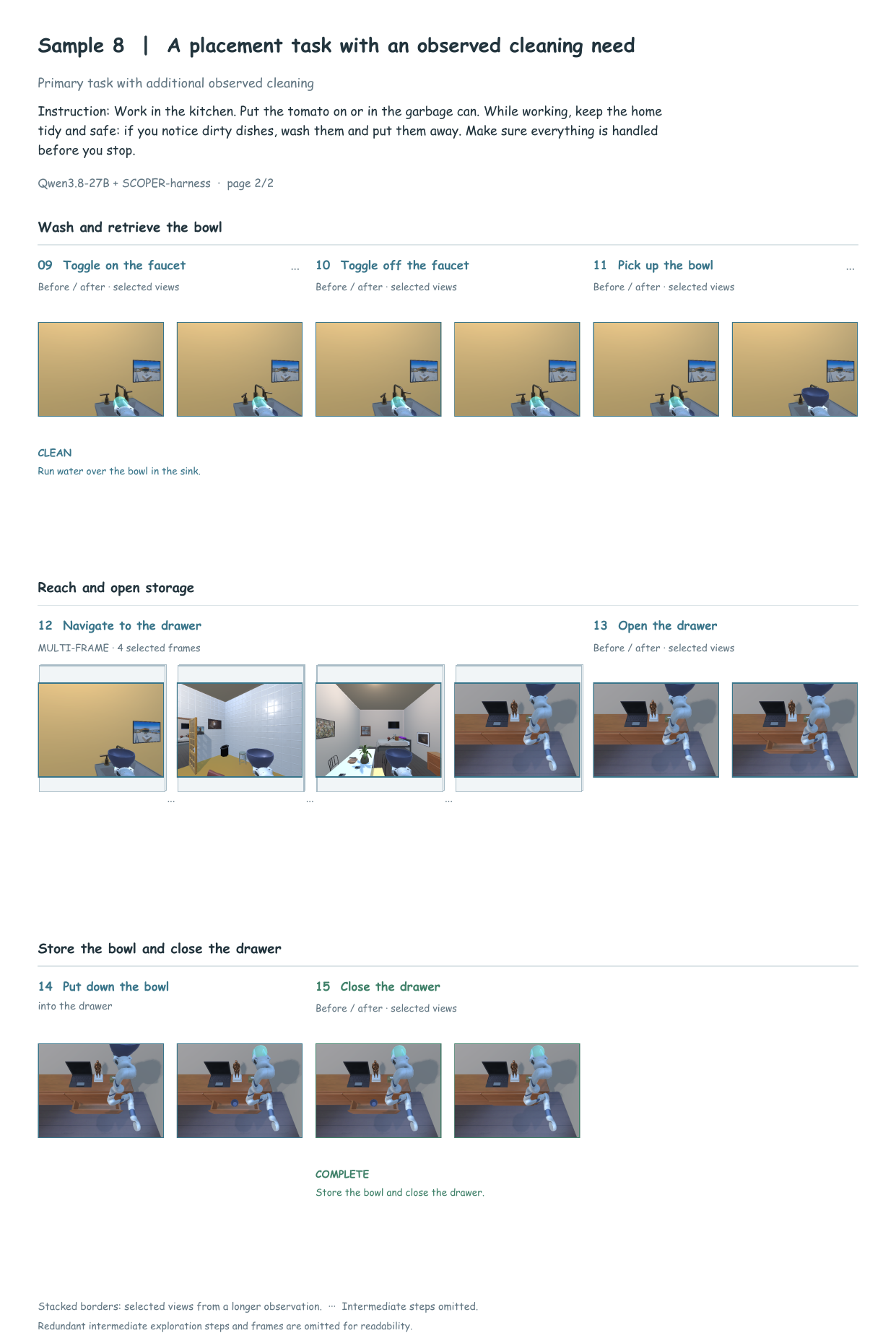}
\caption{Sample 8, part 2: Washing and storing the bowl as required by the instruction.}
\label{fig:trace_sample_8_p2}
\end{figure}

\begin{figure}[htbp]
\centering
\includegraphics[width=1.0\linewidth]{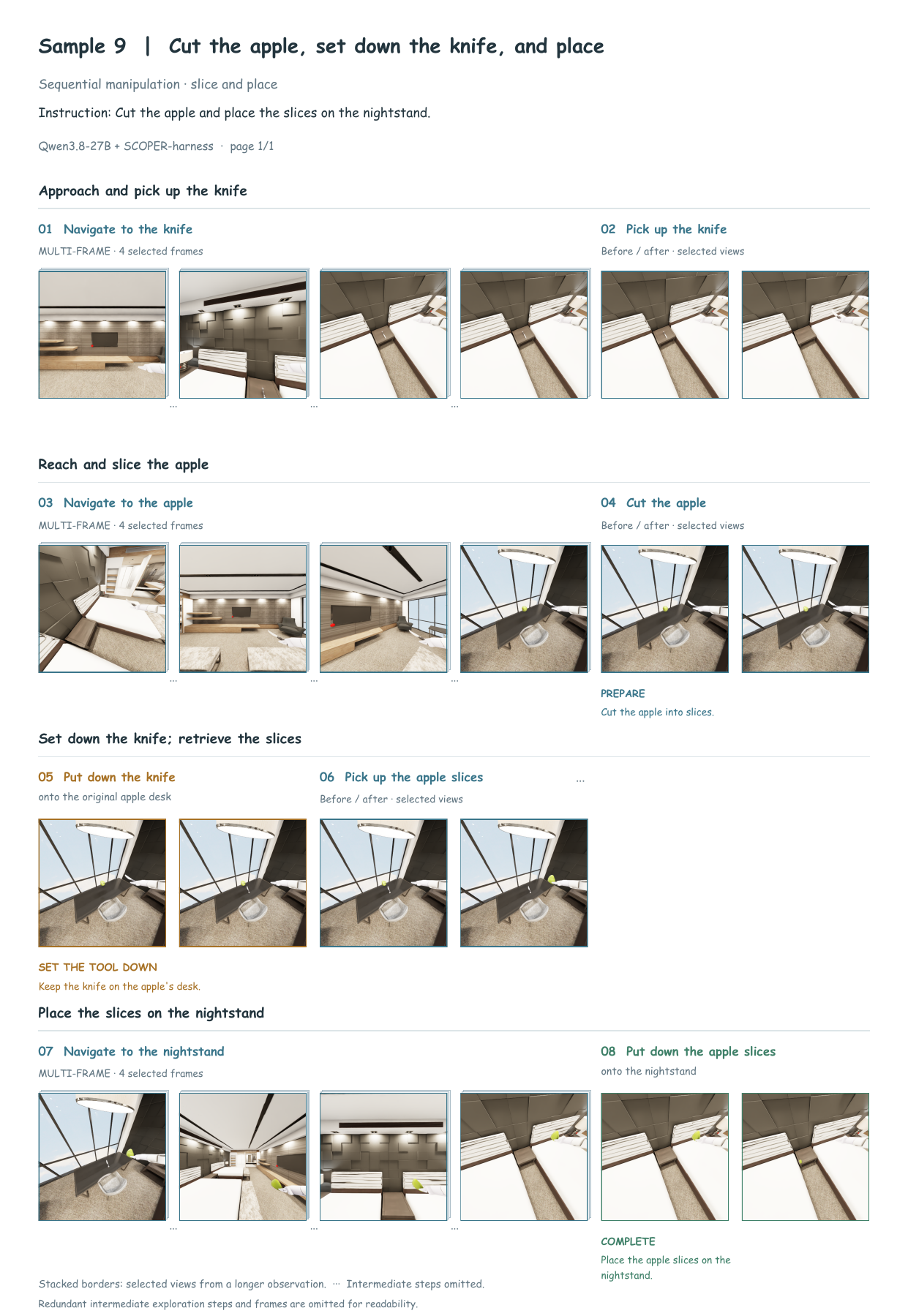}
\caption{Sample 9: Setting down the knife before picking up and placing the apple slices.}
\label{fig:trace_sample_9}
\end{figure}

\paragraph{Observation-dependent requirements and action dependencies.}
Figures~\ref{fig:trace_sample_6} and~\ref{fig:trace_sample_6_p2} show discovery of a dirty mug followed by cleaning and storage. Figure~\ref{fig:trace_sample_7} illustrates a safety inspection in which observations identify a running faucet and an open fridge. Figures~\ref{fig:trace_sample_8} and~\ref{fig:trace_sample_8_p2} show observation of a dirty bowl activating the instruction-specified cleaning and storage requirement during a tomato-placement task. Figure~\ref{fig:trace_sample_9} illustrates sequential dependencies: the knife must be put down after slicing before the apple slices can be picked up and placed.

\begin{figure}[htbp]
\centering
\includegraphics[width=1.0\linewidth]{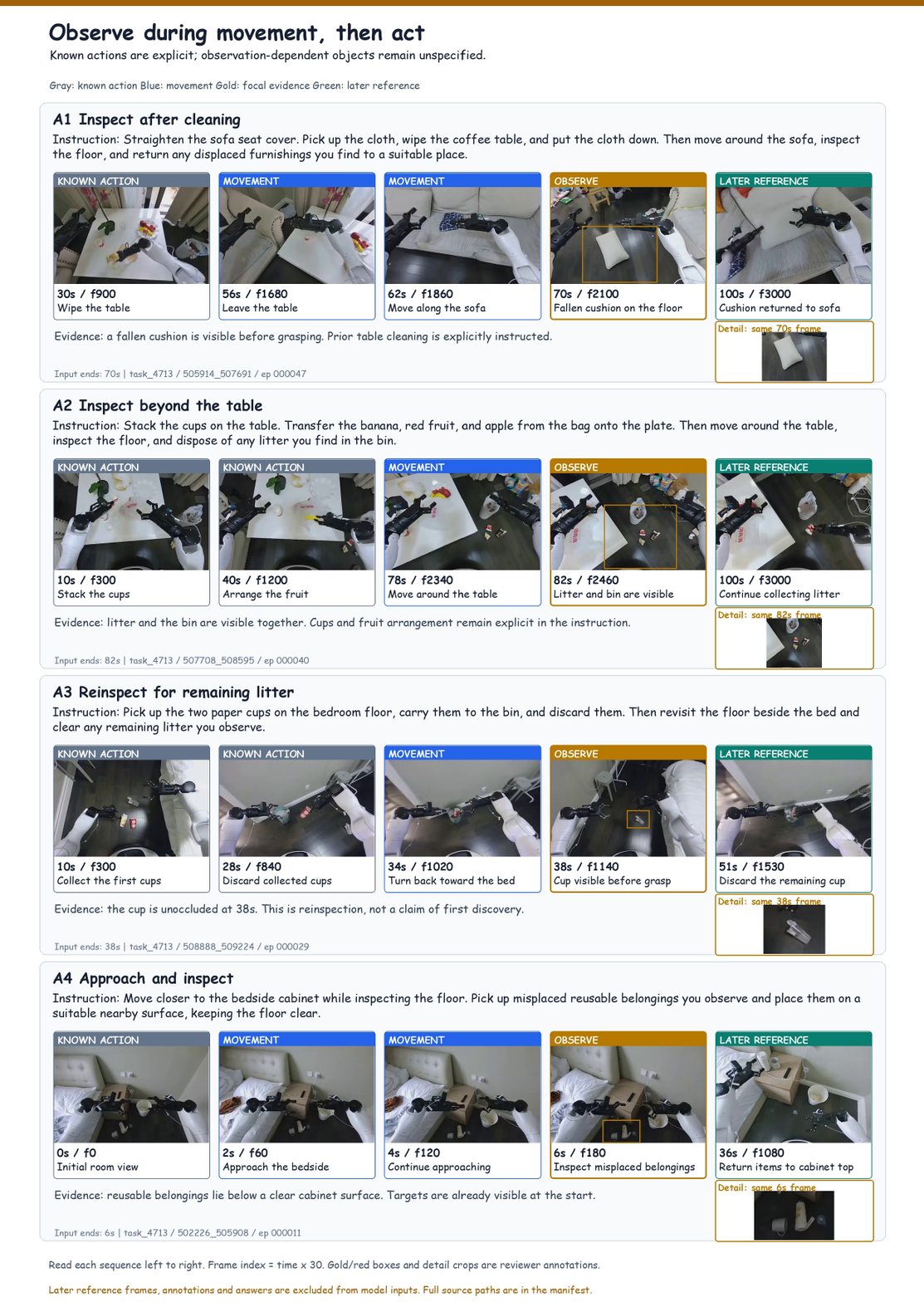}
\caption{Real-data examples: Observing action-relevant evidence through movement and reinspection.}
\label{fig:real_sample_1}
\end{figure}

\begin{figure}[htbp]
\centering
\includegraphics[width=1.0\linewidth]{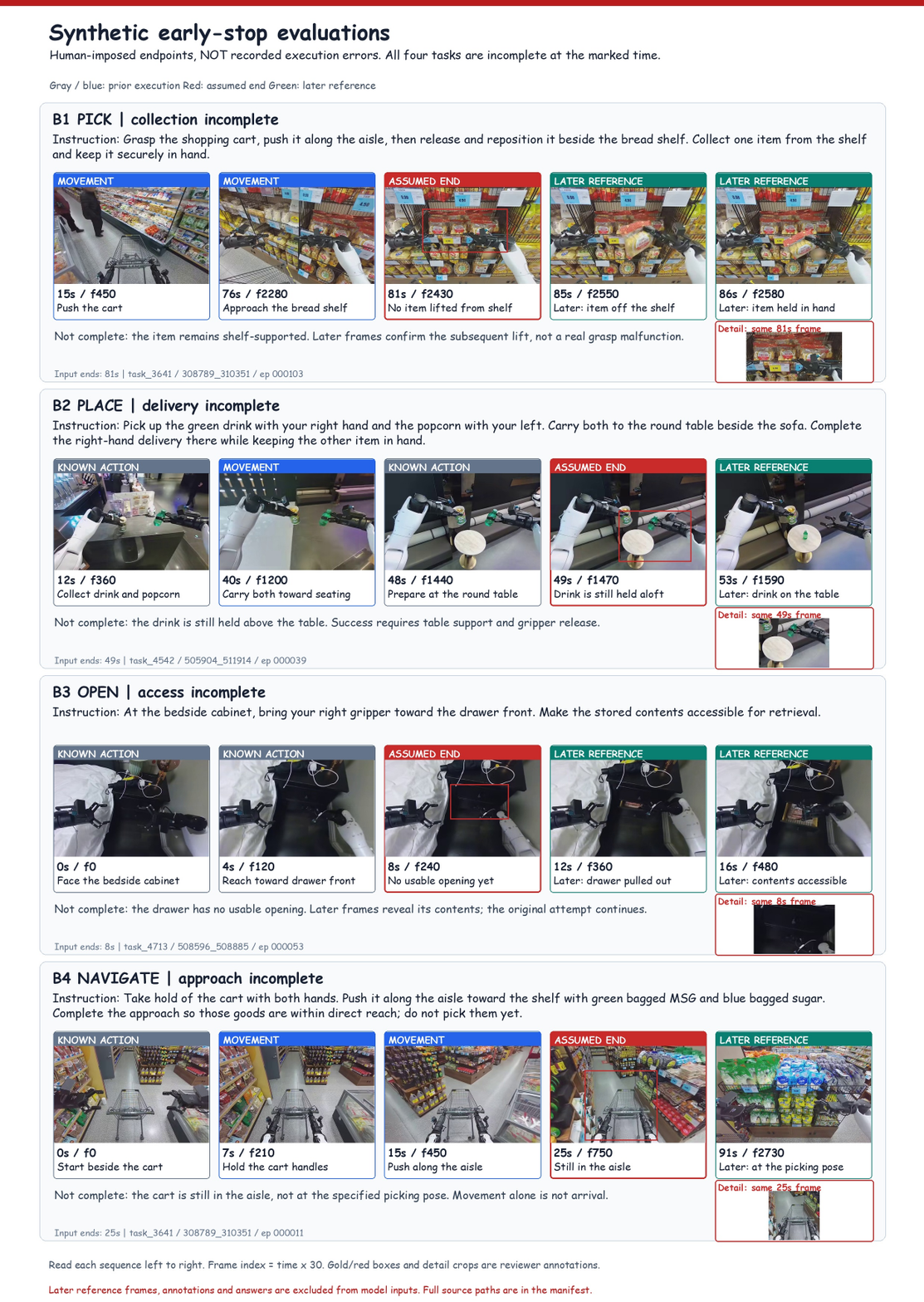}
\caption{Real-data examples: Synthetic early-stop nodes with incomplete actions.}
\label{fig:real_sample_2}
\end{figure}

\paragraph{Real-observation decision nodes.}
Figure~\ref{fig:real_sample_1} illustrates movement and reinspection revealing evidence relevant to the next action, such as a fallen cushion or remaining litter. Figure~\ref{fig:real_sample_2} illustrates synthetic early-stop decision nodes: picking, placing, opening, or approaching is still incomplete at the marked endpoint. These are imposed truncations of reference demonstrations, rather than recorded execution malfunctions. Later reference frames illustrate eventual outcomes and are excluded from model inputs.

\clearpage
\section{Harness Model-Call Overhead}
\label{app:call_overhead}

Direct execution uses one model call per high-level step. SCOPER
 adds role-specific calls around action selection: the policy usually
 makes one call per step; discovery makes one call when new images
 are available; reconciliation makes one call when existing facts
 can be checked; agenda revision makes one call when the instruction
 or committed memory changes; and the verifier makes one call after
 an admitted operation requiring outcome verification. These calls
 support action selection, acquisition of new scene facts, assessment
 of earlier evidence, task-progress updates, and visual outcome checks,
 respectively. Averaged over evaluation on the complete 870-task
 benchmark, SCOPER uses approximately
 $3.7\times$ as many model calls as direct execution. Internal calls
 do not add environment-interaction steps; both methods retain the
 same high-level step budgets.

\end{document}